\documentclass[12pt]{report}
\usepackage[utf8]{inputenc}
\usepackage{amsmath, amsfonts, amssymb}
\usepackage{graphicx}
\usepackage{array}
\usepackage{lipsum}
\usepackage{listings}
\usepackage[colorlinks=true, linkcolor=blue, citecolor=blue, urlcolor=blue]{hyperref}
\usepackage{xcolor}

\definecolor{codegray}{rgb}{0.95,0.95,0.95}
\definecolor{commentgreen}{rgb}{0,0.5,0}

\begin{document}
\title{Future-Proof Yourself: An AI Era Survival Guide}
\author{Prof. Taehoon Kim \\ \href{https://mimic-lab.com}{https://mimic-lab.com}}
\date{}

\maketitle

\tableofcontents

\chapter{Introduction to Key Concepts in Machine Learning and Deep Learning}

\section{Introduction to Machine Learning}
Machine learning is the study of algorithms that improve their performance at some task through experience. A classic definition by Mitchell (1997) states: \emph{``A computer program is said to learn from experience $E$ with respect to some class of tasks $T$ and performance measure $P$, if its performance at tasks in $T$, as measured by $P$, improves with experience $E$.''} \cite{Mitchell-1997}. In simpler terms, rather than being explicitly programmed with fixed rules, an ML system \textbf{learns} patterns from data. For example, instead of writing a specific set of rules to identify spam emails, one can train a machine learning model on many example emails labeled as “spam” or “not spam” so that the model learns to recognize the patterns of spam emails on its own.

There are several broad categories of machine learning. The most common are:
\begin{itemize}
    \item \textbf{Supervised learning}: The algorithm learns from \emph{labeled} data (i.e., data where the desired output is known). The goal is to predict the labels for new, unseen data. This category includes tasks like classification and regression.
    \item \textbf{Unsupervised learning}: The algorithm learns from \emph{unlabeled} data, trying to discover hidden structures or patterns. Here there are no explicit correct outputs given. Clustering and dimensionality reduction are examples of unsupervised learning methods.
    \item \textbf{Reinforcement learning}: The algorithm (often called an \emph{agent}) learns by interacting with an \emph{environment}. Instead of correct input-output pairs, the agent receives \emph{rewards} or \emph{penalties} for its actions and aims to learn a strategy (policy) to maximize cumulative reward. This paradigm is inspired by behavioral learning and is used in scenarios like game-playing, robotics, and control systems.
\end{itemize}

Each of these paradigms addresses different kinds of problems. In the following sections, we will delve into supervised learning (using a simple example of classifying fish), then contrast it with unsupervised and reinforcement learning, all while introducing key concepts such as model generalization, overfitting, features, and the recent advances brought by deep learning.

\section{Supervised Learning Fundamentals}
In supervised learning, we teach the model using examples that are paired with correct outputs. For instance, imagine we want to build a system to automatically sort fish at a seafood plant. We have two kinds of fish: salmon and sea bass. We can take many pictures of fish, and for each picture, we know whether it’s a \textbf{salmon} or a \textbf{bass}. Our task is to create a classifier that looks at a new fish image and predicts the correct type.

This fish classification scenario is a classic toy example often used to illustrate how supervised learning works \cite{Duda-2001}. To make it concrete, suppose we choose a particular feature of the fish from the image — for example, the average intensity (brightness) of the image, which might correlate with the fish’s coloration. Perhaps salmon tend to be lighter on average and sea bass darker (this is hypothetical for illustration). We could plot this feature for many known salmon and bass. If there is a threshold of intensity that separates the two species reasonably well, our classifier could be as simple as: 
\[ \text{If average intensity $>$ some threshold, predict “salmon”; otherwise predict “bass.”} \] 

This would be a very simple \textbf{model} (essentially a threshold-based decision). We would adjust the threshold using our training examples to best separate the salmon from the bass.

\subsection{Generalization to New Examples}
After training the model (setting the threshold in our example) using some collection of labeled fish images (the \emph{training set}), we want it to perform well on new, unseen fish images. The ability to perform well on new data is called \textbf{generalization}. If our classifier works perfectly on the fish images we used to train (because we tuned it to those examples) but fails on new fish pictures, it hasn’t generalized well.

Imagine our fish sorting model becomes too specific to peculiarities of the training images. For example, maybe all the salmon photos in the training set happened to have a blue background and the sea bass photos had a white background. If the model naively uses background color as a key feature, it might classify any fish with a blue background as salmon, which would obviously fail in general. This scenario hints at an important problem in machine learning: even if a model performs well on the training data, it might not perform well on new data if it learned the \emph{wrong patterns} or overly specific details.

\subsection{Underfitting and Overfitting}
When developing supervised learning models, one must be mindful of model complexity relative to the available data. Two common issues are \textbf{underfitting} and \textbf{overfitting} \cite{ISLR-2013}.

\textbf{Underfitting} occurs when a model is too simple to capture the underlying pattern of the data. In our fish example, if we tried to classify fish using a completely irrelevant feature (say, the day of the week the photo was taken), the model would perform poorly on both training and new data. Underfitting means the model has high bias (it’s systematically wrong because it’s too simplistic).

\textbf{Overfitting} happens when a model is too complex relative to the amount and noise of the data, so it ends up learning spurious details or “noise” in the training set as if they were important patterns. An overfitted model might do extremely well on the training data (even memorizing it) but then fail to generalize to test data, because those quirky details it learned do not recur. In the fish example, if we use a very flexible model (imagine a complicated rule or a very deep decision tree that looks at many pixel positions in the image), it might accidentally pick up meaningless patterns, like a reflection in the water in one specific salmon image, and treat that as a feature of “salmon-ness.” This model might correctly classify every training example (since it effectively memorized them), but then classify new fish in a nonsensical way.

A common analogy for underfitting vs. overfitting is fitting a curve to data points. Suppose the true relationship is a smooth curve. A very low-degree polynomial (like a straight line) will underfit (it cannot bend to fit the data), whereas a very high-degree polynomial can pass through every training point and will overfit (wiggling wildly between points). 

The goal is to find a model complexity that is “just right” – complex enough to capture the true structure of the data but simple enough to avoid modeling noise. Techniques like \textbf{cross-validation} (testing the model on held-out data during training), and regularization methods (which penalize overly complex models) are commonly used to combat overfitting \cite{Goodfellow-2016}. We won’t dive into those techniques here, but it’s important to be aware that they exist as part of the toolbox for building robust models.

\section{Features and Feature Selection}
In the process of building our fish classifier, we mentioned using “average intensity” as a \emph{feature}. In machine learning, a feature is an individual measurable property of the phenomenon being observed. Good features are critical for the performance of many traditional (non-deep-learning) ML algorithms. In early approaches to ML, a lot of effort was put into \textbf{feature engineering} – coming up with the right set of features that make it easier for the model to separate classes or make predictions.

For example, in image classification tasks (like our fish example or recognizing handwritten digits), researchers manually designed features such as edges, corners, or textures that could be extracted from the raw pixel data. One classic feature in image recognition is the \textbf{Histogram of Oriented Gradients (HOG)} \cite{Dalal-2005}. HOG features encode local edge directions: essentially, for small patches of the image, one computes how the gradient orientations are distributed. Dalal and Triggs (2005) showed that using HOG features as inputs to a classifier is very effective for detecting pedestrians in images \cite{Dalal-2005}. The HOG feature vector is a summary of shape that is more informative than raw pixels for that task (because it captures outlines of a person without being too sensitive to minor pixel changes).

Another example: for face detection (deciding if an image contains a face), the \textbf{Viola-Jones} algorithm \cite{Viola-2001} was a groundbreaking approach in the early 2000s. It relied on very simple rectangular pattern features (reminiscent of Haar wavelets) that basically measure contrasts (like the difference in pixel intensity between adjacent regions, which can capture things like “the eye region is darker than the cheeks”). Viola and Jones used a machine learning method called AdaBoost to automatically select a small set of these features that are most useful for distinguishing faces from non-faces, and then combined them in a cascade of simple classifiers for efficient detection \cite{Viola-2001}. This approach was efficient enough to run in real time and was the first to enable things like real-time face detection in consumer cameras. It’s a good illustration of how much thought went into designing and selecting the right features in traditional computer vision.

In general, \textbf{feature selection} refers to the process of selecting a subset of relevant features for use in model construction. Especially when you have a very large number of candidate features, some of them may be redundant or irrelevant (and including them could actually hurt the model by making it more prone to overfitting or by slowing it down). Research has shown that eliminating useless features and focusing on the most informative ones can improve learning algorithms’ performance \cite{Guyon-2003}. Feature selection can be done through statistical tests, through algorithms that try different combinations, or via regularization techniques that implicitly drive weights of unimportant features to zero.

To summarize: in traditional machine learning, a lot of the “intelligence” of the solution was in the human-driven step of deciding how to represent the data (which features to use). A famous saying was “\emph{data is the fuel, but feature engineering is the rocket}” – meaning that with the right features, even a simple algorithm can do very well.

\section{From Traditional ML to Deep Learning}
Over the past decade, there has been a shift in how we approach machine learning, especially for complex tasks like image and speech recognition. This shift is from manual feature engineering to \textbf{representation learning}, where the system learns the features by itself. \textbf{Deep learning} refers to machine learning with \textbf{artificial neural networks} that have multiple layers of neurons between input and output. These multi-layer (or “deep”) networks can learn increasingly abstract features from the raw data.

The key advantage of deep learning is that if we provide a large amount of raw data, a deep neural network can learn good features at multiple levels of abstraction, automatically. This has led to extraordinary breakthroughs. For example, in image recognition benchmarks, deep learning approaches started to dominate around 2012. The catalyst was a convolutional neural network by Krizhevsky et al. (commonly known as \textit{AlexNet}) that won the ImageNet image classification challenge by a large margin \cite{Krizhevsky-2012}. AlexNet was trained on millions of images and was able to learn edge detectors in its first layer, simple shape detectors in the next, and eventually very complex structures in deeper layers (as we will discuss in the next section). This was a departure from earlier methods that might use hand-crafted features like HOG or SIFT and then feed them into, say, a support vector machine (SVM) for classification.

To contrast traditional ML with deep learning:
\begin{itemize}
    \item In \textbf{traditional ML}, the pipeline might be: collect data $\to$ manually engineer features $\to$ train a relatively simple model (e.g., linear classifier, decision tree, SVM) on those features.
    \item In \textbf{deep learning}, the pipeline is: collect data (usually a lot more data is needed) $\to$ feed the raw data into a neural network which automatically learns multiple layers of feature transformations $\to$ the final layer of the network produces the predictions. The model (the neural network) is complex but it discovers for itself what features to use.
\end{itemize}

Deep learning has been especially successful in fields like computer vision, speech recognition, and natural language processing, where raw data is high-dimensional and complex, and where we have benefitted from improvements in computing power (GPUs) and the availability of big datasets. LeCun, Bengio, and Hinton (2015) provide a good overview of why deep learning works and what impact it has had \cite{LeCun-2015}. One key reason is that deep networks can express very complicated functions (they are very flexible models, with many tunable parameters), and if regularized and trained on enough data, they can learn the correct complex patterns rather than overfitting random noise. Another reason is that by learning features incrementally (one layer builds on the output of the previous), they can build a hierarchy of concepts – exactly what human engineers were trying to do with multi-stage feature engineering, but now it happens automatically.

However, deep learning models are not always the answer to every problem. They typically require large amounts of data and computational resources to train, and it can be harder to interpret why they make a given decision (they can feel like a “black box”). For many simpler problems with limited data, a well-chosen set of features and a simple model might still be the more practical solution. 

Next, we focus on one of the most important types of deep learning architecture in the context of computer vision: the \textbf{Convolutional Neural Network (CNN)}.

\section{Convolutional Neural Networks (CNNs)}
A Convolutional Neural Network is a type of neural network designed specifically to process grid-like data such as images (which can be seen as a 2D grid of pixels). CNNs are inspired by the way the visual cortex in animals processes images, and they have a special architecture that makes them particularly effective for image recognition and related tasks \cite{LeCun-1998}.

In a regular neural network (often called a fully-connected network), each neuron in one layer is connected to every input in the previous layer. In a CNN, instead, the neurons are arranged in 3D: width, height, and depth (depth corresponds to different feature maps, or channels). The key ideas in CNNs are:
\begin{itemize}
    \item \textbf{Local receptive fields}: Each neuron in a convolutional layer looks at only a small region of the input image (for example, a $5\times5$ patch of pixels) rather than the whole image. This makes sense because useful visual patterns are often local (an edge, a corner, a texture) and we don’t need every neuron to examine the entire image.
    \item \textbf{Convolution operation}: The same set of weights (called a filter or kernel) is used across different positions of the image. The filter “slides” across the image and computes dot products with the local patches. This weight-sharing means the network is looking for the same pattern in different parts of the image. If a certain pattern (say a vertical edge) is useful in one part of an image, it’s probably useful in another part as well. This drastically reduces the number of parameters to learn and encodes translational invariance (the idea that an object’s identity doesn’t change if it shifts position in the image).
    \item \textbf{Pooling}: CNNs often include pooling layers, which down-sample the image representation to make it smaller and more manageable, and to aggregate information. For example, a common operation is \emph{max-pooling} which takes a $2\times2$ block of neurons in one layer and outputs the maximum value in that block to the next layer. Pooling helps make the representation roughly invariant to small translations or distortions (e.g., if an image shifts by one pixel, the pooled representation might stay the same).
\end{itemize}

By stacking multiple convolutional layers (with non-linear activation functions like ReLU in between) and pooling layers, a CNN gradually transforms the input image into higher-level features. The first layer might detect basic edges or color patches, the next layer might combine edges into simple shapes or textures, and deeper layers might detect parts of objects (like an eye, or a wheel of a car) and eventually entire objects or complex scenes.

This hierarchical feature learning is one of the most fascinating aspects of CNNs. For instance, consider a CNN trained to recognize objects in car images. The early layers will likely detect edges and corners. Middle layers might detect patterns like circles (potentially corresponding to tires or headlights) or rectangles (maybe car windows). The later layers, building on those, might activate strongly for entire wheels, car grills, or doors. Finally, the deepest layers integrate all that information to recognize a specific make of car or that the image contains a car at all. In other words, the network builds up a \textbf{composition of features}: edges combine into shapes, shapes into parts, parts into objects. Empirical studies have confirmed this behavior: in a well-trained CNN, neurons in deeper layers often correspond to high-level concepts. For example, one study found that some deep neurons responded to images of cats or dogs, even though no one told the network explicitly what a cat or dog was – those concepts emerged from the data \cite{Zeiler-2014}. 

CNNs have revolutionized image processing. LeCun’s early CNN, called \textbf{LeNet}, was used in the 1990s to recognize handwritten digits on bank checks \cite{LeCun-1998}. It had about 5 layers and was very successful for that task. Modern CNNs are much deeper (often dozens of layers), and thanks to large datasets and powerful GPUs, they can tackle very complex tasks. \textbf{AlexNet} in 2012 had 8 learned layers and was trained on millions of images, achieving unprecedented accuracy on ImageNet \cite{Krizhevsky-2012}. Subsequent architectures like \textbf{VGG}, \textbf{ResNet}, and others went even deeper and introduced new ideas to improve training, leading to further improvements in accuracy.

\subsection{Softmax and Multi-Class Classification}
Many machine learning tasks, including image classification, are not just yes/no decisions but involve choosing between multiple categories. For example, a single CNN might be trained to classify images into 100 different object categories. How does the network output a decision among many classes?

Typically, the final layer of a classification neural network uses a function called \textbf{softmax}. The softmax function converts a vector of raw scores (sometimes called \emph{logits}) from the network into a set of probabilities for each class. Suppose the network’s final outputs (before softmax) are numbers $[z_1, z_2, ..., z_K]$ for $K$ classes. The softmax for class $i$ outputs:
\[ 
P(y = i) = \frac{\exp(z_i)}{\sum_{j=1}^{K} \exp(z_j)}.
\]
Each $P(y=i)$ is in the range $(0,1)$ and all the $P$’s sum to 1, so they can be interpreted as the predicted probability that the input belongs to class $i$. The model’s predicted class would usually be the one with highest probability.

For example, if you input a picture into a trained model and it outputs (after softmax) $[P(\text{cat})=0.1,\; P(\text{dog})=0.7,\; P(\text{car})=0.2]$, then the model is saying it’s 70\% confident the image is a dog, 20\% a car, 10\% a cat (and it would choose “dog” as the final answer).

The training process for such a network uses a loss function called \textbf{cross-entropy}, which measures the discrepancy between the predicted probabilities and the true answer. Without going into formula details, the network adjusts its weights via \textbf{backpropagation} to increase the probability of the correct class for each training example. Over time, the network becomes better at outputting high probabilities for the right class and low for others.

The softmax function is widely used in classification tasks because it neatly generalizes the idea of a sigmoid (logistic) function (which covers the two-class case) to multiple classes, and it provides a probabilistic interpretation of the network’s output. Understanding softmax is important because it’s a core component in many AI systems – from image classifiers to language models.

\section{Unsupervised Learning and Clustering}
Not all learning is about predicting a label. In \textbf{unsupervised learning}, we try to make sense of data without any labeled examples guiding us. One common unsupervised task is \textbf{clustering}: dividing data into groups (clusters) such that items in the same group are more similar to each other than to those in other groups.

Let’s illustrate clustering with a playful example. Imagine we have a collection of animal images that include ducks, rabbits, and hedgehogs, but we have no labels for which image is which animal. A clustering algorithm, when given all the images, might group them into three clusters: one cluster containing mostly ducks, another with rabbits, and another with hedgehogs. In this case, the algorithm has discovered the natural grouping corresponding to the animal types, without ever being told what a “duck” or “rabbit” or “hedgehog” is. We as humans could then look at the clusters and assign meaning (“Cluster 1 seems to be ducks, cluster 2 rabbits, cluster 3 hedgehogs”). This kind of task is useful when you want to find structure in data—like grouping customers by purchasing behavior, grouping news articles by topic, or grouping genes by expression patterns—without pre-specified categories.

A very well-known clustering algorithm is \textbf{k-means clustering} \cite{MacQueen-1967}. The way k-means works is:
\begin{enumerate}
    \item Decide on the number of clusters $k$ you want to find.
    \item Initialize $k$ points in the data space (these will serve as initial \emph{centroids} of clusters).
    \item Assign each data point to the nearest centroid (using a distance metric, typically Euclidean distance).
    \item Recompute each centroid as the mean of all data points assigned to it.
    \item Repeat the assign-and-update-centroids steps until assignments stop changing significantly (i.e., it has converged).
\end{enumerate}
The result is a partition of the dataset into $k$ clusters. This algorithm is simple yet often effective, and it’s been around for a long time (first introduced by MacQueen in 1967 \cite{MacQueen-1967}).

Clustering doesn’t give you definitive answers (because without labels, there isn’t a single “correct” clustering—there can be multiple ways to group data). However, it can be a great exploratory tool. In our animal example, maybe the clustering algorithm actually grouped animals by background color instead of species—then we’d realize we need to extract better features (e.g., focus on the animal shape, not the whole image) for meaningful clustering.

Unsupervised learning includes other techniques beyond clustering, like \textbf{dimensionality reduction} (e.g., Principal Component Analysis) which simplifies data while preserving as much structure as possible, or \textbf{anomaly detection} where the goal is to find unusual data points that don’t fit any cluster well. But clustering is a cornerstone concept to grasp because it contrasts with classification: clustering creates its own labels (cluster IDs) based on inherent similarity, whereas classification needs given labels to learn from.

\subsection{What Makes Reinforcement Learning Different?}
\textbf{Reinforcement Learning (RL)} is distinct from both supervised and unsupervised learning:
\begin{itemize}
    \item In \textbf{supervised learning}, each training example comes with a correct label or target value, and the system tries to map inputs to these targets.
    \item In \textbf{unsupervised learning}, there are no explicit labels. The system seeks structure or patterns in the data.
    \item In \textbf{reinforcement learning}, an agent interacts with an environment and observes rewards or penalties (rather than correct labels). The agent must figure out the best sequence of actions to maximize long-term reward.
\end{itemize}

In many supervised tasks, you have a fixed dataset. The learning algorithm passively observes examples and tries to generalize. By contrast, in RL, the agent’s actions influence the data it sees next, because the environment responds to those actions. This creates a \textbf{feedback loop} where the agent’s decisions directly affect future inputs and future possibilities for earning rewards.

\subsection{The Potential of Simulation-Based Learning}
A key advantage of reinforcement learning is that it does not necessarily require a static dataset. Instead, the agent can generate its own experience by exploring an environment. This environment could be:
\begin{itemize}
    \item A real physical system (for example, a robot interacting with the real world).
    \item A virtual simulator (such as a physics engine for robotics, a video game, or a custom simulation of a supply chain).
\end{itemize}

When an environment can be \textbf{simulated}, the agent has the flexibility to run many trials without the cost or risk of doing so in the real world. This is common in training AI agents for control tasks or robotics, where real-world experimentation could be expensive, slow, or dangerous. In a simulator, the agent can attempt actions and accumulate thousands of hours of experience in a short time. Once it learns a good policy in simulation, the policy can be transferred to the real environment (sometimes requiring additional fine-tuning, but it can still be far more efficient than learning purely in the real environment).

This focus on \textbf{experience-driven} learning is why RL is sometimes described as a more dynamic or interactive style of learning, compared to the static, data-driven approaches of supervised and unsupervised methods. The agent does not need an entire labeled dataset up front; it creates its own data through exploration.

\subsection{The Exploration vs. Exploitation Dilemma}
One of the central challenges in RL is deciding whether to \textbf{exploit} what you already know to earn higher immediate reward, or to \textbf{explore} new actions that might yield even larger rewards in the future. This is often called the \emph{exploration-exploitation trade-off} \cite{Sutton-1998}.

\begin{itemize}
    \item \textbf{Exploitation}: The agent uses its current knowledge to pick the best known action in a given state. If the agent already believes one action has the highest payoff, it consistently chooses that action to maximize immediate reward.
    \item \textbf{Exploration}: The agent tries actions that it is less certain about, to gain more information. Even if these actions seem worse initially, they might lead to discovering better strategies that yield higher rewards in the long run.
\end{itemize}

A simple example is the \emph{multi-armed bandit} problem, where you have multiple slot machines (bandits) each with an unknown payout probability. If you only exploit, you keep using the arm that gave the highest payout so far, but you might miss the chance of finding an even better arm. If you explore too much, you waste time testing different arms instead of reaping the reliable rewards of the best arm found so far. RL algorithms must balance this dilemma, and there are many approaches (like epsilon-greedy, Upper Confidence Bound, and Thompson Sampling) to ensure both adequate exploration and exploitation.

\subsection{Reward and Sequential Decision-Making}
Unlike supervised learning, where you know the correct label for each training example, reinforcement learning’s feedback is a \textbf{reward signal}. The agent does not know the correct action for a given state; it only receives a numerical reward (positive or negative) after it acts. Sometimes the reward might come much later, so the agent also faces the \emph{credit assignment} problem: figuring out which past actions were responsible for receiving a future reward.

Furthermore, RL deals with \textbf{sequential} decisions. Each action the agent takes affects the future state and the rewards that can be obtained. In other words, states, actions, and rewards unfold over time, and the agent seeks to maximize cumulative (or discounted) reward. This makes RL particularly suited to tasks like robotics control, game-playing, or any scenario where decisions happen in a sequence and one needs to plan ahead.

\subsection{Examples and Impact}
One famous success of RL is \textbf{AlphaGo} by DeepMind, which combined reinforcement learning and deep neural networks to master the game of Go at a superhuman level \cite{Silver-2016}. AlphaGo learned by playing millions of Go games against itself (self-play), receiving +1 reward for a win and -1 for a loss. Through exploration and exploitation, it discovered strategies humans had never used.

Another example is training agents to play Atari video games from raw pixels \cite{Mnih-2015}. The agent uses the game score as its reward, exploring different joystick actions to improve performance over time. In robotics, RL can be used to learn locomotion skills, manipulate objects, or navigate around obstacles \cite{Kober-2013}.

\subsection{Model-Based vs. Model-Free Approaches}
In some RL methods, called \textbf{model-based RL}, the agent tries to learn or leverage a model of the environment’s dynamics—basically predicting the next state and reward given the current state and action. It can then perform \textbf{planning} by simulating different future scenarios internally. In \textbf{model-free RL}, the agent does not attempt to learn such a model; it directly learns a policy or value function from experience. Both methods have pros and cons, and ongoing research tries to combine them effectively.

\subsection{Why Is Reinforcement Learning Important?}
Reinforcement learning is compelling because it addresses problems where:
\begin{itemize}
    \item You do not have labeled examples telling you the correct actions.
    \item You can repeatedly interact with a system or simulator to gather experience.
    \item You want to optimize long-term returns rather than immediate gains.
\end{itemize}

This general framework is suitable for many real-world tasks, from resource allocation and scheduling to game-playing and robotics. RL’s power lies in its ability to learn from trial and error, potentially discovering strategies better than those programmed by humans.

The foundational text in this field is the book by Sutton and Barto, \emph{Reinforcement Learning: An Introduction} \cite{Sutton-1998}, which provides both the conceptual framework and the basic algorithms (like Q-learning, policy gradients, etc.) that let agents learn optimal behaviors over time.

\section{Conclusion}
In this chapter, we covered a broad spectrum of fundamental concepts in machine learning and introduced how deep learning extends these ideas. We began with the notion of learning from data, emphasizing the difference between supervised learning (learning from labeled examples), unsupervised learning (finding structure without labels), and reinforcement learning (learning via reward and punishment through interaction). Through the example of the salmon vs. bass fish classifier, we illustrated what it means for a model to learn a decision boundary and the pitfalls of underfitting (model too simple) and overfitting (model too complex and tuned to noise). We discussed the importance of features in traditional ML and how methods like feature selection and hand-crafted descriptors (e.g., HOG for image detection or Haar features for face detection) were crucial in earlier approaches.

We then described the paradigm shift brought by deep learning, where feature extraction is no longer manual but learned by the layers of a neural network. Convolutional Neural Networks exemplify this by learning low-level to high-level features directly from image pixels, enabling state-of-the-art performance in vision tasks. We explained how CNNs work, in intuitive terms, and introduced the softmax function as the key to making multi-class predictions with neural networks.

In unsupervised learning, we saw how algorithms like k-means can let us discover hidden groupings (like clustering animals by similarity), which is a different goal than predicting a specific label. And in reinforcement learning, we saw a completely different learning setup where an agent learns from trial and error using feedback in the form of rewards, allowing AI to achieve goals in dynamic environments (for example, mastering games or controlling robots).

To a newcomer, this might seem like a lot of diverse concepts, but the unifying theme is \textbf{learning from data}. Whether it’s fitting a curve, choosing features, adjusting millions of neural network weights, clustering data points, or updating an agent’s strategy – in all cases the system is improving its performance by observing examples or feedback. The specific techniques differ, but they all move beyond explicitly programming every detail, and instead, the programmer provides a framework (like a model structure or a reward function) and the data or experience, and the algorithm \emph{figures out} a good solution.

As you continue in AI and machine learning, you’ll delve deeper into each of these topics. Supervised learning will lead you to numerous algorithms (from linear regression and decision trees to SVMs and deep networks) and practices for model evaluation. Unsupervised learning will introduce you to methods for discovering patterns and compressing data. Reinforcement learning will teach you about balancing exploration and exploitation and optimizing long-term returns. And deep learning will open up a range of specialized architectures (CNNs for images, RNNs and transformers for sequences, etc.) and tricks for training them. 

The concepts covered here form a foundation: understanding what it means to overfit, why features matter, how learning paradigms differ, and what makes deep learning powerful. With this foundation, you can better appreciate both the potential and the limitations of AI systems. Machine learning is a fast-evolving field, but these core ideas remain relevant and will help you navigate more advanced material and real-world applications. 
\chapter{Optimization}

\section{Introduction}
In the field of artificial intelligence (AI), \emph{neural networks} are computational models inspired by the brain’s interconnected neurons. Modern networks can be \emph{very deep}, containing many layers, and they can solve complex tasks such as image classification, language translation, or speech recognition. But how do these networks actually “learn” appropriate parameters (weights and biases) to make accurate predictions?

The core answer involves two key ideas:
\begin{enumerate}
    \item A \textbf{loss function}, which measures how far off the network's predictions are from the desired answers.
    \item An \textbf{optimization} process (most commonly gradient descent) that iteratively updates the network’s parameters to minimize the loss.
\end{enumerate}

This chapter provides a thorough but beginner-friendly explanation of these building blocks:
\begin{itemize}
    \item \textbf{Loss Functions} and \textbf{Cross-Entropy.} We explore how cross-entropy helps quantify prediction error in classification tasks.
    \item \textbf{Gradient Descent and Backpropagation.} We explain how a network’s parameters are updated, and how backpropagation efficiently computes all the necessary gradients.
    \item \textbf{Difficulties in Training Deep Networks.} We address the \emph{vanishing gradient problem} and see why deeper networks can be harder to train.
    \item \textbf{Skip Connections.} We learn how architectural innovations such as \emph{ResNet} address these difficulties and allow us to train very deep networks effectively.
\end{itemize}

No prior background in AI or machine learning is required. Our goal is to provide enough detail for you to understand both the “big picture” and the trickier subtleties that often confuse newcomers.

\section{Loss Functions: Measuring Model Error}

\subsection{Why We Need a Loss Function}
At a high level, a neural network is simply a function mapping an input \(\mathbf{x}\) (e.g., an image) to an output \(\hat{\mathbf{y}}\) (e.g., predicted probabilities over classes). During \emph{training}, we compare the network’s predictions \(\hat{\mathbf{y}}\) to the \emph{true labels} \(\mathbf{y}\) in our dataset, then adjust the network’s internal parameters to make \(\hat{\mathbf{y}}\) match \(\mathbf{y}\) more closely.

But how do we quantify “how close” \(\hat{\mathbf{y}}\) is to \(\mathbf{y}\)? This is where a \textbf{loss function} (sometimes called a \emph{cost function} or \emph{objective function}) comes in. It outputs a single number indicating the overall discrepancy between predictions and the actual target. The lower the loss, the more accurate (or more confident) the predictions are.

\subsection{Probability and Likelihood in ML (A Deeper Look)}
In statistics and machine learning, understanding \emph{probability} and \emph{likelihood} helps clarify why certain loss functions are favored:
\begin{itemize}
    \item \textbf{Probability} measures how likely an outcome is, given a specific model. For example, if a model says “there is a 70\% chance of heads,” that is a probability statement for a coin flip. 
    \item \textbf{Likelihood} measures how well a set of parameters (within a model) explains observed data. Suppose we have data (e.g., a sequence of coin flips), and we want to see which parameter setting best “fits” these observations. We compute the \emph{likelihood} of those parameters by considering the product of probabilities assigned to each observed outcome.
\end{itemize}
In \textbf{classification}, we typically assume our neural network outputs “probabilities” for each class. Then, \emph{maximizing likelihood} of the observed training data is equivalent to \emph{minimizing the negative log-likelihood}, which leads us directly to \textbf{cross-entropy loss}.

\subsection{Information Theory and Entropy}

To really see where cross-entropy comes from, let us step briefly into \emph{information theory}:
\begin{itemize}
    \item \textbf{Entropy} (\(H\)) measures the uncertainty in a distribution. For a discrete distribution \(p(x)\) over outcomes \(x\), the entropy is:
    \[
    H(p) = -\sum_{x} p(x) \log_2 p(x).
    \]
    If an event is certain (\(p=1\)), its entropy contribution is 0 (no surprise). If multiple outcomes are equally likely, the entropy is higher because each event is more “surprising.”
    \item \textbf{Cross-Entropy} (\(H(p,q)\)) measures how well one distribution \(q\) \emph{predicts} or \emph{models} another distribution \(p\). Formally:
    \[
    H(p,q) = -\sum_{x} p(x)\,\log_2\,q(x).
    \]
    If \(p\) and \(q\) match perfectly, \(H(p,q) = H(p)\). If they differ, cross-entropy is larger.
\end{itemize}

\subsubsection{Relating Cross-Entropy to KL Divergence}
The difference between \(H(p,q)\) and \(H(p)\) is the \textbf{Kullback–Leibler (KL) Divergence} \(D_{\mathrm{KL}}(p\|q)\). Symbolically:
\[
H(p,q) = H(p) + D_{\mathrm{KL}}(p\|q).
\]
Because \(D_{\mathrm{KL}}(p\|q) \ge 0\), cross-entropy is always at least the entropy of the true distribution. Intuitively, the worse \(q\) fits \(p\), the bigger the gap.

\subsection{Cross-Entropy in Classification}

Consider a classification problem with \(C\) possible classes. We denote the true label of a training example by a \emph{one-hot} vector \(\mathbf{y}\). For example, if the correct class is 2 out of \(\{1,2,3\}\), then \(\mathbf{y} = [0,\,1,\,0]\). Let \(\hat{\mathbf{y}} = [\hat{y}_1,\dots,\hat{y}_C]\) be the network’s predicted probabilities for each class, with \(\sum_{c=1}^C \hat{y}_c = 1\).

The cross-entropy loss for this one example is:
\[
L_{\text{CE}}(\mathbf{y},\hat{\mathbf{y}}) 
= -\sum_{c=1}^{C} y_c\,\log \hat{y}_c 
= -\log \hat{y}_{\text{correct}},
\]
because \(y_{\text{correct}} = 1\) and \(y_{c\neq \text{correct}} = 0\). In words, it is the negative log of the predicted probability of the correct class. 

\paragraph{Why is this a good loss?}
\begin{itemize}
    \item \textbf{Differentiability.} The cross-entropy loss is differentiable almost everywhere, so it works nicely with gradient-based optimization.
    \item \textbf{Strongly penalizes wrong confident predictions.} If \(\hat{y}_{\text{correct}} = 0.001\), the loss is large (\(\approx -\log(0.001)= 6.9\)), giving a strong signal to “correct” the parameters.
    \item \textbf{Maximum likelihood interpretation.} Minimizing cross-entropy is equivalent to maximizing the probability (likelihood) assigned to the correct labels. This ties in neatly with fundamental statistical principles \cite{Bishop2006}.
\end{itemize}

Thus, cross-entropy is now the most common loss for classification tasks, from simple digit recognition to large-scale image classification or language modeling.

\section{Gradient Descent: An Overview of Optimization}
Once we have a loss function that tells us how far off our predictions are, \emph{how do we make the loss smaller}? We need an \textbf{optimization} procedure. The dominant choice in neural networks is \textbf{gradient descent} (GD) or one of its variants.

\subsection{Local Minima, Global Minima, and Saddle Points}
It helps to visualize the loss function as a high-dimensional “landscape.” The shape can be quite complicated, with many peaks (high loss) and valleys (low loss). A \emph{global minimum} is the absolute lowest point. A \emph{local minimum} is a point where the loss cannot be decreased by any small step, but it might not be the global lowest. Additionally, neural network loss landscapes can have \emph{saddle points}, where the gradient is zero yet the point is neither a true local minimum nor a maximum. 

A key fact in modern neural networks is that we rarely find a single global minimum. Instead, we converge to a “good enough” local minimum or a low-loss region. Empirically, in high-dimensional spaces, many local minima or saddle regions can yield similarly good generalization performance, so the distinction between a local vs. global minimum is not always as critical as once feared.

\subsection{Basic Gradient Descent}
Let \(\theta\) represent all network parameters (weights and biases). We define our loss over the entire training set as \(L(\theta)\). \emph{Gradient descent} updates \(\theta\) iteratively:
\[
\theta \leftarrow \theta - \eta \,\nabla_{\theta} L(\theta).
\]
Here, \(\nabla_{\theta} L(\theta)\) is the \textbf{gradient} of \(L\) w.r.t.\ \(\theta\). It is a vector pointing in the direction of steepest \emph{increase} in the loss. By subtracting this term, we move \emph{downhill} in parameter space. The constant \(\eta\) is the \textbf{learning rate} that controls step size:
\begin{itemize}
    \item If \(\eta\) is too large, we might jump over minima or diverge.
    \item If \(\eta\) is too small, training becomes very slow.
\end{itemize}

\paragraph{Why the Gradient?}
In high-dimensional spaces with millions of parameters, you need a systematic way to figure out how changes in each parameter affect the loss. The gradient precisely captures these sensitivities. By following the negative gradient, each parameter moves in whatever direction decreases the loss \emph{most quickly}, at least locally.

\subsection{Stochastic and Mini-Batch Gradient Descent}
\textbf{Batch gradient descent} recalculates the gradient using the entire dataset at each step, which can be computationally expensive. Thus, two common alternatives are:
\begin{itemize}
    \item \textbf{Stochastic Gradient Descent (SGD):} Update parameters using the loss gradient from a single randomly chosen example at a time. This introduces noise (because one example might not be representative of the entire dataset), but it can help jump out of local minima.
    \item \textbf{Mini-Batch Gradient Descent:} A hybrid approach that uses a small batch of examples (e.g., 32, 64, or 128) to compute the gradient. This is efficient on modern hardware (which can process batches in parallel) and is the de facto standard in deep learning practice \cite{Goodfellow2016}.
\end{itemize}

\subsection{Variants: Momentum, Adam, and Beyond}
The straightforward gradient descent update can be improved by incorporating:
\begin{itemize}
    \item \textbf{Momentum.} We keep a “velocity” vector \(\mathbf{v}\) that accumulates a fraction \(\beta\) (e.g., 0.9) of previous updates. This helps the parameter updates move smoothly in directions consistent across batches and damps oscillations in directions that frequently change sign.
    \item \textbf{Adam.} Combines ideas from momentum and adaptive learning rates, where each parameter’s update scale is adapted based on recent gradients’ magnitudes. This often leads to faster convergence in practice.
\end{itemize}

Regardless of the specific variant, all rely on the \textbf{gradient}, which must be computed efficiently. For neural networks, this is done through \textbf{backpropagation}.

\section{Backpropagation: Computing Gradients in Deep Networks}
\subsection{Chain Rule Refresher}
In a multi-layer network, each layer transforms its input via a function, and the final output is used to compute the loss. To find \(\frac{\partial L}{\partial \theta}\) for each parameter \(\theta\), we repeatedly use the \textbf{chain rule} of calculus:

\[
\frac{\partial y}{\partial x} 
= \frac{\partial y}{\partial u}\cdot \frac{\partial u}{\partial x},
\]
for a function \(y(u(x))\). In a deep network, we have many compositions: \(\mathbf{h}^{(1)}= f_1(\mathbf{x})\), \(\mathbf{h}^{(2)}= f_2(\mathbf{h}^{(1)})\), \(\dots\). The chain rule tracks how changes in early layers propagate through subsequent layers to alter the final loss.

\subsection{Backpropagation Step-by-Step}

The \emph{backpropagation} algorithm, developed by Rumelhart et al.\ \cite{Rumelhart1986}, efficiently applies the chain rule in layered architectures:
\begin{enumerate}
    \item \textbf{Forward pass:} Compute the output of each layer, culminating in the final loss \(L\).
    \item \textbf{Backward pass initialization:} Start from the derivative of \(L\) w.r.t.\ the final output (often \(\frac{\partial L}{\partial \mathbf{h}^{(n)}}\) for the \(n\)-th layer’s output). Since \(L\) w.r.t.\ itself is 1, we typically use the partial derivative of the loss with respect to the output layer’s pre-activation or post-activation.
    \item \textbf{Layer-by-layer backprop:} Moving from layer \(n\) down to layer \(1\):
    \begin{itemize}
        \item Compute how changes in the layer’s activations affect the loss. This yields \(\frac{\partial L}{\partial \mathbf{h}^{(n)}}\).
        \item Use the chain rule to find \(\frac{\partial L}{\partial \theta_n}\) (the partial derivatives of the loss w.r.t.\ the layer’s parameters \(\theta_n\)).
        \item Pass the gradient \(\frac{\partial L}{\partial \mathbf{h}^{(n-1)}}\) further down to the previous layer.
    \end{itemize}
    \item \textbf{Parameter update:} Use the accumulated gradients \(\frac{\partial L}{\partial \theta_i}\) with whichever gradient descent variant you prefer (SGD, Adam, etc.).
\end{enumerate}

\paragraph{Why is Backprop so Efficient?}
If you tried to compute each partial derivative from scratch for millions of parameters, you would do a lot of redundant work. Backprop reuses intermediate gradients in a systematic way, letting you compute \textit{all} needed partial derivatives in about the same order of time as a few forward passes. 

\subsection{A Simple Example}
Imagine a tiny “network” (just an arithmetic expression):
\[
z = (x + y)\,w.
\]
We want \(\frac{\partial z}{\partial x}\), \(\frac{\partial z}{\partial y}\), and \(\frac{\partial z}{\partial w}\). The forward pass is:
\[
u = x + y \quad (\text{addition step}), 
\quad z = u \times w \quad (\text{multiplication step}).
\]
The backward pass uses the chain rule:
\[
\frac{\partial z}{\partial w} = u, \quad 
\frac{\partial z}{\partial u} = w, 
\quad \text{and} \quad 
\frac{\partial u}{\partial x} = 1,\; \frac{\partial u}{\partial y} = 1.
\]
Hence:
\[
\frac{\partial z}{\partial x} = \frac{\partial z}{\partial u}\cdot \frac{\partial u}{\partial x} 
= w \times 1 = w,
\]
\[
\frac{\partial z}{\partial y} = w,\quad
\frac{\partial z}{\partial w} = x + y.
\]
In a real neural network, each “addition” or “multiplication” is replaced by more complex operations, but the principle is identical.

\section{Challenges in Training Deep Networks}
\subsection{The Vanishing Gradient Problem}
When you have a \emph{deep} neural network—say, 20 or more layers—early layers often suffer from \textbf{vanishing gradients}. Because the chain rule repeatedly multiplies derivatives from each layer, if each derivative is less than 1, the product can shrink exponentially. By the time you get to the first few layers, the gradient is nearly zero, so those layers barely learn.

\subsubsection{Example of Vanishing}
If each layer’s typical derivative magnitude is 0.9, and you have 30 layers, the gradient shrinks roughly like \(0.9^{30} \approx 0.042\). That is less than 5\% of its original magnitude, severely slowing or halting meaningful learning in early layers.

\paragraph{Common Solutions}
\begin{itemize}
    \item \textbf{ReLU-type Activations:} ReLU activations do not saturate in the positive domain, so their derivative is 1 for positive inputs. This helps keep gradients from systematically decaying across layers.
    \item \textbf{Weight Initialization:} Careful initialization ensures that signals neither explode nor vanish at the start of training. Methods like Xavier (Glorot) or He initialization set the variance of weights to keep the standard deviation of activations stable.
    \item \textbf{Normalization Layers:} Techniques such as Batch Normalization or Layer Normalization rescale activations so that each layer’s outputs have consistent means and variances, preventing extreme values and helping maintain stable gradients.
\end{itemize}

\subsection{When Adding Layers Degrades Performance}
Even after mitigating vanishing gradients, researchers noticed that going from, say, a 20-layer model to a 50-layer model often failed to improve results and could \emph{degrade} both training and test accuracy. This phenomenon suggested that simply stacking more layers was not beneficial unless we changed the network architecture to allow better flow of information and gradients. 

\section{Skip Connections and Residual Networks}

A landmark improvement came with \textbf{skip connections}, popularized by \textbf{ResNet} (Residual Network) architectures \cite{He2016}. Instead of forcing every layer’s output to flow \textit{only} through the next layer, skip connections add a direct path that bypasses one or more layers. Mathematically, in a simple “residual block,” the input \(\mathbf{x}\) is combined with a learned transform \(F(\mathbf{x})\):
\[
\mathbf{y} = \mathbf{x} + F(\mathbf{x}).
\]
Here, \(F(\mathbf{x})\) might consist of a few convolutional layers, batch normalization, and ReLU activations. Meanwhile, \(\mathbf{x}\) is passed directly to the output and added on—hence the name “skip” or “shortcut.”

\subsection{Why Skip Connections Help}
\begin{itemize}
    \item \textbf{Direct Gradient Flow.} Because \(\mathbf{y}\) depends on \(\mathbf{x}\) by direct addition, any gradient from \(\mathbf{y}\) back to \(\mathbf{x}\) does not need to be multiplied by many small factors in the chain rule. This significantly alleviates vanishing gradients in very deep networks.
    \item \textbf{Identity Mapping is Easy.} If a certain layer does not need to change the input (i.e., the best transform is the identity), it is straightforward for \(F(\mathbf{x})\) to learn zero, letting \(\mathbf{y} = \mathbf{x}\). In ordinary networks, learning the identity through multiple weighted layers is more complicated.
    \item \textbf{Empirical Success.} ResNets with 50, 101, and 152 layers became feasible to train and outperformed shallower networks in ImageNet recognition tasks \cite{He2016}. This “deeper is better” approach unleashed a new wave of highly deep architectures in both vision and language models.
\end{itemize}

\subsection{Beyond ResNet}
Since ResNet’s introduction, skip connections have appeared in many forms:
\begin{itemize}
    \item \textbf{DenseNet.} DenseNets connect each layer to every other layer, further enhancing gradient flow.
    \item \textbf{Transformers.} Popular in natural language processing, Transformers also incorporate skip connections (plus normalization) in each attention and feed-forward sub-layer.
    \item \textbf{U-Nets.} In image segmentation, skip connections link downsampling layers to corresponding upsampling layers to preserve spatial detail.
\end{itemize}
In all these cases, the principle remains: letting information “skip” or bypass certain transformations can greatly ease the optimization difficulties of deep networks.

\section{Putting It All Together: Practical Training Steps}
Bringing the concepts together, training a typical deep neural network involves the following:

\begin{enumerate}
    \item \textbf{Specify the architecture:} Number of layers, type of layers (convolutional, fully connected, residual blocks, etc.), activation functions (ReLU, etc.), and whether to include skip connections.
    \item \textbf{Choose a loss function:} For classification tasks, \emph{cross-entropy} is standard. This aligns well with probability-based interpretations.
    \item \textbf{Initialize parameters:} Use a method like He or Xavier initialization to avoid extreme initial values that lead to fast vanishing or exploding gradients.
    \item \textbf{Select an optimizer:} Often Adam or mini-batch SGD with momentum. Tune the learning rate or use a schedule that reduces it over epochs.
    \item \textbf{Forward pass:} Feed a batch of training examples through the network, compute outputs, and calculate the loss.
    \item \textbf{Backward pass (backpropagation):} Compute gradients of the loss w.r.t.\ each parameter.
    \item \textbf{Parameter update (gradient descent):} Adjust weights and biases according to the chosen optimizer.
    \item \textbf{Repeat for many epochs:} Over multiple passes (\emph{epochs}) through the dataset, the loss typically decreases, and model accuracy improves. Monitor validation data to detect overfitting or underfitting.
\end{enumerate}

\paragraph{Note on Overfitting.}
If your model fits the training set extremely well but fails on validation or test data, you are overfitting. Techniques like \emph{dropout}, \emph{weight decay}, or \emph{data augmentation} are then used to regularize training.

\section{Extended Explanations of Tricky Concepts}

\subsection{Local Minima, Saddle Points, and High-Dimensional Landscapes}

Newcomers often worry that gradient descent might get stuck in “bad” local minima. In low-dimensional problems (like a 2D function), local minima can be a big issue. But neural networks are typically very high-dimensional (often millions of parameters). In such huge spaces, local minima that are truly “bad” everywhere are statistically rare. More common are \emph{flat or saddle-like regions} where gradients are very small, causing slow progress. 

Practically, modern gradient-based optimizers still find solutions that generalize well, even if they are not \emph{global} minima. Researchers have shown that many local minima yield similarly good performance. Consequently, while local minima and saddle points exist, they do not typically ruin training the way one might initially fear. Tuning hyperparameters (e.g., learning rate, batch size) often has a more direct impact on final performance than stressing over local vs. global minima.

\subsection{Chain Rule and Partial Derivatives: Why It's Intuitive}
The \textbf{chain rule} might feel abstract at first. Imagine a long assembly line of small transformations from your input \(\mathbf{x}\) to the final output \(\mathbf{y}\). Each part of the assembly line either amplifies, shrinks, or reshapes signals. When you see an error at the end, you can trace back to see how each stage contributed: 
\[
\frac{\partial \mathbf{y}}{\partial \mathbf{x}} 
= \frac{\partial \mathbf{y}}{\partial \mathbf{z}_n} \cdot \frac{\partial \mathbf{z}_n}{\partial \mathbf{z}_{n-1}} \cdots \frac{\partial \mathbf{z}_1}{\partial \mathbf{x}},
\]
where \(\mathbf{z}_i\) are intermediate variables. This process systematically breaks down the effect of each transformation (each layer). That’s all the chain rule is—an organized approach to keep track of \emph{cause-and-effect} at every step.

\subsection{Vanishing Gradients: A Numerical Example}
To illustrate how vanishing gradients can occur, imagine a fully connected network with 10 layers, each with a typical derivative magnitude of 0.8. If you try to pass a gradient backward:
\[
0.8^{10} = 0.8^{10} \approx 0.107.
\]
Only around 10\% of the original signal remains. Increase layers to 30, and you end up with about \(0.8^{30} \approx 0.015\). That is barely 1.5\% of the original gradient. Those earliest layers do not get enough signal to update effectively. In practice, factors can be even smaller, leading to near-total \emph{vanishing}. 

\subsection{Skip Connections in More Depth}
In a ResNet block, \(\mathbf{x} \to F(\mathbf{x})\) might represent a small “sub-network” of a few layers. Without a skip connection, you would rely on the chain rule through these multiple layers. If each is \(\alpha < 1\) in derivative magnitude, after two or three layers you might get \(\alpha^3\) scaling. However, with the skip connection
\[
\mathbf{y} = \mathbf{x} + F(\mathbf{x}),
\]
the gradient from \(\mathbf{y}\) back to \(\mathbf{x}\) has a direct path with derivative 1, circumventing the repeated multiplication by \(\alpha\). This architecture ensures deeper networks (such as 50 or 100 layers) can still learn effectively because early layers receive strong gradient signals. 

\section{Conclusion and Next Steps}
\subsection{Chapter Recap}
We have explored:
\begin{itemize}
    \item \textbf{Loss Functions, Probability, and Cross-Entropy.} Cross-entropy emerges naturally from ideas in information theory and maximum likelihood estimation. It is ideal for classification because it is differentiable and strongly encourages correct, confident predictions.
    \item \textbf{Gradient Descent and Backpropagation.} Gradient descent is the workhorse for optimizing neural networks. Backpropagation applies the chain rule to efficiently compute all partial derivatives, even in large, deep networks.
    \item \textbf{Challenges in Deep Networks.} Vanishing gradients can hamper learning in early layers. Merely adding more layers sometimes degrades performance without additional strategies.
    \item \textbf{Skip Connections (ResNets).} By introducing shortcuts, skip connections ease gradient flow, enabling training of very deep networks and often improving accuracy.
\end{itemize}

Understanding how neural networks train, why gradient descent works, and how skip connections solve deep-network challenges is foundational. Whether you aim to design new architectures or simply train existing ones effectively, these concepts will guide you in debugging training issues, improving performance, and appreciating the theory behind deep learning breakthroughs.

\chapter{Artificial Neural Networks}

\section{Introduction}

Artificial Neural Networks (ANNs) are computational models inspired by the neural connections in the human brain. They serve as a powerful approach to approximating a wide range of functions, which has led to their success in tasks such as image recognition, language understanding, speech processing, and more. Over the past decade, improvements in hardware and algorithms have brought deep learning to the forefront of machine learning research and real-world applications. 

In these expanded lecture notes, we will begin by recalling the concept of linear regression and see why a neural network without activation functions behaves similarly to a single linear model. We will then explore how \emph{activation functions} transform these stacked linear layers into highly expressive models capable of modeling non-linear relationships. 

Moving forward, we will introduce \emph{Multi-Layer Perceptrons (MLPs)} and discuss how the parameter count can explode when dealing with high-dimensional data. This naturally motivates \emph{Convolutional Neural Networks (CNNs)}, which reduce the number of parameters by sharing weights in local regions of the input. We will also turn to \emph{Recurrent Neural Networks (RNNs)}, which maintain a hidden state across time for sequential data, before exploring \emph{Transformers} that rely on attention mechanisms for parallelizable, long-range sequence modeling. Finally, we will see how modern deep learning techniques handle data-scarce situations through \emph{semi-supervised} and \emph{self-supervised} learning, culminating in the introduction of \emph{Vision Transformers (ViT)} for image tasks.

These notes aim to be accessible to undergraduate students and beginners with no prior background in AI or advanced mathematics. Wherever possible, we will focus on intuitive explanations that build a solid foundation for further study.

\section{From Linear Regression to Neural Networks Without Activations}

\subsection{Linear Regression in a Nutshell}
Linear regression is one of the simplest yet most foundational models in machine learning. Suppose you have an input vector $\mathbf{x} = (x_1, x_2, \dots, x_n)$ that you want to map to an output $y$ (which can be a real number). The core assumption of linear regression is that $y$ is approximately a \emph{linear} function of the inputs:
\[
    y \approx w_1 x_1 + w_2 x_2 + \dots + w_n x_n + b,
\]
where $w_1, w_2, \dots, w_n$ are weights (or parameters) that determine the importance of each input feature, and $b$ is a bias term that shifts the output. The learning process in linear regression is about finding suitable values for these parameters based on training data.

Though linear regression is intuitive and often effective for simple tasks, it can only model \emph{linear} relationships. Real-world data often exhibit more complexity, so researchers looked to artificial neural networks as a way to capture richer relationships.

\subsection{Stacks of Linear Layers Without Activations}
In principle, one might try to make a linear model more expressive by stacking multiple linear layers. For example, you could have:
\[
    \mathbf{z}_1 = W_1 \mathbf{x} + \mathbf{b}_1, \quad
    \mathbf{z}_2 = W_2 \mathbf{z}_1 + \mathbf{b}_2, \quad \dots
\]
However, if you never apply any \emph{non-linear} function (commonly called an \emph{activation function}) between these layers, you are effectively just applying a sequence of linear transformations. A composition of purely linear transformations is still equivalent to \emph{one} linear transformation. Mathematically, $W_2(W_1 \mathbf{x} + \mathbf{b}_1) + \mathbf{b}_2$ is just another linear mapping from $\mathbf{x}$ to the output. This means that no matter how many such layers you stack, you never increase the true modeling capacity beyond that of a single linear regression model \cite{DeepLearningBook}.

This observation underscores the reason why \emph{activation functions} are crucial to the power of neural networks. By introducing non-linearity, activation functions enable the network to learn more complex, nonlinear relationships in the data.

\section{Activation Functions}

\subsection{Why Non-Linearity Matters}
Activation functions are placed after each linear operation in a neural network. They transform a linear combination of inputs into a nonlinear output, allowing the network to capture curved or more intricate decision boundaries. In many real-world tasks, relationships between inputs and outputs are far from linear. For instance, determining whether an image contains a cat is not a simple matter of summing pixel intensities. Non-linear activation functions allow networks to represent such complex decision boundaries effectively \cite{HahnloserReLU}.

Consider an example from image recognition. Even something as simple as classifying handwritten digits requires distinguishing curved shapes and angles. A linear model would struggle if the variations cannot be separated by a single plane in a high-dimensional space. By using a non-linear activation (like ReLU) after each layer, the network can combine simple patterns in complex ways, effectively carving multiple non-linear decision boundaries.

\subsection{Types of Activation Functions}
\begin{itemize}
    \item \textbf{Sigmoid}: Outputs values between 0 and 1. It was historically popular but tends to saturate (output very close to 0 or 1) for large magnitude inputs. This saturation makes gradients very small (vanishing gradient problem) and can slow training.
    
    \item \textbf{Tanh}: Similar “S”-shaped curve but outputs in the range (-1, 1). This can help center the data around zero. However, it still saturates at large positive or negative inputs.
    
    \item \textbf{ReLU (Rectified Linear Unit)}: Defined as $\max(0, x)$. This is piecewise linear: it outputs 0 for negative $x$ and $x$ itself for positive $x$. ReLU is computationally efficient and has a gradient of 1 for all positive inputs, which reduces the vanishing gradient issue \cite{HahnloserReLU}. It has become the default choice for many hidden layers in modern deep networks.
    
    \item \textbf{Leaky ReLU, ELU, Swish, etc.}: Variations of ReLU exist to fix minor issues. For instance, Leaky ReLU assigns a small negative slope (like 0.01) instead of 0 for negative inputs. ELU and Swish offer smoother transitions, potentially improving convergence.
    
    \item \textbf{Softmax}: Mainly used in the output layer for multi-class classification. It turns a set of raw scores (logits) into probabilities that sum to 1 across all classes.
\end{itemize}

Choosing the right activation can affect both convergence speed and final performance, but the common thread is that \emph{any} non-linear activation helps a network escape the limitations of purely linear models.

\section{Multi-Layer Perceptrons (MLPs) and Parameter Growth}

\subsection{Fully-Connected Layers}
A Multi-Layer Perceptron (MLP) is built from layers of neurons, where each neuron in one layer is connected to every neuron in the next layer. These layers are called \emph{fully-connected} or \emph{dense} layers. When we say a network has $L$ layers, we typically count only the hidden layers plus an output layer (with an input layer preceding them to feed data in).

In an MLP, each neuron computes a weighted sum of inputs plus a bias term. After computing this sum, the neuron applies an activation function. Hence, a single hidden layer with $H$ neurons (ignoring bias terms) has $H \times d$ weights if $d$ is the number of input features.

\subsection{Explosive Parameter Counts in High Dimensions}
While MLPs are conceptually straightforward, a major downside appears when the input dimension is large. For example, consider a color image of size $224 \times 224$ pixels (around 150k pixels), with three color channels. If we feed these pixels directly into a single hidden layer of 1000 neurons, that first layer alone has about $150,000 \times 1,000$ weights, which is 150 million parameters, plus 1000 biases. Even with modern hardware, this is large and can lead to overfitting or slow training \cite{AlexNet}.

Furthermore, images exhibit spatial structure (nearby pixels often relate to each other) that fully-connected layers do not exploit. This inefficiency in parameter usage is a key motivation for more specialized architectures. Still, MLPs remain relevant for \emph{structured data} or lower-dimensional data where this explosive growth is less severe.

\subsection{General Approximation Power vs. Practicality}
In theory, MLPs can approximate a wide range of functions. This is often quoted as the \emph{universal approximation theorem}, stating that a sufficiently large single hidden layer network can approximate any continuous function under mild conditions. However, practicality matters. A naive MLP approach to large-scale tasks results in impractically large parameter counts and high computational costs.

Hence, while MLPs are often the first deep network many students encounter, most modern architectures for images or sequences build upon the insight that we can share or reuse parameters more efficiently. Thus, we turn to Convolutional Neural Networks (CNNs) for image tasks.

\section{Convolutional Neural Networks (CNNs)}

\subsection{Motivation for CNNs}
Traditional MLPs treat each input feature independently. For images, this means ignoring that pixels near each other are often highly correlated. Convolutional Neural Networks (CNNs) address this by:
\begin{itemize}
    \item Using \emph{local connectivity}, where each neuron focuses on a small \emph{patch} of the image (e.g., a 3 by 3 region).
    \item Sharing the same filter weights (known as \emph{parameter sharing}) across different spatial locations. 
\end{itemize}
This approach dramatically reduces the number of parameters while also leveraging the fact that certain features (like edges) can appear anywhere in an image.

\subsection{Core Operations: Convolution and Feature Maps}
At the heart of a CNN layer is the \emph{convolution operation}, which involves a small filter (kernel) matrix. For a $3 \times 3$ filter, you slide it over the input image. At each position, you compute the dot product between the filter and the underlying region of the image, obtaining a single number for that position. Doing this across all spatial positions yields a 2D \emph{feature map} \cite{LeCunGradient}.

By using multiple filters, the network can detect different visual features. For example, one filter may become an “edge detector,” another a “corner detector,” and so on, and these are \emph{learned} from data rather than hand-designed. Sharing the same filter across the entire image drastically cuts the parameter count compared to an MLP with equal coverage.

\subsection{Stride, Padding, and Pooling}

\paragraph{Stride}
The \emph{stride} tells us how many pixels to skip each time we move the filter. A stride of 1 means shifting the filter one pixel at a time, covering every possible location. A stride of 2 or more reduces overlap between filter applications, decreasing the spatial size of the resulting feature maps. Larger strides can speed up computation but at the cost of losing some fine detail \cite{AlexNet}.

\paragraph{Padding}
When the filter approaches the edges of an image, part of the filter would lie outside the image if we did not add \emph{padding}. A common strategy is zero-padding, where extra rows and columns of zeros are added around the image. This influences the output size and ensures that the filter can be applied at the boundary. Without padding, the output shrinks after each convolution, which might be undesirable in some designs.

\paragraph{Pooling}
\emph{Pooling} layers reduce the spatial resolution of feature maps by aggregating values in small non-overlapping regions (e.g., $2 \times 2$). The most common is \emph{max pooling}, taking the maximum in each region, though average pooling is also used. Pooling helps in two ways: it diminishes the computational burden for deeper layers by shrinking the representation, and it achieves a degree of translational invariance. Small shifts in the input image result in less dramatic changes in the pooled representation.

\subsection{Receptive Field and Deep CNNs}
In a deep CNN, the \emph{receptive field} of neurons in higher layers grows. A neuron in the second or third convolutional layer indirectly sees a larger patch of the original image, because each layer stacks on top of previous feature maps. Early layers detect low-level features (edges, corners) while deeper layers combine them into higher-level structures (textures, object parts). Ultimately, the final layers can represent entire objects \cite{ResNet}.

CNNs revolutionized computer vision. Since the groundbreaking success of AlexNet on the ImageNet competition in 2012, CNN architectures such as VGG, ResNet, and others have repeatedly raised the bar on image classification, detection, and segmentation tasks \cite{AlexNet, ResNet}.

\section{Recurrent Neural Networks (RNNs)}

\subsection{Sequential Data and the Need for Memory}
Not all data is spatial like images. Many tasks involve \emph{sequences}: text, speech signals, time series data from sensors, etc. In such cases, the order in which data points appear is meaningful. We might want to process a sentence word by word, or a speech waveform frame by frame. Traditional feed-forward networks or CNNs are less suited to capturing these temporal dependencies.

\textbf{Recurrent Neural Networks (RNNs)} maintain a \emph{hidden state} $h_t$ that is passed from one time step $t$ to the next, effectively giving the network a form of \emph{memory}. At time $t$, an RNN takes both the current input $x_t$ and the previous hidden state $h_{t-1}$, combining them to produce $h_t$. This hidden state can store information from earlier time steps \cite{GRU}.

\subsection{Parameter Sharing Over Time}
One of the strengths of RNNs is that the same set of weights is used at each time step, regardless of the sequence length. This “weight recycling” means the model can handle inputs of variable length without needing an exponential increase in parameters. It also captures patterns regardless of their position in the sequence, allowing the network to generalize across different parts of the input \cite{GRU}.

\subsection{Backpropagation Through Time (BPTT)}
To train an RNN, we “unroll” it over the sequence. For example, a sequence of length $T$ is turned into $T$ layers in a computational graph, each layer sharing weights. Then we apply \emph{backpropagation} on this unrolled graph, a process called \emph{Backpropagation Through Time (BPTT)}. 

However, when $T$ is large, the gradient must backpropagate through many time steps. This can lead to \emph{vanishing gradients} (where updates become extremely small) or \emph{exploding gradients} (where they become excessively large). Gated architectures like LSTM (Long Short-Term Memory) and GRU (Gated Recurrent Unit) help mitigate vanishing gradients by introducing gating mechanisms that allow selective forgetting or retention of information. These architectures have propelled RNNs to success in language modeling, machine translation, and other sequential tasks.

\section{Transformers and Self-Attention}

\subsection{Limitations of RNNs and the Rise of Transformers}
Although RNNs handle sequences, they do so one step at a time, which is difficult to parallelize. They also sometimes struggle with very long-range dependencies, where relevant information might be hundreds of steps away. 

\textbf{Transformers} address these issues by discarding recurrence in favor of an \emph{attention mechanism} that lets every position in a sequence directly attend to every other position. This parallelizable approach allows for more efficient computation on modern hardware. Originally introduced by Vaswani et al. \cite{VaswaniAttention} for machine translation, Transformers have since become a standard architecture in natural language processing (NLP).

\subsection{Self-Attention Mechanism}
In a Transformer, each element of the sequence (for instance, each word in a sentence) is transformed into three vectors: a \emph{Query}, a \emph{Key}, and a \emph{Value}. For each query, we compute scores with every key to gauge how much attention the model should pay to that key’s corresponding value. These scores are normalized via a softmax, resulting in attention weights. The output is then a weighted sum of the value vectors. 

This is often referred to as \emph{Scaled Dot-Product Attention}:
\[
\text{Attention}(Q, K, V) = \text{softmax}\bigl(\frac{Q K^\top}{\sqrt{d_k}}\bigr) \, V,
\]
where $d_k$ is the dimensionality of keys. Because every element can attend to every other element in parallel, Transformers can capture long-range relationships in a single layer, rather than through multiple recurrent steps \cite{VaswaniAttention}.

\subsection{Parallelization and Multi-Head Attention}
Another advantage of Transformers is parallelization. In RNNs, time steps must be processed sequentially, but in a Transformer layer, we can process all positions simultaneously because the attention scores are computed pairwise between all positions at once.

To enhance the learning capacity, \emph{multi-head attention} splits the Query, Key, and Value vectors into multiple “heads,” each with its own parameters. Each head performs attention independently, then the outputs are concatenated. This allows the model to attend to different kinds of information within the same layer. For instance, one head might focus on local context while another identifies global patterns.

\section{Semi-Supervised Learning}

\subsection{Motivation and Basic Ideas}
Real-world datasets frequently have limited labels, because labeling data is often time-consuming, expensive, or requires domain expertise. However, we may have a large pool of unlabeled data. \textbf{Semi-supervised learning} attempts to leverage both labeled and unlabeled data to improve model performance \cite{LeePseudo}.

\subsection{Entropy Minimization}
One simple idea is \emph{entropy minimization}, which encourages the model to produce confident (low-entropy) predictions on unlabeled data. Suppose the network’s output layer is a softmax distribution over classes. If the distribution for an unlabeled sample is uniform or uncertain, the entropy is high. By penalizing high entropy, the model is nudged to produce sharper, low-entropy predictions \emph{without} explicit labels. This technique can push the decision boundary into low-density regions of the data space.

\subsection{Pseudo-Labeling}
Another popular method is \emph{pseudo-labeling} or self-training \cite{LeePseudo}. The model uses its own high-confidence predictions on unlabeled samples as if they were ground-truth labels. Then, these pseudo-labeled samples are added to the training set. This simple approach can be surprisingly effective, though it requires care, because if the model is confident yet wrong, it can reinforce its mistakes. Researchers often combine pseudo-labeling with confidence thresholds or repeated refinement to mitigate error propagation.

By combining these ideas with strong data augmentation and consistency constraints, modern semi-supervised methods narrow the gap between fully supervised training and training with limited labeled data.

\section{Self-Supervised Learning}

\subsection{Overview and Motivation}
\textbf{Self-supervised learning} goes a step further by completely bypassing human-provided labels. Instead, the model constructs a surrogate or “pretext” task from the data itself. This approach opens the door to using massive amounts of unlabeled data. After solving the pretext task, the resulting model typically learns useful representations that can be fine-tuned for downstream tasks \cite{SimCLR}.

Examples include:
\begin{itemize}
    \item Masking parts of the input and asking the model to reconstruct them.
    \item Predicting one part of the data from another (e.g., future frames in a video from past frames).
    \item Distinguishing augmented views of the same sample from views of different samples (contrastive methods).
\end{itemize}

\subsection{Masked Modeling: BERT}
In natural language processing, \textbf{BERT} \cite{DevlinBERT} popularized \emph{masked language modeling}. About 15 percent of tokens in a sentence are randomly replaced with a special [MASK] symbol (or a placeholder), and the model must predict those hidden words from the context. This forces the model to learn a deep, bidirectional understanding of text. Once pretrained in this self-supervised fashion on large corpora, BERT can be fine-tuned on many NLP tasks with fewer labeled examples.

\subsection{Autoregressive Generation: GPT}
Another self-supervised approach is the \emph{autoregressive} method, used by GPT (Generative Pretrained Transformer) \cite{RadfordGPT}. Here, the model is trained to predict the next token in a sequence given all previous tokens. Because every next token in a large text corpus is a training label, the model naturally learns grammar, facts, and contextual relationships. GPT has shown remarkable capabilities in text generation, question answering, and other tasks.

\subsection{Contrastive Learning}
In computer vision, \emph{contrastive learning} is widely used to learn robust visual features without labels \cite{SimCLR}. The main idea is to take two random augmented views of the same image and push their representations to be similar, while pushing apart representations of different images. By doing so, the model uncovers factors of variation that are invariant to augmentations, leading to strong transferable features for tasks like classification or detection.

\section{Vision Transformers (ViT)}

\subsection{Adapting Transformers to Images}
Though CNNs dominate image-based tasks, there has been growing interest in adapting Transformers to vision. The \textbf{Vision Transformer (ViT)} \cite{DosovitskiyViT} breaks an image into patches, each patch treated like a token in the Transformer pipeline. For instance, a $224 \times 224$ image can be split into $16 \times 16$ patches, leading to $(224/16)^2 = 14^2 = 196$ patches. Each patch is flattened and projected into a vector embedding.

A special “class token” can be prepended, and standard Transformer layers process all these patch embeddings in parallel using multi-head self-attention. This lets each patch “attend” to other patches regardless of spatial distance, allowing the model to capture both local and global patterns in the first layer.

\subsection{Performance Considerations}
Vision Transformers can match or exceed state-of-the-art CNNs on image classification tasks, \emph{given} large-scale training data \cite{DosovitskiyViT}. Unlike CNNs, ViTs do not explicitly encode translational invariance or local connectivity. They rely on self-attention to learn these properties. As a result, ViTs often require very large labeled datasets (or self-supervised pretraining on large unlabeled sets) to reach their full potential. When such data is available, ViTs provide a flexible, attention-based approach that can be easier to scale than very deep CNNs.

Recent work applies ViTs beyond classification: in object detection, segmentation, and even generative modeling. Sometimes, hybrid architectures combine CNN-like local attention with the global attention of Transformers. These trends suggest the gap between CNNs and Transformers in computer vision might keep narrowing as research advances.

\section{Conclusion}

We have traversed a broad landscape of deep learning approaches:

\begin{itemize}
    \item \textbf{Linear Regression to Neural Networks Without Activations:} We observed how a stack of purely linear layers collapses to one linear mapping, making activation functions essential.
    \item \textbf{Activation Functions:} Introduced Sigmoid, Tanh, ReLU, and others. Non-linearities unlock the power of neural networks to learn complex patterns.
    \item \textbf{Multi-Layer Perceptrons (MLPs):} Showed how parameters grow rapidly with input size, motivating more specialized architectures for high-dimensional inputs.
    \item \textbf{Convolutional Neural Networks (CNNs):} Leveraged local connectivity and weight sharing for images, drastically reducing parameters and respecting spatial structure.
    \item \textbf{Recurrent Neural Networks (RNNs):} Provided a way to handle sequential data through hidden states passed over time. BPTT allowed training but brought issues of vanishing gradients, mitigated by LSTM/GRU cells.
    \item \textbf{Transformers:} Replaced recurrence with self-attention, enabling parallel processing of sequences and better handling of long-range dependencies. Became a core architecture in NLP and are expanding to other domains.
    \item \textbf{Semi-Supervised Learning:} Employed entropy minimization and pseudo-labeling to utilize unlabeled data, bridging the gap between supervised and unsupervised settings.
    \item \textbf{Self-Supervised Learning:} Created training signals from data itself, as seen in masked language modeling (BERT), autoregressive modeling (GPT), and contrastive learning (SimCLR).
    \item \textbf{Vision Transformers (ViT):} Extended the Transformer architecture to images by splitting them into patches, enabling global attention in a single layer and achieving competitive performance with CNNs.
\end{itemize}

Taken together, these developments underscore the flexibility and strength of deep learning. The same broad principles – layering transformations with non-linear activations, leveraging shared parameters for efficiency, and learning from data directly – drive the architecture design in various modalities (images, text, time series, etc.). Where labeled data is scarce, semi-supervised and self-supervised approaches continue to expand the reach of AI, enabling large models to learn universal features from unlabeled corpora or images.

As you progress, you may dive deeper into each architecture's detailed mathematics or attempt to implement them from scratch. While new techniques and variations will surely arise, the foundations covered here remain essential to understanding modern deep learning systems. Mastery of these concepts will prepare you to adapt and innovate in the constantly evolving field of AI.

\chapter{History of AI}

\section{Machine Learning vs.\ Deep Learning}
Machine Learning (ML) is a field where algorithms learn patterns from data. Traditional ML relies on humans to select the most relevant features, a process called \emph{feature engineering}. Deep Learning, on the other hand, uses multiple layers of artificial neural networks to automatically learn features directly from raw data, requiring less manual design.

\paragraph{Balancing Bias and Variance}
A well-trained model minimizes both \emph{bias} (systematic errors caused by oversimplified assumptions) and \emph{variance} (over-sensitivity to variations in training data). The ideal balance helps the model generalize to new, unseen data without overfitting.

\section{Information Theory and Cross-Entropy}
Information theory provides mathematical tools to quantify how much uncertainty or ``information'' is present in events or data.

\paragraph{Self-Information and Entropy}
The \emph{self-information} of an event reflects how ``surprising'' it is. The \emph{entropy} of a distribution is the average self-information over all possible outcomes, representing the overall uncertainty in that distribution.

\paragraph{Cross-Entropy in Model Training}
In supervised learning for classification, a common loss function is the cross-entropy:
\[
H(p, q) = -\sum_{x} p(x)\,\log\bigl(q(x)\bigr),
\]
where \(p(x)\) is the true distribution (often represented as one-hot labels) and \(q(x)\) is the model's predicted distribution. Cross-entropy reaches its minimum when \(p(x)\) and \(q(x)\) match exactly.

\section{Gradient Descent}
Gradient descent is an optimization method used to minimize a loss function \(L(\theta)\), where \(\theta\) represents the model parameters. At each iteration:
\[
\theta \leftarrow \theta - \alpha \nabla_{\theta} L(\theta),
\]
where \(\alpha\) is the \emph{learning rate} and \(\nabla_{\theta} L(\theta)\) is the gradient of the loss with respect to the parameters.

\paragraph{Steepest Descent Analogy}
Imagine you are on a hillside and want to reach the bottom by always taking a step in the steepest downward direction. Similarly, in gradient descent, each update step aims to reduce the loss function's value.

\section{Backpropagation and Optimizer}
\paragraph{Backpropagation}
Backpropagation applies the chain rule of calculus to compute the contribution of each parameter to the final loss. It propagates the error from the network's output layer backward to all layers:
\[
\frac{\partial L}{\partial x_i} 
= \frac{\partial L}{\partial x_j}
\cdot \frac{\partial x_j}{\partial x_i},
\]
where \(x_i\) and \(x_j\) are intermediate variables or node outputs along the computational path.

\paragraph{Optimization Methods}
Optimizers such as \emph{SGD with Momentum}, \emph{Nesterov Momentum}, \emph{RMSProp}, or \emph{Adam} add extra terms or adaptive learning rates to stabilize and speed up convergence.

\section{Loss Landscape and Skip-Connection}
\paragraph{Loss Landscape}
A neural network's loss landscape describes how the loss function changes as we move through the space of possible parameter values. Difficult landscapes with many ``cliffs'' or sharp local minima can slow or derail training.

\paragraph{Skip-Connection}
Modern architectures (ResNet, for example) use \emph{skip connections} to let information bypass certain layers. This design often produces smoother loss landscapes and improves gradient flow, making training more stable.

\section{Hopfield Network}
A Hopfield network is an older type of neural system where each unit is connected to every other unit. It uses the principle ``neurons that fire together, wire together'' to store and recall patterns. Once trained, these networks can retrieve complete patterns from partial or noisy inputs through an iterative update process. 

\paragraph{Key Characteristics}
\begin{itemize}
  \item \textbf{Binary Neurons}: Classic Hopfield networks typically use binary (on/off) neurons. Each neuron's state can be represented as $+1$ or $-1$ (or sometimes $0$ or $1$).
  \item \textbf{Symmetric Weights}: The connection weight from neuron $i$ to neuron $j$ equals the weight from $j$ to $i$, i.e.\ $w_{ij} = w_{ji}$. This symmetry ensures that the network's energy function is well-defined.
  \item \textbf{Energy Function}: The network is often described by an energy function
  \[
    E(\mathbf{s}) = -\frac{1}{2} \sum_{i,j} w_{ij} \, s_i \, s_j + \sum_i b_i \, s_i,
  \]
  where $\mathbf{s}$ denotes the states of all neurons, $w_{ij}$ the weight between neurons $i$ and $j$, and $b_i$ the bias term for neuron $i$. The network naturally evolves toward states that minimize this energy.
  \item \textbf{Iterative Update}: Neurons update their states one at a time or in small groups. Each update rule typically involves switching the neuron's state to the sign of the weighted sum of its inputs.
\end{itemize}

\paragraph{Associative Memory}
Hopfield networks can store a set of patterns (e.g.\ binary vectors) by choosing appropriate weights. When a partial or noisy version of a stored pattern is given as initial input, the network iteratively updates its neuron states, eventually settling into the closest stored pattern. This property makes Hopfield nets useful for pattern completion and noise correction.

\paragraph{Applications and Limitations}
\begin{itemize}
  \item \textbf{Pattern Recognition and Completion}: Commonly applied to small, simple memory tasks like recovering corrupted images.
  \item \textbf{Capacity Constraints}: The number of patterns that can be reliably stored is proportional to the number of neurons ($\sim 0.14N$ for binary Hopfield networks).
  \item \textbf{Energy Minimization Perspective}: Introduced the idea that certain neural networks can be viewed as minimizing an energy function, influencing later research on energy-based models.
\end{itemize}

\section{Boltzmann Machine}
A Boltzmann machine generalizes Hopfield networks by adding \emph{hidden units} that are not directly observed. This allows the model to capture more complex structure in data. Boltzmann machines are \emph{stochastic} (probabilistic) networks, meaning the state of each neuron is updated based on a probability rather than a strict sign or threshold rule.

\paragraph{Energy Function and Probability Distribution}
Like the Hopfield network, Boltzmann machines have an energy function:
\[
  E(\mathbf{v}, \mathbf{h}) 
  = -\sum_{i,j} w_{ij}\, v_i\, v_j 
    - \sum_{k,l} w_{kl}\, h_k\, h_l 
    - \sum_{i,k} w_{ik}\, v_i\, h_k
    - \sum_i b_i\, v_i 
    - \sum_k c_k\, h_k,
\]
where $\mathbf{v}$ represents the visible units, $\mathbf{h}$ the hidden units, $w_{ij}$ the connections among visible units, $w_{kl}$ the connections among hidden units, $w_{ik}$ the cross-connections, and $b_i$, $c_k$ the biases for visible and hidden units respectively. The network defines a probability distribution over the states $(\mathbf{v}, \mathbf{h})$ via the Boltzmann distribution:
\[
  P(\mathbf{v}, \mathbf{h}) 
  = \frac{1}{Z} \exp\bigl(-E(\mathbf{v}, \mathbf{h})\bigr),
\]
where $Z$ is the partition function ensuring that all probabilities sum to 1.

\paragraph{Stochastic Learning Process}
\begin{itemize}
  \item \textbf{Gibbs Sampling}: The network typically uses Gibbs sampling to update the state of each neuron based on the states of all other neurons. This involves calculating the probability that a neuron should be $1$ (or $+1$) given the current states of its neighbors, and then randomly deciding its next state according to that probability.
  \item \textbf{Contrastive Divergence (CD)}: A practical algorithm introduced by Geoffrey Hinton for training Restricted Boltzmann Machines (RBMs). RBMs are a simplified version of Boltzmann machines with no hidden-to-hidden or visible-to-visible connections. CD approximates the gradients needed to update the weights and biases.
  \item \textbf{Incremental Improvement}: By iteratively sampling and updating weights to better match the data distribution, the Boltzmann machine converges to a set of parameters that (ideally) models complex relationships between visible variables.
\end{itemize}

\paragraph{Foundations for Deep Architectures}
\begin{itemize}
  \item \textbf{Restricted Boltzmann Machines and Deep Belief Networks (DBNs)}: RBMs serve as building blocks for deeper models. Stacking multiple RBMs leads to \emph{Deep Belief Networks}, one of the early successes in deep learning.
  \item \textbf{Energy-Based View}: Boltzmann machines strengthen the concept that learning can be framed as finding parameter configurations that minimize an energy function, influencing other energy-based models.
  \item \textbf{Limitations}: 
  \begin{itemize}
    \item Full Boltzmann machines can be challenging to train due to slow mixing times in Gibbs sampling and high computational costs.
    \item Architectural constraints (like the restrictions in RBMs) are often used to make training tractable.
  \end{itemize}
\end{itemize}
\section{Activation Functions}
Activation functions bring non-linearity into a neural network. Without them, each layer would just perform a linear transformation of the previous layer, severely limiting what can be learned.

\paragraph{Examples}
\begin{itemize}
  \item \textbf{Sigmoid}: \(\sigma(x) = \frac{1}{1 + e^{-x}}\)
  \item \textbf{tanh}: \(\tanh(x)\)
  \item \textbf{ReLU}: \(\text{ReLU}(x) = \max(0, x)\)
  \item \textbf{Leaky ReLU}: Allows a small gradient for negative \(x\).
  \item \textbf{Maxout} and \textbf{ELU}: Variations that improve gradient flow in certain scenarios.
\end{itemize}

\section{Multi-Layer Perceptron}
A Multi-Layer Perceptron (MLP) fully connects each neuron in one layer to all neurons in the next layer. For large inputs, MLPs can grow to have hundreds of millions or even billions of parameters. Modern language models push this design to extremes, showing that vast capacity can capture a remarkable range of patterns.

\section{Convolutional Neural Network}
\paragraph{Convolutional Layers}
Instead of full connections, convolutional layers use small filters (kernels) that slide across the input. These filters are \emph{shared} across locations, greatly reducing parameters and capturing local structures (e.g., edges in images).

\paragraph{Receptive Field}
In a CNN, the \emph{receptive field} is the region of the input that influences a neuron in the output. Deeper layers combine information from earlier layers, expanding the receptive field. This allows higher layers to detect more complex features.

\section{Recurrent Neural Network}
RNNs handle sequential data (like text or audio) by updating a \emph{hidden state} at each timestep:
\[
h_t = f(h_{t-1}, x_t),
\]
where \(x_t\) is the current input and \(h_{t-1}\) is the previous hidden state. Training RNNs on long sequences can be challenging. One approach to make it more tractable is to truncate the backpropagation so that it does not go all the way back to the earliest timesteps.

\section{Transformer}
Transformers remove the RNN-style recurrence. They rely on \emph{attention}, which compares all pairs of tokens in a sequence simultaneously:
\[
\text{Attention}(Q, K, V) = \text{softmax}\left(\frac{QK^\top}{\sqrt{d}}\right)V,
\]
where \(Q\), \(K\), and \(V\) are query, key, and value matrices respectively, and \(d\) is a scaling factor. This parallel approach speeds up training, especially for long sequences.

\section{Self-Supervised Learning}
Self-supervised learning methods allow models to train on unlabeled data by creating proxy tasks. One such task is \emph{self-prediction}, where certain parts of the data are masked or removed, and the model tries to predict them. This reduces the need for extensive human-generated labels.

\subsection{Autoregressive Generation}
Autoregressive models predict the next element of a sequence from past elements:
\[
p(x) = \prod_{t=1}^{T} p(x_t \mid x_{1}, \ldots, x_{t-1}).
\]
This approach underlies many modern language models, such as GPT, which generate text one token at a time.

\subsection{Generative Pre-Training (GPT)}
\paragraph{Basic Idea}
GPT-style models train on massive text corpora to predict the next token. By seeing large amounts of text, they learn complex language patterns.

\paragraph{In-Context Learning}
Once trained, GPT can adapt to new tasks simply by reading example prompts in the input:
\begin{itemize}
  \item \textbf{Zero-shot}: No examples of the task are given, so GPT must rely on broad knowledge.
  \item \textbf{One-shot}: Only a single example is provided.
  \item \textbf{Few-shot}: A few examples guide GPT to perform tasks more effectively.
\end{itemize}

\subsection{Masked Generation / Prediction}
Unlike autoregressive models, masked language modeling (for example, in BERT) conceals some tokens and trains the model to fill them in. This forces the model to learn context from both sides of the missing tokens.

\subsection{Vision Transformer (ViT)}
Vision Transformers adapt the Transformer architecture to images by splitting them into patches and treating each patch as a token. The attention mechanism then learns how patches relate to one another, enabling the model to classify or interpret an entire image.

\subsection{Contrastive Learning}
Contrastive learning methods train models to distinguish between \emph{similar} and \emph{dissimilar} examples. For instance, two augmentations of the same image are treated as similar, while different images are treated as dissimilar. This approach helps learn robust embeddings without explicit labels.

\section{Evolution of AI}

AI has progressed through several major stages, each marked by key theoretical and practical advances as well as new, larger datasets.

\subsection{Foundational Neural Network Theories (1982 to 2011)}
\begin{itemize}
  \item \textbf{Perceptron and Early Neural Networks}: Researchers like Frank Rosenblatt pioneered the \emph{perceptron} in the late 1950s, focusing on simple binary classifiers. While the perceptron was limited, it sparked interest in computational models of learning.
  \item \textbf{Hopfield Networks (1982)}: John Hopfield introduced a fully connected, recurrent neural network that could serve as a form of associative memory. This laid the groundwork for more research into energy-based models.
  \item \textbf{Boltzmann Machines}: Building on energy-based concepts, Boltzmann Machines introduced hidden units and stochastic (random) sampling methods for learning. This opened the door to modeling more complex data distributions.
  \item \textbf{Backpropagation (1986)}: David Rumelhart, Geoffrey Hinton, and Ronald Williams popularized the backpropagation algorithm, providing a systematic way to compute how each weight contributes to network error. This solved a crucial training bottleneck for multi-layer neural networks.
  \item \textbf{Early Computational Limits}: Despite theoretical advancements, hardware and data constraints meant that neural networks saw limited success until faster processors and bigger datasets became widely available.
\end{itemize}

\subsection{Supervised Learning and Specialized Architectures (2012 to 2016)}
\begin{itemize}
  \item \textbf{Explosion of Labeled Data}: The availability of large, labeled datasets (for example, ImageNet) allowed neural networks to train on real-world problems at a new scale.
  \item \textbf{Deep Convolutional Neural Networks (CNNs)}: In 2012, Alex Krizhevsky and colleagues used CNNs to achieve a groundbreaking result on the ImageNet challenge. CNNs became the standard for image tasks like object recognition, detection, and segmentation.
  \item \textbf{Recurrent Neural Networks (RNNs) and LSTM}: Techniques like Long Short-Term Memory (LSTM) networks gained popularity for sequence-related tasks such as language modeling and speech recognition.
  \item \textbf{Wider Adoption in Industry}: Improved accuracy on benchmarks convinced tech companies to invest heavily in neural networks for speech-to-text, image tagging, and other applications. This drove further research and funding.
\end{itemize}

\subsection{Attention-Based Universal Architectures (2017 to 2024)}
\begin{itemize}
  \item \textbf{Transformer Architecture (2017)}: Proposed by Vaswani et al., the Transformer removed the need for recurrent operations in sequence modeling, relying instead on attention mechanisms. This made training faster and more parallelizable.
  \item \textbf{Scale and Generality}: Larger models, especially in language tasks, demonstrated unprecedented capabilities. Models like BERT (2018), GPT-2 (2019), GPT-3 (2020), and subsequent generations performed well on a wide variety of tasks without specialized architectures for each task type.
  \item \textbf{Multimodality and Cross-Domain Applications}: Research extended Transformers to handle not just text but also images, audio, and combinations of different data types. Vision Transformers (ViT) exemplify applying attention-based methods to image classification.
  \item \textbf{Data-Driven Progress}: Breakthroughs in these years relied on collecting or generating ever-larger training sets. Researchers also explored how to effectively sample and preprocess huge amounts of data from varied sources (web pages, images, videos, etc.).
\end{itemize}

Across each era, \textbf{larger datasets} have consistently enabled bigger models and more advanced techniques, acting as a key driver in the evolution of AI.

\section{Datasets}

Datasets form the foundation for training and evaluating AI models. Below are some major datasets that have propelled research in machine learning and computer vision.

\paragraph{MNIST}
\begin{itemize}
  \item Description: A collection of 70,000 handwritten digits (60,000 for training, 10,000 for testing).
  \item Content: Each image is a 28x28 grayscale picture of a digit from 0 to 9.
  \item Usage: Often the first benchmark for image classification experiments, serving as a standard for testing new algorithms.
\end{itemize}

\paragraph{CIFAR-10 and CIFAR-100}
\begin{itemize}
  \item Description: A set of 60,000 color images (32x32 pixels each).
  \item Classes: CIFAR-10 has 10 classes (for example, airplane, bird, cat) while CIFAR-100 has 100 classes with fewer examples per class.
  \item Usage: Widely used for evaluating image classification models. Although the images are small, the variety of classes introduces more complexity than MNIST.
\end{itemize}

\paragraph{ImageNet}
\begin{itemize}
  \item Description: Contains over a million high-resolution images spanning about 1,000 different object categories.
  \item Significance: The 2012 ImageNet Large Scale Visual Recognition Challenge (ILSVRC) was crucial in popularizing deep learning when CNNs drastically improved classification accuracy.
  \item Impact: Models pre-trained on ImageNet are often used as a starting point for various computer vision tasks, demonstrating transfer learning.
\end{itemize}

\paragraph{MS-COCO (Microsoft Common Objects in Context)}
\begin{itemize}
  \item Description: Over 200,000 images depicting everyday scenes.
  \item Detailed Annotations: Includes object bounding boxes, segmentation masks, and textual captions describing the scene.
  \item Usage: Enables a wide range of tasks including object detection, instance segmentation, and image captioning. Considered more challenging than ImageNet because of cluttered backgrounds and multiple objects per image.
\end{itemize}

\paragraph{OpenImages}
\begin{itemize}
  \item Description: Millions of images with rich annotations (object bounding boxes, segmentation masks, and labels).
  \item Variety: Covers diverse categories and object types with multiple annotations per image.
  \item Usage: Encourages research in object detection, instance segmentation, and other computer vision tasks at scale.
\end{itemize}

\paragraph{Common Crawl}
\begin{itemize}
  \item Description: A massive collection of raw web data, continually updated from billions of webpages.
  \item Relevance to Language Models: Used in large-scale text pre-training for models like GPT. Contains diverse, real-world text samples.
  \item Size: Petabytes of data, making it invaluable for training state-of-the-art natural language processing systems.
\end{itemize}

\paragraph{LAION}
\begin{itemize}
  \item Description: A publicly available dataset of millions of image-text pairs.
  \item Purpose: Aims to advance multimodal research, enabling tasks where both visual and textual understanding are crucial.
  \item Use Cases: Training large-scale models that can associate images with their captions. Models trained on LAION data can generate text descriptions for images, perform image-based question answering, and more.
\end{itemize}

These datasets each address different needs in AI development. Whether it's benchmarking simple classification (MNIST), tackling complex image tasks (ImageNet, MS-COCO), or training vast language models (Common Crawl), they have collectively shaped the progress and capabilities of modern AI systems.

\section{Reinforcement Learning from Human Feedback (RLHF)}
\paragraph{Core Idea}
Reinforcement Learning from Human Feedback (RLHF) is a process that incorporates human judgments or preferences into the training loop of a model. Rather than relying on numerical reward signals from an environment (as in traditional Reinforcement Learning), RLHF uses data collected directly from human evaluators who compare different model outputs and indicate which one is better.

\paragraph{Why RLHF is Important}
\begin{itemize}
  \item \textbf{Alignment with user goals}: Helps ensure that model responses match users' needs or instructions.
  \item \textbf{Reduces unintended behavior}: The model can be guided away from producing harmful or irrelevant outputs.
  \item \textbf{Human-centric approach}: Emphasizes human values, preferences, and oversight when shaping AI behavior.
\end{itemize}

\paragraph{Training Steps}
\begin{enumerate}
  \item \textbf{Supervised Fine-Tuning}:
  \begin{itemize}
    \item The model starts with a base pre-trained checkpoint (for instance, GPT).
    \item It is then fine-tuned on carefully curated, human-written examples of the desired output style or instructions.
    \item This step teaches the model basic instruction-following skills before introducing preference comparisons.
  \end{itemize}

  \item \textbf{Reward Model Training}:
  \begin{itemize}
    \item Human evaluators compare two or more candidate responses for the same input and select which response they prefer.
    \item These comparisons form training data for the \emph{reward model}, which outputs a numerical score indicating how well a response aligns with human preference.
    \item The reward model is trained to predict these human preferences from the text of the response.
  \end{itemize}

  \item \textbf{Reinforcement Learning with the Reward Model}:
  \begin{itemize}
    \item Using the reward model as a proxy for human approval, the system optimizes the policy (i.e., the original language model) to maximize the reward score.
    \item This is often done with methods such as Proximal Policy Optimization (PPO), although other RL algorithms can be used.
    \item During this process, the model learns to produce outputs that humans find more acceptable or useful.
  \end{itemize}
\end{enumerate}

\paragraph{Reward Model}
The reward model is central to RLHF. It converts human comparisons into a scalar feedback signal. By looking at multiple model outputs for the same prompt:
\begin{itemize}
  \item Human evaluators label which output they like more.
  \item The reward model then learns to predict these preference labels.
  \item After training, the reward model can be applied to new model outputs to assign scores indicating how closely they match human desires.
\end{itemize}

\section{Red Teaming}
Red Teaming is a separate process that tests the limits and vulnerabilities of a trained model. Here, ``red teamers'' or testers try to force the model to produce problematic or unintended outputs through adversarial prompts or tricky scenarios.

\paragraph{Goals of Red Teaming}
\begin{itemize}
  \item \textbf{Identify weaknesses}: Expose scenarios where the model fails, misbehaves, or produces harmful content.
  \item \textbf{Improve safety and robustness}: Insights from red teaming guide further training, model revisions, or prompt engineering.
  \item \textbf{Promote responsible deployment}: Understanding potential failure modes helps in setting usage policies and mitigations.
\end{itemize}

\paragraph{Examples of Red Teaming Strategies}
\begin{itemize}
  \item \textbf{Provocative questions}: Asking offensive or controversial questions to test model responses.
  \item \textbf{Ambiguous phrasing}: Seeing how the model handles unclear or incomplete queries.
  \item \textbf{Trick prompts}: Using unusual wording, code-like text, or hidden cues to see if the model can be manipulated.
\end{itemize}

\section{Chain-of-Thought (CoT)}
Chain-of-Thought (CoT) is a technique where a model breaks down a complex question or task into multiple steps, effectively mimicking how a human might outline or reason through a problem. While the final answer is still a single output, the intermediate reasoning steps can help the model arrive at more accurate or logical conclusions.

\paragraph{Key Benefits}
\begin{itemize}
  \item \textbf{Enhanced reasoning}: By sequentially explaining its thought process, the model can tackle tasks requiring multiple steps (e.g., multi-stage math problems or logical puzzles).
  \item \textbf{Reduced errors}: Seeing intermediate steps can help catch mistakes earlier, leading to more refined final answers.
  \item \textbf{Better interpretability}: In some cases, CoT outputs reveal how the model arrived at its conclusion, making it easier to diagnose errors or biases.
\end{itemize}

\paragraph{Implementation Approaches}
\begin{itemize}
  \item \textbf{Prompt-based}: The user asks the model to ``explain step-by-step'' or to ``show your reasoning.'' 
  \item \textbf{Training-based}: The model is fine-tuned on examples that already include chain-of-thought style reasoning, encouraging it to learn a structured explanation approach.
\end{itemize}

\section{Self-Instruct}
Self-Instruct is a method where a language model uses a small set of ``seed'' instructions and outputs. It then generates new instructions and potential answers, effectively teaching itself how to handle various requests.

\paragraph{How Self-Instruct Works}
\begin{enumerate}
  \item \textbf{Seed Prompts}: Collect a small number of high-quality tasks and example answers.
  \item \textbf{Generation}: The model creates new tasks by imitating the style of the seed prompts, then attempts to answer them.
  \item \textbf{Filtering \& Refinement}: Self-generated data may be noisy. A filtering step removes low-quality prompts and answers, refining the dataset.
  \item \textbf{Iterative Improvement}: The filtered data can be used to fine-tune the model again, further improving performance.
\end{enumerate}

\paragraph{Advantages}
\begin{itemize}
  \item Reduces reliance on large human-labeled instruction datasets.
  \item Can adapt to new domains or task types without extensive human effort.
  \item Encourages the model to explore a broader space of instructions and responses.
\end{itemize}

\section{Direct Preference Optimization (DPO)}
\paragraph{DPO}
Direct Preference Optimization (DPO) is a more direct alternative to the multi-step approach of RLHF. Instead of building a reward model and then running RL, DPO compares preferred vs. non-preferred responses directly and pushes the model to favor the preferred ones.

\paragraph{Comparison to RLHF}
\begin{itemize}
  \item \textbf{Simplicity}: DPO tries to simplify pipeline complexity by eliminating a separate policy optimization loop.
  \item \textbf{Contrastive Learning Setup}: The model is trained in a contrastive manner. It sees pairs of ``preferred'' and ``less preferred'' outputs, and learns to rank the preferred output higher than the non-preferred one.
  \item \textbf{Potential Trade-offs}: DPO might be easier to implement, but it may not capture all the nuances that a full RL approach could, depending on the task complexity and data availability.
\end{itemize}

\section{Retrieval-Augmented Generation (RAG)}
\paragraph{Core Idea}
In Retrieval-Augmented Generation (RAG), a language model is paired with an external retrieval system or database. When the model receives a prompt, it first looks up relevant documents from an external source and then incorporates that information into its final response.

\paragraph{Motivation}
\begin{itemize}
  \item \textbf{Address Knowledge Gaps}: Large models can contain a lot of information, but they may still be out-of-date or incomplete.
  \item \textbf{Focus on Specific Domains}: RAG allows the model to draw upon a specialized collection of documents or structured data, making the generated content more accurate and detailed.
  \item \textbf{Memory and Parameter Efficiency}: Instead of storing all facts inside the model's parameters, external documents can be used as needed, which is more flexible.
\end{itemize}

\paragraph{Three Approaches in Practice}
\begin{enumerate}
  \item \textbf{Finetuning}:
  \begin{itemize}
    \item The model's internal weights are updated using domain-specific data.
    \item Improves performance on a particular type of task but may reduce generality if too narrowly fine-tuned.
  \end{itemize}

  \item \textbf{Prompt Engineering}:
  \begin{itemize}
    \item Carefully designing how the question or task is presented to the model.
    \item Helps guide the model's attention and reasoning without changing the model's weights.
  \end{itemize}

  \item \textbf{Retrieval-Augmented Generation}:
  \begin{itemize}
    \item The model queries an external source, such as a search index or database, to find relevant context.
    \item The retrieved context is combined with the original prompt to produce a final response.
  \end{itemize}
\end{enumerate}

\section*{Conclusion}
Throughout the history of AI, new algorithms and architectural innovations have often led to significant leaps in capability. This trend has closely tied each breakthrough to the availability of larger or more specialized datasets. By understanding these key developments and techniques, students can appreciate both where AI comes from and where it might go next.

\chapter{Model Scaling}

\section{Introduction to Model Scaling}
Model scaling refers to the practice of increasing a model’s capacity or complexity in order to achieve better performance. In deep learning, scaling typically means using \textbf{larger neural networks}—for example, increasing the number of layers, the number of neurons/filters per layer, or other architectural dimensions. Historically, bigger models trained on more data have often yielded breakthrough improvements in tasks ranging from image recognition to natural language understanding. This trend is sometimes nicknamed the \emph{scaling hypothesis}: given enough data and a large enough model, performance will keep improving (albeit with diminishing returns). Indeed, many milestones in AI have been achieved not just by clever algorithms, but by dramatically scaling up neural networks using more computation and data.

There are multiple ways to scale a neural network model:
\begin{itemize}
\item \textbf{Depth (Layers):} Increase the number of layers in the network, allowing more complex sequential feature transformations.
\item \textbf{Width (Units per layer):} Increase the number of neurons in each layer (for fully-connected networks) or the number of channels/filters (for convolutional networks), allowing more features to be processed in parallel.
\item \textbf{Model Resolution / Input Size:} For vision models, use higher-resolution inputs (larger images) or, for sequence models, larger embedding sizes or sequence lengths. This can increase the amount of information the model can handle.
\item \textbf{Overall Parameters:} Any combination of the above that increases the total parameter count (and usually the required compute). Often used as a general measure of model size.
\item \textbf{Compute per Inference:} Use more computation at test-time per sample, for example through multiple forward passes or iterative refinement, without necessarily increasing the number of parameters.
\end{itemize}

Crucially, scaling up a model is not as simple as just stacking more layers or adding more neurons; naively increasing size can lead to problems such as vanishing gradients, overfitting, or infeasible training times. Over the years, researchers have developed techniques to scale models effectively while maintaining trainability and efficiency. For instance, architectural innovations (like residual connections and normalization) have enabled much deeper networks than were previously possible, and optimization strategies (like distributed training) have allowed these gigantic models to be trained on available hardware.

In this lecture, we will explore the various facets of model scaling in deep learning. We will start with depth-based scaling in vision models, looking at how architectures like VGG, ResNet, Inception, and DenseNet \cite{VGG, ResNet, InceptionNet, DenseNet} progressively increased network depth and complexity to improve accuracy. Next, we will discuss more \emph{efficient} scaling strategies, including both automated methods (neural architecture search) and human-designed approaches, exemplified by EfficientNet and RegNet \cite{EfficientNet, RegNet, NAS}. We will then examine scaling in the context of Transformers and large language models, where growth in parameter count (GPT-3, Chinchilla \cite{GPT3, Chinchilla}) has led to surprising new capabilities. Another important angle is scaling the \emph{computation at inference time}, or ``test-time compute,’’ which involves allowing models to perform more reasoning steps when answering a question or making a prediction. We will also highlight the often-underappreciated role of \textbf{normalization techniques} (like Batch Normalization) in enabling stable training of deep networks.

Moreover, scaling a model is not only about model architecture and training algorithms, but also about practical engineering: we will cover \textbf{efficient building blocks} (such as depthwise separable convolutions in MobileNet \cite{MobileNet}) that make models lighter, as well as \textbf{distributed training} techniques (data/model parallelism and tools like DeepSpeed) that allow training massive models across many devices. We must also consider the \textbf{environmental and computational tradeoffs}: larger models can be extremely costly to train and deploy, so efficiency and sustainability are growing concerns. We’ll discuss how specialized \textbf{AI accelerators} (GPUs, TPUs, etc.) and inference-time optimizations (quantization, pruning, etc.) are used to manage these costs. Finally, we’ll conclude with a summary and outlook on the future of model scaling.

\section{Depth-Based Scaling of Networks}
One of the most straightforward ways to scale a neural network is by increasing its depth (i.e., the number of layers). Deeper networks can compute more complex functions and hierarchical features, as each additional layer can learn to represent increasingly abstract aspects of the data. However, making networks deeper historically encountered obstacles: deeper models were harder to train due to vanishing/exploding gradients, and they often required more data to avoid overfitting. In this section, we discuss how a series of landmark convolutional network architectures successfully pushed to greater depths and achieved higher performance:

\subsection{Going Deeper: VGG Networks}
The \textbf{VGG network} \cite{VGG} (by Simonyan and Zisserman, 2014) was one of the first architectures to demonstrate the power of significantly increased depth in CNNs. Prior to VGG, the state-of-the-art ImageNet model (AlexNet, 2012) had only 8 learnable layers. VGG extended this to 16 and 19 layers in its two main variants (VGG-16 and VGG-19). The architecture was built with a very uniform design: it used a stack of $3\times 3$ convolutional layers throughout, with periodic downsampling via max-pooling, and doubled the number of filters after each pool (64, 128, 256, 512, …). Despite the simplicity of using only $3\times3$ conv filters, VGG showed that just making the network deeper (with more layers) and narrower (small filter sizes, small stride) can greatly improve accuracy. The deepest VGG model (19 layers) achieved top-tier performance in the 2014 ImageNet competition, proving that depth itself was a key to better representations.

However, VGG networks also highlighted some downsides of naive depth scaling. The model had \textit{very large numbers of parameters} (e.g., about 138 million in VGG-16, largely due to the fully-connected layers at the end) and was computationally expensive, making it slow to train and use. The large parameter count meant VGG was prone to overfitting and heavily reliant on regularization and huge datasets. Nevertheless, the success of VGG was a turning point: it established a new baseline that deeper (and conceptually simpler) networks can outperform shallower ones, inspiring further exploration into depth.

\subsection{Residual Networks (ResNets)}
While VGG proved depth can help, training extremely deep networks remained challenging. Simply stacking more layers beyond a certain point often led to higher training error due to optimization difficulties (a phenomenon observed as “degradation” of training accuracy with depth). The breakthrough that overcame this hurdle was the introduction of \textbf{Residual Networks (ResNets)} by He et al. (2015) \cite{ResNet}. ResNets introduced \textbf{skip connections}, which are identity shortcuts that add the input of a layer (or block of layers) to its output. In other words, a residual block computes a function $F(x)$ and then returns $x + F(x)$ as the output. These skip connections effectively turn the network into learning residual mappings, which are easier to optimize than trying to learn a full mapping from scratch.

Residual connections alleviated the vanishing gradient problem, because gradients can flow directly through the skip paths back to earlier layers. This enabled the successful training of networks far deeper than anything before. The original ResNet paper showed results with 50, 101, even 152 layers, all trained to high accuracy on ImageNet \cite{ResNet}. Remarkably, a ResNet-152 (152 layers) not only outperformed shallower models, but it did so with fewer parameters than VGG-19. This is because ResNets made heavy use of $1\times1$ convolutions inside the residual blocks to reduce and then re-expand the number of channels (the \textbf{bottleneck} design), which keeps the parameter count manageable even as depth increases. The success of ResNets was so convincing that residual connections have become a standard component in virtually all very deep networks since. For example, later architectures in vision, such as ResNeXt and EfficientNet, and in NLP, such as Transformers, all utilize some form of residual (skip) connection to ease training of deep layers.

In summary, ResNet demonstrated that with the right architecture (residual blocks with normalization, as we’ll discuss later), depth scaling has essentially no fundamental barrier: networks with 100+ layers can \emph{not only} be trained, but also generalize well, as long as they’re designed to mitigate optimization issues. This opened the door to going deeper and deeper whenever more performance was needed, without being stuck by training divergence. After ResNet, researchers even tried thousand-layer networks as experiments (e.g., ResNet-1001) and found they could still train, although in practice, other limitations like data and compute become the bottleneck.

\subsection{Inception Modules and Going Wider}
Around the same time as VGG and ResNet, another line of research explored scaling not just by making chains of layers deeper, but by making individual layers \emph{wider and more structurally complex}. The \textbf{Inception architecture} (also known as GoogLeNet) introduced by Szegedy et al. (2014) \cite{InceptionNet} is a prime example. Instead of a strictly sequential layer-by-layer design, Inception networks use modules where multiple types of transformations are done in parallel and then concatenated. An Inception module typically contains parallel branches: one with a $1\times1$ convolution (for dimensionality reduction), one with a $3\times3$ conv, one with a $5\times5$ conv (or two $3\times3$ convs to reduce cost, in a later Inception version), and one with a pooling layer. The outputs of these branches are concatenated along the channel dimension and fed to the next stage.

The motivation was to let the network learn both local features (small receptive fields) and more global features (larger receptive fields) at the same layer, instead of committing to one filter size. This can be seen as increasing the \emph{width} of the network in a clever way: the model is wider (multiple parallel paths) but each path is constrained (some are small convs, some are big convs, etc.), so the overall parameter count and compute are kept under control. Inception v1 (GoogLeNet) achieved very strong performance on ImageNet with 22 layers, but importantly, it did so with a much smaller footprint than VGG (GoogLeNet had only ~5 million parameters, vs VGG’s 138 million) by virtue of these efficient modules with lots of $1\times1$ compressions.

Subsequent versions, Inception v2 and v3, further improved the module’s efficiency (for example, factorizing $5\times5$ conv into two $3\times3$ convs, and introducing BatchNorm which we’ll discuss later). Inception v4 and Inception-ResNet later combined the ideas of Inception and ResNet, adding residual connections to Inception-style modules for easier training. The Inception family illustrates another aspect of scaling: you can scale \textit{depth}, but also \textit{width and multi-branch complexity}, to get better performance. While ResNet showed raw depth can be pushed, Inception showed that a thoughtfully structured wide architecture can extract richer features without blowing up computation.

\subsection{Dense Connections: DenseNets}
Following the ResNet success, researchers wondered if one could connect layers even more densely than the simple skip connections. The \textbf{DenseNet} architecture (Huang et al., 2017) \cite{DenseNet} is an extreme case of connectivity: every layer in a DenseNet is connected to \emph{every earlier layer} (within each dense block). In a dense block of $L$ layers, layer $m$ receives as input the feature-maps of all previous $m-1$ layers (they are concatenated together), and its own output is then passed to all future layers. This means the $n$-th layer has $n$ inputs (all previous feature maps), and it computes some new feature maps which are then added to the growing set.

Why do this? Similar to ResNet, dense connections also help with gradient flow (any layer can directly access the loss gradient from later layers because of the connections). But additionally, DenseNet encourages extreme \textbf{feature reuse}. Earlier layer outputs are fed directly into many later layers, so later layers can cheaply leverage features computed earlier, rather than re-computing similar features. This can make the network more parameter-efficient. In fact, DenseNets were shown to achieve comparable or better accuracy than ResNets with significantly fewer parameters. For example, a DenseNet with about 8 million parameters can match or surpass a ResNet with 20+ million on ImageNet, by virtue of this feature reuse.

DenseNet’s scaling was primarily in \emph{depth} (they built DenseNets with depths like 121 or 201 layers), but with the caveat that each layer is very small (e.g., producing only 32 feature maps) since many layers contribute to the final representation. One downside is that the memory usage can be high (because all earlier layer outputs are kept around to feed into later ones), and the computations due to concatenation can also become expensive as the effective width grows with depth. Nonetheless, DenseNet demonstrated another successful strategy for scaling depth: make the network deeper but mitigate redundant computation by connecting everything, so each layer does only a small incremental transformation. DenseNets also have a regularizing effect in that no feature can go “unseen” by the rest of the network – if a layer produces something useless, a later layer could learn to ignore it since all features are available.

In summary, depth-based scaling has been a fundamental driver of progress:
\begin{itemize}
\item VGG showed we can dramatically increase depth (with a homogeneous architecture) to gain accuracy, at the cost of more computation.
\item ResNet solved the key optimization problem, enabling very deep networks to train successfully through residual connections.
\item Inception showed that in addition to depth, increasing the internal width and multi-scale processing of layers can yield efficient accuracy gains.
\item DenseNet proved that dense feature reusage can allow going deep without a blow-up in parameters, by weaving layers together.
\end{itemize}
These innovations are often combined in modern architectures (e.g., a modern CNN might have residual blocks, some Inception-like splits, use $1\times1$ bottlenecks, etc.). Depth remains a critical axis of scaling: today, having dozens or even hundreds of layers is commonplace in state-of-the-art models for vision and other domains.

\section{Efficient Model Scaling: Neural Architecture Search and Compound Scaling}
While simply increasing depth or width can improve performance, it often does so with diminishing returns and increasing costs. A key question is: \emph{how do we scale a model in the most efficient way}? Instead of blindly making a network larger, can we allocate capacity in a smarter manner to get more accuracy per parameter or per FLOP? This has led to two important approaches: (1) using automation (\textbf{Neural Architecture Search, NAS}) to discover better architectures and scaling strategies, and (2) deriving principled scaling rules or design guidelines (sometimes called \textbf{compound scaling}) to balance different aspects like depth, width, and resolution. We’ll discuss two representative works: EfficientNet, which combined NAS with a simple but effective scaling rule, and RegNet, which explored manual design to find a family of well-behaved scalable models.

\subsection{Neural Architecture Search (NAS) for Scaling}
Neural Architecture Search (NAS) refers to techniques that automate the design of neural network architectures. Instead of a human manually specifying the number of layers, filter sizes, etc., NAS uses an algorithm (like a genetic algorithm, reinforcement learning agent, or gradient-based method) to search through the space of possible architectures and find high-performing ones. Early NAS work by Zoph and Le (2017) \cite{NAS} demonstrated that it’s possible to learn convolutional architectures that rival or even beat manually-designed networks. They employed a reinforcement learning controller to sample architecture descriptions (like how many filters, kernel sizes, skip connections, etc.), trained each candidate on data, and used the performance as a reward signal to improve the controller. This process, while conceptually straightforward, was extremely computationally expensive at first (requiring tens of thousands of training runs). Nonetheless, it proved the point that algorithms can discover non-intuitive architectures. For instance, NAS found motifs like skip connections and convolutions of varying sizes, some of which resembled Inception-like modules or other patterns.

One notable result of NAS was the \textbf{NASNet-A} architecture (Zoph et al., 2018) and the \textbf{AmoebaNet} series (Real et al., 2018, evolved with evolutionary strategies). NASNet-A, when scaled up to a large model for ImageNet, slightly outperformed human-designed models of similar cost. However, the discovered architecture was quite complex and irregular, which made it harder to interpret or implement efficiently. NASNet did introduce the idea of \textbf{cells}: rather than searching the entire large network structure, they searched for a small cell (a subgraph of layers) that can be stacked repeatedly to form the full network. This cell-based design made it easier to scale the discovered architecture to different depths and widths.

\subsection{Compound Scaling and EfficientNet}
A significant advancement in scaling strategy came with \textbf{EfficientNet} (Tan \& Le, 2019) \cite{EfficientNet}. EfficientNet tackled the question: given a baseline model, how should we increase its depth, width, and input resolution to get the best improvement for a given increase in compute? Before EfficientNet, many practitioners would do ad-hoc scaling like “let’s try doubling the number of filters and see” or “let’s add more layers until we hit memory limit”. Instead, EfficientNet proposed a \textbf{compound scaling} method. They introduced coefficients $\phi$ that uniformly scale the network’s depth, width, and resolution according to preset ratios. For example, in EfficientNet, increasing $\phi$ by 1 might mean: increase depth by factor $\alpha$, width by factor $\beta$, and image size by factor $\gamma$, where $\alpha, \beta, \gamma$ are chosen such that the overall FLOPs roughly multiply by a certain amount (e.g., $2^\phi$ growth in FLOPs).

Concretely, EfficientNet started from a small but efficient baseline model (EfficientNet-B0), which itself was found via NAS (they used a mobile-sized NAS search similar to how \textbf{MnasNet} was developed, focusing on depthwise separable conv blocks with squeeze-and-excitation SE layers for efficiency). Then they set $\alpha=1.2$, $\beta=1.1$, $\gamma=1.15$ (these numbers were determined via small grid search under constraints) such that when $\phi$ increases, depth $\approx \alpha^\phi$, width $\approx \beta^\phi$, and resolution $\approx \gamma^\phi$. By scaling up from B0 to B1, B2, … up to B7 using this compound rule, they obtained a family of models from small to large, each roughly optimal in accuracy for its model size.

The results were impressive: EfficientNet models significantly outperformed other networks at the same level of compute. For instance, EfficientNet-B4 achieved accuracy similar to ResNet-50 but with much less compute, and the largest model EfficientNet-B7 (with about 66M parameters) attained \textbf{state-of-the-art ImageNet accuracy of $\sim$84.4\% top-1} at the time, while being an order of magnitude smaller and faster than previous best models. To put it in perspective, one contemporary model called GPipe (an extremely large NAS-based model with 557M parameters) had slightly lower accuracy (84.3\%) than EfficientNet-B7, despite B7 being 8$\times$ smaller and faster \cite{EfficientNet}. This demonstrated that a well-scaled mid-sized model can beat an unscaled giant model.

EfficientNet’s approach has two key takeaways: First, when scaling up a model, it is beneficial to balance multiple factors (depth, width, input size) rather than just one. Intuitively, if you only make the network deeper but keep it very narrow, it might become too bottlenecked at each layer; if you only make it wider but not deeper, it might not have enough sequential layers to abstract high-level concepts; if you use huge images but the network is too small, it can’t take advantage of the extra detail. EfficientNet formalized one way to achieve this balance. Second, using NAS to find a good \textit{starting} block or cell can complement scaling: EfficientNet didn’t search every model size, it just searched for a good base architecture (B0) in a small regime, and then scaled that up with a rule. This was much more computationally feasible than doing a full NAS for a large model, and it gave excellent results.

Today, the EfficientNet paper and models are a standard reference for how to do principled model scaling. Variants like EfficientNetV2 have further improved aspects like training speed and used progressive learning (smaller resolution to larger during training). But the core idea remains widely influential, even beyond vision: the notion of compound scaling can be seen in some transformer scaling setups too (e.g., scaling width vs depth of transformer layers).

\subsection{Designing Network Families: RegNet}
Another approach to efficient scaling is to manually derive design principles that produce a family of models parameterized by size. Facebook AI Research’s \textbf{RegNet} (Radosavovic et al., 2020) \cite{RegNet} is an example of this philosophy. Instead of searching for an architecture via NAS, the authors systematically explored the space of CNN architectures by varying network depth, width (per stage), group convolution widths, etc., and observing what patterns yield the best trade-offs. They were looking for a parameterized design space where simple functions describe the optimal model configurations at each scale.

Surprisingly, they discovered that the best-performing networks across different sizes followed a simple regular pattern: specifically, the number of channels in each stage of the network as a function of the stage index tends to increase in a roughly linear or quantized-linear fashion (rather than arbitrary nonlinear patterns). In simpler terms, a RegNet is basically a \textbf{regularly shaped network} – e.g., it might start with 32 channels, then 48, 80, 112,… increasing by a constant amount or ratio at each block, and with a certain number of blocks. These RegNet models are described by just a few parameters (like initial width, slope of width increase, depth, group width) and yet they achieved excellent performance across a range of FLOP budgets.

One of the selling points of RegNet was that these networks, being very regular and simple, are \emph{hardware-friendly} and fast in practice. They compared RegNet vs EfficientNet: while both had good accuracy for a given FLOP count, RegNets were up to 5x faster on GPUs in some cases \cite{RegNet}, because EfficientNet’s compound scaling led to some layers being very bottlenecked or having odd dimensions (and lots of small depthwise convs and SE modules), whereas RegNet uses mostly standard convs in a regular pattern. This shows an interesting trade-off: NAS can find maybe the absolute optimal accuracy, but the resulting architecture might be complex. A human-designed regime like RegNet might be slightly less optimal in accuracy per FLOP, but win in actual efficiency and simplicity.

RegNet and EfficientNet families both illustrate the concept of \textbf{model families} that span from small to large. Instead of designing a one-off model, researchers now think in terms of scalable families: provide a recipe to get a model at 50M FLOPs, 200M FLOPs, 1B FLOPs, etc., all with similar design DNA. This is very useful in practice because one often needs models of different sizes for different use cases (mobile vs server, etc.).

In summary, efficient scaling approaches aim to get the most out of each parameter or each operation:
\begin{itemize}
\item NAS methods automate the search for good architectures. They have found novel architectures, especially for smaller models, that can then be scaled up. However, pure NAS is expensive and can yield complicated designs.
\item Compound scaling (EfficientNet) provides a simple formula to scale models in multiple dimensions simultaneously. It resulted in record-breaking efficient models by balancing depth/width/resolution.
\item Manual design exploration (RegNet) can reveal simple patterns for model scaling, giving families of networks that are easy to implement and still very performant.
\end{itemize}
Ultimately, these approaches share the goal of pushing accuracy higher \emph{without simply throwing exponentially more resources} – they try to use parameters and compute in a smarter way. This has become increasingly important as we reach scales where each new model can cost millions of dollars to train; we want to ensure that such investments are as optimal as possible.

\section{Scaling Transformers and Language Models}
The deep learning scaling journey is perhaps most dramatically illustrated in the realm of natural language processing with \textbf{Transformer-based language models}. The Transformer architecture (Vaswani et al., 2017) introduced self-attention mechanisms that made it feasible to train very large sequence models due to their parallelizability and stable training dynamics (thanks in part to layer normalization and residual connections in every layer). Over the past few years, we have seen an unprecedented growth in the size of language models, leading to qualitatively new capabilities.

\subsection{From Millions to Billions of Parameters}
Early NLP models in the 2010s, such as LSTMs or even the first Transformers for translation, had on the order of tens of millions of parameters. The landscape shifted with unsupervised pretraining: models like \textbf{BERT} (Devlin et al., 2018) had hundreds of millions of parameters (BERT-Large ~340M) and showed that pretraining on large text corpora could give universal language understanding that transfers to many tasks. BERT’s success encouraged scaling up model size and data together.

OpenAI’s \textbf{GPT} series took this further. \textbf{GPT-2} (Radford et al., 2019) had 1.5 billion parameters, a jump of nearly an order of magnitude from BERT. GPT-2 was trained to simply predict the next word on a massive dataset of internet text, and it surprised the community with its ability to generate coherent paragraphs of text, showing that unsupervised language modeling at scale leads to fluent generative capabilities.

The real watershed moment was \textbf{GPT-3} (Brown et al., 2020) \cite{GPT3}, which contained a staggering 175 billion parameters. GPT-3 was more than 100$\times$ larger than GPT-2. This scaling was enabled by significant computing power and engineering (for instance, using hundreds of GPUs in parallel, as we’ll discuss in distributed training). GPT-3 demonstrated something remarkable: at this scale, the model was able to perform tasks \emph{in zero-shot or few-shot settings} that normally would require training or fine-tuning. For example, given a question, GPT-3 can answer it reasonably well without any task-specific training, or given a few examples of a new task in the prompt (few-shot learning), it can often continue the pattern and complete the task. This emergent ability was not clearly present in smaller models. Thus, scaling up to 175B parameters unlocked a new regime where the model starts to act as a general-purpose intelligent agent, to an extent.

The trend did not stop at GPT-3. Other organizations built even larger models:
Google’s \textbf{PaLM} (2022) had 540 billion parameters, Microsoft/Nvidia’s \textbf{Megatron-Turing NLG} had 530 billion, and there were even experimental sparse models with trillions of parameters (like the Switch-Transformer with 1.6 trillion parameters, though only a fraction are active per token due to a mixture-of-experts design). Each time, these models set new records on language benchmarks and demonstrated increasingly sophisticated behavior (e.g., better understanding of nuance, some reasoning ability, code generation, etc.).

One interesting observation made during this period is that larger models often follow a \textbf{scaling law}: performance (e.g., measured in perplexity or accuracy on some task) tends to improve as a power-law as we increase model size, dataset size, and compute. Initially, studies (e.g., by Kaplan et al., 2020 at OpenAI) suggested that model performance improves predictably with more parameters if you also feed it enough data, and they extrapolated that trend outwards. This provided a theoretical justification for building bigger models: if you can afford 10x more compute, a 10x bigger model (with appropriately more data) will reliably give better results, following a log-linear trend.

However, simply making the model huge without adjusting other factors can be suboptimal. A pivotal study from DeepMind in 2022, often referred to by the codename \textbf{Chinchilla} \cite{Chinchilla}, revisited these scaling laws by considering \emph{compute budget as the fundamental constraint}. They asked: for a given amount of compute (FLOPs spent in training), what is the optimal model size and amount of training data? The surprising finding was that many existing large models were actually \emph{undrained} in terms of data. For instance, GPT-3 used 300 billion tokens for training; Chinchilla analysis suggested that with the compute GPT-3 used, it should have used about 4 times more data and a smaller model to get the best results. The rule of thumb they found was to scale model size and training data \textbf{in tandem} — roughly, parameter count should be proportional to the number of training tokens (specifically, for every doubling of model parameters, also double the dataset size).

To prove this, they trained \textbf{Chinchilla}, a model with only 70B parameters (much smaller than GPT-3’s 175B), but on 1.4 trillion tokens (about 4.7x more data than GPT-3). Importantly, the total compute used was kept similar to that of a larger model like Gopher (280B) or GPT-3. The result was that Chinchilla \emph{outperformed} Gopher (280B), GPT-3 (175B), and other models on a wide range of language tasks \cite{Chinchilla}. In other words, if you have e.g. X GPU-days to train a model, you’re better off with a moderately sized model trained on lots of data than an extremely large model trained on a limited data budget. This was a course-correction in the scaling narrative: bigger is not better unless you also increase the training duration/data.

The Chinchilla finding has important implications. It suggests current large models might not be fully utilizing their capacity because they haven’t seen enough data to learn all that they could. For future projects, it advises an optimal balance: don’t just blindly push parameter counts; also invest in gathering more data or training for more steps. It also means that if one is willing to train for longer, one could get away with a smaller (cheaper) model with equal or better performance, which has downstream benefits like faster inference and easier deployment.

\subsection{Emergent Abilities and Limits of Scaling}
As language models have scaled, there are anecdotal reports of \textbf{emergent behaviors} — capabilities that were absent in smaller models but appear once a certain scale threshold is crossed. We already mentioned few-shot learning as one example emerging around GPT-3’s scale. Others include better understanding of human instructions (like instruction-following ability which becomes much stronger at large scale, especially with fine-tuning like InstructGPT), basic common-sense reasoning, and the ability to write code or perform arithmetic with increasing accuracy. Some tasks show a sudden jump in performance when model size increases from, say, 10B to 100B, rather than a smooth continuum, which is a fascinating phenomenon under study.

On the other hand, scaling alone doesn’t solve all problems. Very large models still make mistakes, can be brittle or output toxic or incorrect information, and they become increasingly expensive to train and run. There are also fundamental limits: eventually, one might run out of relevant training data on the internet, or the returns diminish to the point that the cost isn’t justified by the small accuracy gain. Some researchers are investigating alternatives and complements to pure scaling, such as incorporating knowledge databases or doing more efficient reasoning (which brings us to the next topic of test-time compute).

In summary, the Transformer and LLM era has exemplified model scaling:
\begin{itemize}
\item Parameter counts went from millions to hundreds of billions within a few years, yielding qualitatively new capabilities in language generation and understanding.
\item The positive feedback loop: larger models $\to$ better results $\to$ justification for even larger models, was strong, but tempered by findings like Chinchilla that emphasize efficient use of compute.
\item Scaling laws provide a rough roadmap, but careful tuning of data vs model size is required to truly get optimal performance.
\item These enormous models require advanced techniques (model parallelism, etc.) to train, which we’ll touch on later.
\end{itemize}

Transformers are particularly amenable to scaling because their basic block (self-attention + feed-forward sublayers) is highly flexible and the model is mostly homogeneous and can be copied many times (like stacking more transformer layers). This modularity, combined with residual connections and LayerNorm, means you can often just increase the number of layers or the hidden size and things still train well (provided you adjust hyperparameters properly). That isn’t always the case for other architectures, making Transformers something like the ``VGG of NLP’’ initially (simple block repeated) but with the trainability of ResNet, thus an ideal candidate for scale-up.

\section{Test-Time Compute: Scaling Reasoning at Inference}
All the scaling we discussed so far involves making the model or training process larger. But another dimension of scaling is how much computation a model is allowed to perform \emph{when it is actually used} to make a prediction or generate an output. Typically, once a model is trained, we fix its architecture and weights, and for each input, we run a fixed sequence of operations (one forward pass) to get an output. However, what if for harder problems, we let the model do more work – think longer or consult external knowledge – before finalizing an answer? This idea is sometimes called \textbf{test-time compute} scaling or \textbf{reasoning-based scaling}.

The core observation is that humans faced with a difficult question will spend more time and break the problem into steps, whereas most neural networks by default give an answer in a single pass no matter the complexity of the query. There has been a recent shift towards allowing “slow thinking” in AI models:
\begin{itemize}
\item In the context of language models, techniques like \textbf{Chain-of-Thought (CoT) prompting} allow the model to generate intermediate reasoning steps (as if it’s writing down its thought process) before giving a final answer. For example, instead of directly asking the model a math word problem and expecting an answer, we prompt it to first produce a step-by-step solution. This effectively multiplies the amount of computation (each step is another forward pass or another segment of output it must produce) but greatly improves accuracy on tasks requiring reasoning. The model is using more compute per query to think things through.
\item Another approach is using \textbf{self-refinement or majority voting}. A model can generate multiple candidate answers or reasoning traces and then either pick the most consistent one or refine its answer based on those attempts. This is like performing inference multiple times and aggregating results, which again uses more compute for a better outcome.
\item We also see forms of test-time compute in earlier AI systems: e.g., \textbf{AlphaGo/AlphaZero} did a forward pass through a neural network to get move probabilities and values, but then ran an extensive Monte Carlo Tree Search (MCTS) simulation (which involves many network evaluations and combining them via tree search) to decide on the best move. The network itself wasn’t huge (millions of parameters), but the test-time search made the overall system far stronger than the raw network policy. In effect, search allowed a relatively compact model to achieve superhuman performance by leveraging more computation.
\item Some research models incorporate an explicit iterative loop, such as \textbf{Recurrent Relational Networks} or \textbf{Neural Reasoners}, where the model can perform multiple computation cycles per input. For instance, there’s an idea of treating the depth of a network as dynamic: a model might apply the same layers repeatedly until some condition is met (Adaptive Computation Time, Graves 2016). This means easy inputs exit quickly, hard inputs take more iterations. This is one way to scale computation on demand.
\item Retrieval-based models: Another form of increasing test-time compute is allowing the model to query external knowledge (like a search engine or database). Systems like OpenAI’s WebGPT or retrieval-augmented generation will perform search queries or lookups for a question and then use those results to compose an answer. This pipeline uses more processing per query (the cost of searching the web or database lookups), but it can dramatically improve accuracy and keep the model itself smaller since it doesn’t have to memorize everything.
\end{itemize}

The concept of scaling test-time compute is closely tied to the idea of \textbf{reasoning}. Instead of relying purely on what the static network weights encode, the model can engage in a computation process (potentially involving sequences of reasoning steps or multiple passes) to arrive at an answer. This can sometimes compensate for not having an extremely large parametric memory. For example, a 6-billion-parameter model with a well-implemented reasoning strategy and multiple inference steps can potentially solve tasks that a 6-billion parameter model in one-shot cannot, and might even rival a larger 100B model on some complex tasks, by virtue of “thinking harder”.

One concrete demonstration is in mathematical problem solving. A big model like GPT-3 (175B) might only solve a certain fraction of multi-step math problems if it answers in one go. But a much smaller model that is allowed to do scratch work via chain-of-thought and even check intermediate results can solve a higher fraction of those problems, albeit taking a few passes to do so. Essentially, compute is an alternative currency to parameters: you can either pre-compute a lot (big training, big weights) or compute more on the fly (reasoning steps) to achieve an outcome.

It’s worth noting that scaling test-time compute has its own challenges. The model needs to be guided to use the extra compute effectively (hence methods like chain-of-thought prompting explicitly tell it to output intermediate steps). If the model isn’t trained or prompted properly to do multi-step reasoning, simply giving it a loop or more time might not help. There’s active research in training models to plan, to self-reflect, and to use tools so that when confronted with a new task, they can break it down into manageable subtasks.

Also, more compute at inference means slower responses, which might be a trade-off. In some applications, you can’t afford to have the model think for a whole minute if the user expects an answer in one second. But if you do have the luxury (like offline analysis, or non-real-time tasks), then you can squeeze more quality out of the model by letting it churn longer on the problem.

In summary, \textbf{reasoning-based scaling} at test time is an exciting complement to the traditional parameter-based scaling:
\begin{itemize}
\item It allows even a fixed-size model to become more accurate by using additional computation per input (like an ensemble of one model with itself, or an internal dialogue).
\item It is particularly useful for tasks that naturally involve multiple steps or search (math, logic puzzles, planning, knowledge lookup).
\item In effect, it treats the neural network not just as a one-shot function approximator, but as a component in a larger iterative algorithm.
\end{itemize}
We should expect future AI systems to leverage this more, combining large parametric models with clever inference-time algorithms to get the best of both worlds.

\section{The Role of Normalization in Stable Scaling}
One crucial enabler behind training very deep or very large models is the use of \textbf{normalization techniques} in the network. Without normalization, many of the scaling successes (ResNets, Transformers, etc.) would not have been possible or would have required much more careful tuning. The idea of normalization is to adjust the distributions of layer inputs or outputs during training to maintain stability.

The most famous example is \textbf{Batch Normalization (BatchNorm)} introduced by Ioffe and Szegedy (2015) \cite{BatchNorm}. BatchNorm operates by normalizing the activations of a layer for each mini-batch. Concretely, for each channel/feature $j$ in a layer, BatchNorm will compute the mean and variance of that feature across the examples in the current batch, then subtract the mean and divide by the standard deviation, and finally apply a learned linear scaling and offset (gamma and beta parameters) to allow the layer to still represent identity transformations if needed. This process keeps the activations in a relatively stable range as the network trains.

Why was BatchNorm so revolutionary? It addressed the problem of \textbf{internal covariate shift}, which refers to the way that as lower layers’ parameters change, the distribution of inputs to higher layers shifts, making training harder (each layer has to continuously adapt to changing distribution from the layer below). By normalizing, each layer sees a more stationary distribution of inputs over the course of training. Practically, BatchNorm allowed much higher learning rates and made the training of deep networks far more robust to initialization. In the original paper, they demonstrated that BatchNorm could enable training of networks that completely failed to converge otherwise, and it often also improved final accuracy. It also had a mild regularization effect (because each batch’s statistics add some noise), often reducing the need for other regularizers like Dropout in convolutional nets.

BatchNorm was a key ingredient in the success of VGG-like networks and ResNets. For example, ResNets insert a BatchNorm after each convolution (and before adding the residual) which keeps the residual addition stable. Without BatchNorm, extremely deep ResNets might still have struggled. In fact, after BatchNorm’s introduction, virtually all high-performance CNNs incorporated it (or a variant) – it became almost implied when scaling depth.

However, BatchNorm has a limitation: it depends on batch statistics, which means the behavior can be tricky during inference (when you typically switch to using accumulated moving averages of means/variances) and it doesn’t work as well for very small batch sizes or certain tasks like recurrent sequence modeling. For such cases, other normalization methods were developed:
	•	\textbf{Layer Normalization (LayerNorm)} \cite{LayerNorm} (Ba et al., 2016) normalizes across the neurons in a layer for each single example (rather than across examples). This doesn’t depend on other examples in a batch and is suitable for RNNs/Transformers. In Transformers, every sub-layer (attention or feed-forward) is preceded or followed by a LayerNorm. This keeps the scale of activations under control even as the model depth (number of transformer layers) grows. Without layer norm, training large Transformers might diverge or be very sensitive to learning rate.
	•	\textbf{Instance Normalization}, \textbf{Group Normalization} (Wu \& He, 2018), etc., are other variants that normalize over different dimensions, useful in specific contexts (instance norm for style transfer, group norm as a replacement for BN when batch sizes are small).
	•	More recently, \textbf{RMSNorm} (Root Mean Square Norm) and other normalization tweaks have been used especially in very large language models (some LLMs use RMSNorm which is like layer norm without the mean subtraction, to simplify things).

Normalization helps with scaling in another way: it prevents activation magnitudes from blowing up or collapsing when networks get deeper or when learning rates are high. For example, if you tried to stack 100 fully connected layers without any normalization or special initialization, the activations and gradients might either explode to infinity or shrink to zero by the time they reach the end, due to multiplicative effects. Techniques like careful weight initialization (e.g., Xavier/He initialization) can mitigate this to some extent, but normalization actively keeps things in check throughout training.

In practice, when designing a scaled-up model, adding normalization layers is now a standard part of the recipe:
	•	In a CNN, we usually do Conv -> BatchNorm -> ReLU as a basic trio, repeated.
	•	In a Transformer block, we do LayerNorm -> Attention -> LayerNorm -> FFN (plus residual adds in between).
	•	Even in very deep MLPs or other architectures, some form of normalization or scaled initialization is used to ensure gradient flow.

It’s worth noting that normalization itself has some computational and memory overhead, and at inference time batchnorm can be folded into the preceding linear layer (since it’s a linear operation when using fixed mean/var), so it’s not a big cost there. But the benefits during training far outweigh the slight cost.

There have been attempts to remove the need for batch normalization, for instance \textbf{self-normalizing networks} (SELUs) or other normalization-free networks. For example, there’s a concept of \textbf{Normalization-Free ResNets} (Brock et al., 2021) that use careful initialization and activation scaling (and sometimes smaller learning rates) to train deep nets without BatchNorm. They did manage to train 1000-layer nets without BN. This is interesting academically, but in most cases it’s just easier to use BN or LN.

In summary, normalization techniques like BatchNorm and LayerNorm have been unsung heroes in enabling stable scaling:
\begin{itemize}
\item They dramatically improved the trainability of deep networks by keeping activations well-behaved.
\item BatchNorm in particular made the optimization landscape smoother, allowing faster training and often better generalization.
\item LayerNorm proved essential for Transformers, which are now the backbone of large language models.
\item The presence of normalization is one reason why architectures can be scaled to great depth/size without losing performance or encountering optimization failures.
\end{itemize}

Anyone building or training a deep neural network today will almost always include some normalization after each major layer. It’s part of the standard pattern for a reason—without it, many of the successes in going big might not have materialized.

\section{Efficient Building Blocks for Scalable Models}
Another aspect of model scaling is the design of more \textbf{efficient layer building blocks} that allow the creation of large networks without proportional increases in computation. If each layer of a network can be made cheaper (in terms of FLOPs or parameters) while still expressive, you can afford to have more layers or a wider network under the same resource constraints. This is particularly important for deploying models on limited hardware (like mobile phones) or when trying to train very large models with fixed computing budgets. We’ll discuss a few such building block innovations, notably \textbf{depthwise separable convolutions} popularized by MobileNet \cite{MobileNet}, as well as others like bottleneck layers and group convolutions.

\subsection{Depthwise Separable Convolutions and MobileNets}
A standard convolutional layer in a CNN is quite expensive: for an input with $M$ feature maps (channels) and output with $N$ feature maps, a kernel size $k \times k$, the layer has $M \times N \times k^2$ weights and for each output feature map computation it does similar number of multiplications. In other words, computational cost grows with $M \times N$. In a typical CNN, both $M$ and $N$ are in the tens or hundreds, so this can be large.

A \textbf{depthwise separable convolution} factorizes the convolution into two steps:
\begin{enumerate}
\item \textbf{Depthwise convolution:} Perform a $k \times k$ convolution independently on each input channel (hence “depthwise”), producing $M$ intermediate feature maps (one per channel). Since each such convolution has $k^2$ weights and operates on one channel, the total weights used here is $M \times k^2$. There is no mixing of information between channels at this stage.
\item \textbf{Pointwise convolution:} Now use $1 \times 1$ convolutions (which we can think of as simple linear combinations) to combine the intermediate features across channels and produce the final $N$ output channels. A $1\times1$ conv that goes from $M$ channels to $N$ channels has $M \times N$ weights (each 1x1 filter has $M$ inputs and we have $N$ such filters).
\end{enumerate}
The total number of weights in a depthwise separable conv is $M \times k^2 + M \times N$. Compare this to a standard convolution’s $M \times N \times k^2$. For typical values (say $k=3$ and $N$ on the order of $M$), the separable conv is much cheaper. For example, if $M = N = 128$ and $k=3$, standard conv uses $128 \times 128 \times 9 \approx 147k$ weights, whereas depthwise separable uses $128 \times 9 + 128 \times 128 \approx 16k + 16k = 32k$ weights, which is nearly 5 times smaller. Similar reductions occur in computation.

This idea was actually used in some earlier architectures in part (like Inception modules effectively used 1x1 convs to reduce channels then a spatial conv). \textbf{MobileNet} (Howard et al., 2017) \cite{MobileNet} was the architecture that fully embraced depthwise separable convolutions to create an extremely lightweight model for mobile devices. MobileNet v1 is basically a streamlined CNN where every convolution is replaced by a depthwise conv + pointwise conv pair (except the very first conv layer). This allowed the network to be very deep and still very fast. The original MobileNet had 28 layers of convs (depthwise + pointwise counted separately) and only 4.2 million parameters, yet achieved around 70\% top-1 accuracy on ImageNet. In contrast, a much larger model like VGG-16 had 138 million parameters and about 74\% accuracy. So MobileNet delivered reasonable accuracy at a tiny fraction of the size by using an efficient building block.

MobileNet v1 also introduced a \textbf{width multiplier} hyperparameter $\alpha$ that could scale down every layer’s channel counts by a factor (to trade accuracy for even more speed if needed) and a resolution multiplier for input image size. These gave developers flexibility to deploy smaller variants if 70\% accuracy was not needed and 60\% could suffice for an even lighter model.

Following MobileNet v1, there was \textbf{MobileNet v2} (Sandler et al., 2018), which further refined the block by introducing \textbf{inverted residual blocks with linear bottlenecks}. This sounds complex but the idea was:
	•	Instead of doing depthwise conv on a narrow set of channels (which might become a bottleneck for information), they \emph{first expand} the number of channels with a $1\times1$ conv (say from $M$ to $t \cdot M$ where $t$ is expansion factor, like 6), then do a depthwise conv on this larger space, then \emph{project down} with a linear $1\times1$ conv back to a smaller number of channels (possibly even smaller than $M$, hence “bottleneck”).
	•	They also added a residual connection around each such block (if input and output dimensions were the same) — hence “inverted residual”, inverted because a traditional ResNet bottleneck first reduces then expands, whereas this block first expands then reduces back.
	•	The use of a linear activation at the final projection (no ReLU on the last layer of the block) was important to not destroy information during the bottleneck projection (ReLU could kill information when collapsing dimensions).

MobileNet v2 achieved even higher accuracy (~72\% on ImageNet) with similar or fewer operations than v1, making it one of the most efficient models for its time. These MobileNets were instrumental for on-device AI, and they also influenced larger model design by showing the effectiveness of depthwise convs.

\subsection{Other Efficient Layer Techniques}
Depthwise separable convs are one powerful tool, but there are others:
	•	\textbf{Bottleneck layers:} We’ve mentioned this in ResNet and MobileNet context. The use of $1\times1$ conv to reduce the dimensionality before a costly operation (like a $3\times3$ conv) and then possibly expand back is a general pattern. ResNets used a $1\times1$ to go from say 256 channels to 64, then a $3\times3$ on 64 (cheaper), then another $1\times1$ to go to 256 (back to original channel count). This significantly cuts down FLOPs while keeping representational power. Most modern CNNs use bottlenecks for anything beyond the smallest layers.
	•	\textbf{Grouped Convolutions:} This is a middle ground between a full conv and depthwise conv. Instead of each filter using all input channels, we split input channels into groups and each filter only sees channels within its group. For example, in AlexNet (2012), group conv (with 2 groups) was originally used for a different reason (split across two GPUs), but ResNeXt (Xie et al., 2017) turned it into a purposeful architectural feature. ResNeXt showed that you can increase the number of groups (making each convolution narrower in scope) while increasing the total number of filters, to get a better trade-off. Essentially, group conv is like having multiple smaller convs operating in parallel on partitioned channels. Depthwise conv is an extreme form of group conv where number of groups = number of input channels.
	•	\textbf{Channel Shuffle:} One issue with group conv is that if you always keep the same grouping, a filter never sees features from other groups, potentially limiting connectivity. \textbf{ShuffleNet} (Zhang et al., 2018) introduced a simple operation called channel shuffle that permutes the channels between groups after each layer, ensuring that information is mixed across groups over layers. ShuffleNet combined grouped 1x1 convs, depthwise convs, and channel shuffle to create an extremely efficient model for mobile (on par with MobileNet v2 in complexity). It essentially generalizes the depthwise separable idea: they had pointwise group conv (cheaper than full conv) + depthwise conv + another pointwise group conv, with shuffling in between.
	•	\textbf{Squeeze-and-Excitation (SE) blocks:} These were introduced in the SENet model (Hu et al., 2018). SE blocks are not about reducing computation of a single conv, but about adding a tiny neural network that does channel-wise attention. Specifically, an SE block takes the output of some layers, pools it to a vector (of length equal to number of channels), passes it through a small bottleneck MLP and sigmoid to produce weights for each channel, and multiplies those weights to the channels (re-weighting them). This is a lightweight way for the network to recalibrate channel importance and yielded significant accuracy improvements (~+1-2\% ImageNet accuracy) for a very minor cost (maybe ~0.5-2\% extra compute). EfficientNet and many other models incorporated SE blocks because they improve efficiency (accuracy per parameter) even though they add a few parameters.
	•	\textbf{Transformer efficient blocks:} In the transformer world, efficient building blocks mean things like optimized attention mechanisms (like sparse attention patterns for long sequences, or replacing softmax attention with linear attention approximations for better scaling). For example, the Performer or Linformer try to reduce the $O(n^2)$ cost of attention for long sequence length $n$. That’s another kind of efficiency which is about scaling to longer inputs rather than scaling model size.

In summary, efficient building blocks allow networks to scale to either larger depths or to work within constrained environments:
\begin{itemize}
\item Depthwise separable convolutions (MobileNet) drastically cut down computation, enabling deep models on low-power devices.
\item Bottlenecks and grouped convs let us widen networks without an explosion of parameters, which was key in models like ResNeXt and also indirectly in transformers (where multi-head attention is somewhat analogous to group processing on different heads).
\item Attention to efficiency at the micro level (like SE blocks adding a big accuracy boost for a few extra ops) yields models that dominate accuracy-vs-compute benchmarks.
\end{itemize}

When designing a network for a specific parameter or FLOPs budget, combining these tricks is now standard. For instance, EfficientNet’s MBConv is essentially a MobileNet v2 inverted residual (which itself uses depthwise + bottleneck + SE). That combination was chosen because it delivers a lot of bang for the buck, allowing the model to devote saved compute elsewhere (like to more layers). Thus, model scaling isn’t just about macro architecture (how many layers) but also about the micro operations chosen for each layer.

\section{Distributed Training for Scalable Deep Learning}
As models and datasets have grown, a single compute device (like one GPU or one TPU chip) is often not enough to train them in a reasonable time (or at all, if the model doesn’t fit in memory). \textbf{Distributed training} refers to using multiple processors (GPUs/TPUs or even whole machines) in parallel to train a single model. Without distributed training, the large-scale deep learning breakthroughs of recent years (like GPT-3, which was trained on thousands of GPUs) would be impossible.

There are several parallelism paradigms, each addressing different scaling challenges:
\begin{itemize}
\item \textbf{Data Parallelism:} This is the most common and conceptually simplest form. If you have $K$ GPUs, you give each GPU a different subset of the training data at each step (like each GPU processes a different batch of examples). All GPUs have a copy of the model parameters. They each compute gradients on their mini-batch, and then these gradients are aggregated (for example, summed or averaged) across GPUs. After that, each GPU updates its copy of the model with the aggregated gradient (which is equivalent to the gradient on the combined batch of all GPUs). The effect is that with $K$ GPUs you can process $K$ times more data per step (effectively a larger batch size, or you can keep batch per GPU same to speed up training by $K$x). Data parallelism scales well if the communication (aggregating gradients) is fast enough not to become a bottleneck. Modern interconnects (NVLink, InfiniBand, etc.) and strategies like \textbf{All-Reduce} make this efficient for dozens or even hundreds of GPUs in parallel. Most medium-sized model training (CNNs, smaller transformers) in practice use data parallelism across multiple GPUs or nodes.
\item \textbf{Model Parallelism:} This is used when the model is so large it cannot even fit on a single GPU, or when we want to split up the compute of a single example across devices. In model parallelism, different GPUs hold different parts of the model. For example, if you have a 100-layer network, you might put 50 layers on one GPU and the next 50 on another; or for a single giant layer (like a huge fully-connected layer with a massive weight matrix), you might split the neurons between GPUs. During a forward pass, the data has to move from one GPU to the next for each layer (like pipelining through the layers), or if splitting within a layer, intermediate results need to be shared. Model parallelism is more complex because it requires partitioning the computation graph and managing communication of activations and gradients between devices. It’s typically not as efficient as data parallelism due to communication overhead, but it’s indispensable for ultra-large models. For example, GPT-3 (175B) model weights could not fit in one GPU memory (which might be 16 GB for a V100 or 40GB for an A100, and GPT-3 175B in half precision would be >300GB), so it must be sharded across many GPUs. One scheme is \textbf{tensor slicing}: break big matrices so each GPU stores a slice and compute collectively. Another is \textbf{pipeline parallelism}: assign contiguous layers to different GPUs and pass micro-batches sequentially through them (this keeps each GPU busy with different samples in different pipeline stages).
\item \textbf{Mixed Parallelism:} In practice, large-scale training uses a combination. For instance, one might use data parallelism across nodes, and within each node use model parallelism to split a large model. Or use pipeline parallelism combined with data parallel groups to strike a balance. The combination is often necessary to scale to many devices without hitting network bandwidth limits or memory limits.
\end{itemize}

Managing all this complexity led to the development of software frameworks and libraries:
	•	\textbf{Horovod} (Uber) was an early library to ease data parallel training across many GPUs, by abstracting the all-reduce communications.
	•	\textbf{PyTorch Distributed Data Parallel (DDP)} is now a built-in way to do data parallel training in PyTorch efficiently with minimal code changes.
	•	\textbf{TensorFlow’s Distribution strategies} similarly handle multi-GPU or multi-node training.

However, as model parallelism and memory sharding became more important, more specialized libraries emerged:
	•	\textbf{Mesh TensorFlow} (Google) and \textbf{Megatron-LM} (NVIDIA) provided patterns to split transformer models across many GPUs, handling the details of splitting matrices for multi-GPU operations (like splitting the heads of multi-head attention across GPUs, etc.).
	•	\textbf{DeepSpeed} (Microsoft) \cite{DeepSpeed} and \textbf{FairScale} (Facebook) introduced the concept of \textbf{ZeRO (Zero Redundancy Optimizer)} and \textbf{Fully Sharded Data Parallel (FSDP)}. The idea of ZeRO is to reduce memory usage in data parallelism by not replicating all the training states on each GPU. In normal data parallelism, if you have $K$ GPUs, you have $K$ copies of the model and optimizer states (gradients, momentum, etc.), which is wasteful. ZeRO partitions these states across GPUs so each GPU might only store 1/K of the gradients, 1/K of the optimizer moments, etc., while still each GPU has the full model for forward/backward. This sharding allows, say, 8 GPUs each storing 1/8 of the optimizer states, thus collectively handling a model 8x larger than one GPU could with the same memory. DeepSpeed and similar systems implement this transparently, along with offloading to CPU memory or NVMe for parts of the model not in active use, gradient checkpointing (trading compute for memory by not storing some intermediates), etc.
	•	\textbf{Pipeline parallel frameworks}: DeepSpeed and others also allow defining pipeline stages easily and manage the scheduling (like the 1F1B algorithm for efficient pipeline utilization). Pipeline parallelism slices the mini-batch into micro-batches and overlaps the computation of different micro-batches on different stages, to keep all GPUs busy most of the time.

Through these tools, researchers trained models with tens or hundreds of billions of parameters. For example, the 175B GPT-3 was trained using model parallelism (sharding each matrix across multiple GPUs) combined with data parallel across many nodes. The specifics: they might have used 8-way model parallel per model, and 128-way data parallel (just as an illustrative breakdown). The end result is as if training one giant model on one giant “virtual GPU” that is an aggregate of 1024 physical GPUs.

Another aspect of distributed training is communication and synchronization. There is an overhead when GPUs sync gradients or exchange activations. Techniques like \textbf{gradient compression} or \textbf{lazy communication} can reduce overhead (e.g., quantize gradients before sending, or overlap communication with computation). Also, using faster network hardware (InfiniBand, NVLink, NVSwitch) is crucial for keeping scaling efficient. At large scale, one can measure how training speed scales with number of GPUs: ideally linear (100 GPUs = 100x speed of 1 GPU), but often sub-linear due to overhead.

Finally, training large models often requires careful \textbf{learning rate scheduling and batch size tuning}. With data parallel, if you increase total batch size (say 32 GPUs each with batch 32, so total batch 1024), you often need to adjust the learning rate (linear scaling rule: multiply LR by number of GPUs, sometimes with a small warmup). Too large a batch can hurt generalization, so there’s active research on how far you can scale batch size without losing accuracy, or how to adjust optimization hyperparameters accordingly.

In short, distributed training is the backbone of modern deep learning at scale:
\begin{itemize}
\item It enables the use of many processors to train a single model faster or to train models that are too large for one processor.
\item Data parallelism is easiest and widely used; model parallelism is necessary for the largest models.
\item Sophisticated frameworks like DeepSpeed combine multiple parallelism strategies and memory optimization to maximize the model size that can be trained on a given hardware cluster.
\item For someone building very large models, understanding and utilizing distributed training techniques is as important as the model architecture itself.
\end{itemize}

As an example relatable to our students: training a ResNet-50 on ImageNet in a few hours uses 8 GPUs with data parallelism. Training GPT-3 required on the order of 1024 GPUs for over a month. Without distribution, the latter would take centuries on a single GPU! So distributed training really unlocks the ability to do things that are otherwise practically impossible.

\section{Environmental and Computational Trade-offs}
Scaling up models and training comes with significant \textbf{computational and environmental costs}. It’s important to recognize these trade-offs as we push the boundaries of model size and performance.

On the computational side, large models require enormous amounts of processing. Training a state-of-the-art model can cost millions of dollars in cloud compute or require specialized hardware setups only available to a few organizations. For example, it was estimated that training the GPT-3 model (175B parameters) took tens of thousands of GPU-hours; one estimate put the cloud compute cost at around \$4-5 million for a single training run of GPT-3. As another data point, training PaLM (540B parameters) likely used even more compute (Google hasn’t disclosed exact cost, but one can imagine it’s higher). This puts such efforts out of reach for most academic labs and startups, raising concerns about the democratization of AI research. If only giant tech companies can afford to train the most powerful models, progress could become more siloed. However, the open-source community and some academic consortiums are working on reproducing large models collaboratively (e.g., the EleutherAI group reproduced GPT-like models at smaller scales).

Environmentally, the energy consumption and resulting carbon footprint of large-scale training runs is substantial. A well-cited study by Strubell et al. (2019) \cite{Strubell2019} highlighted that training a big NLP model with hyperparameter tuning could emit on the order of \textbf{hundreds of thousands of pounds of CO$_2$}. Specifically, one experiment they analyzed (a Transformer trained with neural architecture search) was estimated to emit ~626,000 pounds of $CO_2$ (approximately 284 metric tons), which they noted is roughly five times the lifetime emissions of an average car. Even training a single large model without extensive tuning can use as much electricity as several households would use in a year. And these numbers have likely grown with the size of models in 2020-2023.

These environmental costs have sparked a movement towards \textbf{Green AI}, which calls for more focus on efficiency and for reporting the compute/energy used in research publications. Researchers are encouraged to consider the computational cost vs. benefit of model improvements. For example, if a new model achieves 1\% higher accuracy but requires 10x more computation, is it worth it? Could we find a more efficient way to get that improvement?

There’s also the perspective of \textbf{diminishing returns}. Often, scaling up yields smaller and smaller improvements: the first 10x increase in model size might give a huge jump in performance, but the next 10x might only give a marginal gain. At some point, the gain might not justify the cost. For instance, going from a 1B to a 10B parameter model might yield a big boost, but going from 100B to 1T might yield relatively less new capability (unless new emergent behaviors appear, which is uncertain). Understanding where these inflection points are is important for decision-making. The Chinchilla result we discussed is an example where bigger was not better because resources were misallocated; it showed a way to be more compute-efficient by balancing data and model size.

Another trade-off is \textbf{inference cost}. A model that’s huge not only costs a lot to train, but also to deploy: running GPT-3 (175B) for a single user query can require multiple GPU seconds of compute. If you have millions of users, that quickly becomes untenable. This is why companies often distill or compress large models for deployment, or why they invest in super-efficient serving infrastructure. There’s a direct cost (in electricity and hardware wear) for each inference as well.

However, we also see that investing in a large model can sometimes reduce costs in other ways: for example, a powerful model might handle many tasks (reducing the need to train separate specialized models for each task). There’s a notion of \textbf{model reuse} and \textbf{foundation models} – train one giant model and use it for many purposes. The computational cost is front-loaded in training, and then you get a general model. This could be efficient at a societal level if managed well (instead of everyone training their own medium model from scratch, they fine-tune a shared giant model). But it also concentrates the cost at the initial training.

From an educational perspective: when you plan an AI project, you should consider if you really need the largest model, or if a smaller efficient model can solve the problem. There’s often an elegance in achieving the same result with less. A well-known phrase by Google researchers is “the best model is the one that is the most efficient while meeting the task requirements.”

Research is actively addressing these trade-offs:
	•	Techniques like model pruning, quantization, and distillation (discussed in the next section) aim to reduce model size and compute while preserving performance.
	•	Algorithms like gradient checkpointing can reduce memory (thus enabling fewer GPUs to train a model, albeit with more computation).
	•	New training methods (like retrieval-based training that doesn’t have to store everything in weights, or using smaller models with external memory) might bypass the need for gargantuan monolithic networks.
	•	Use of renewable energy and more efficient hardware can mitigate the carbon footprint. For example, some companies schedule training for times when renewable electricity is abundant, or site their data centers in regions with clean energy.
	•	Finally, there’s interest in algorithms that could make use of large models more sample-efficiently or allow training to converge faster so we don’t waste as much energy on trial-and-error.

In summary, while model scaling has delivered incredible results, it comes with heavy computational and environmental costs:
\begin{itemize}
\item Training large models consumes a lot of electricity and hardware resources, sometimes only accessible to large institutions.
\item There is a carbon footprint concern; it’s important to strive for efficiency and consider the environmental impact.
\item Researchers must weigh if the accuracy gains justify the resource usage, and seek innovative ways to get more out of less.
\item There is a responsibility in the AI community to pursue \textbf{sustainable AI} practices, making sure that as we push for higher performance, we also push for higher efficiency.
\end{itemize}
Ultimately, the goal is to keep advancing AI in a way that is not only effective but also sustainable and broadly accessible.

\section{AI Accelerators and Inference-Time Optimization}
The final piece of the model scaling puzzle involves the \textbf{hardware} on which models run and the techniques to optimize models for deployment. As models became larger and more computationally demanding, specialized hardware known as \textbf{AI accelerators} have been developed to handle the heavy math operations much faster than general CPUs. Simultaneously, to make these models practical in real-world applications, engineers apply various \textbf{inference-time optimizations} to reduce latency and resource usage when serving models to users.

\subsection{AI Accelerators: GPUs, TPUs, and more}
In the early days, neural networks were trained on CPUs, but as soon as networks grew a bit, training became painfully slow. The big shift came when people started using \textbf{Graphics Processing Units (GPUs)} for neural network training around the mid-2000s. GPUs are designed to do many simple operations in parallel (originally for rendering graphics), which happens to be very similar to the large matrix and vector operations in neural networks. Libraries like CUDA and cuDNN (NVIDIA’s deep neural network library) made it easier to run backpropagation on GPUs, leading to massive speedups. A single modern GPU can be tens or hundreds of times faster than a CPU for training a neural net.

As deep learning took off, GPU manufacturers like NVIDIA started optimizing their hardware specifically for AI. For example, they introduced \textbf{Tensor Cores} (first in the Volta architecture, around 2017) which are units that perform matrix multiply–accumulate operations very fast at lower precision (like FP16 or BF16). These cores are tailored for the kind of computations in training deep nets, and using them can provide another 5-10x speed boost compared to using normal GPU cores, albeit requiring using mixed-precision training (which is now standard; it trains faster and uses less memory, with no loss in model quality in most cases).

Meanwhile, Google developed the \textbf{Tensor Processing Unit (TPU)} \cite{Jouppi2017}, an ASIC (application-specific integrated circuit) specifically for neural network workloads. TPUs were deployed in Google’s datacenters starting mid-2010s. They excel at matrix operations and are used both for training (Cloud TPU) and for inference (Edge TPU for smaller devices, for instance). One advantage of TPUs is that they can be built into very large pods with fast interconnect, allowing Google to train very large models (like their 11B parameter T5 in 2019, and more recently PaLM 540B were trained on TPU v4 pods).

Other companies/efforts have produced accelerators:
	•	\textbf{FPGAs} (Field Programmable Gate Arrays) can be configured to run neural nets efficiently and are sometimes used in low-latency environments (like high-frequency trading with AI, because FPGAs can get very low latency).
	•	\textbf{ASICs from startups}: There are Graphcore’s IPU, Cerebras’s Wafer-Scale Engine (which is essentially a huge chip containing many cores for deep learning), Habana Labs’ Gaudi (now owned by Intel), etc. These often promise either more speed, better energy efficiency, or memory advantages.
	•	\textbf{Neuromorphic chips}: Though not mainstream for deep learning, some research chips model brain-like spiking neural nets (IBM TrueNorth, Intel Loihi) which are very power efficient for certain tasks, but they’re a bit outside the typical deep learning deployment.

Even at smaller scale, modern smartphones come with \textbf{NPUs} or DSPs optimized for AI. Apple’s A-series chips have a “Neural Engine”, Qualcomm has the Hexagon DSP that accelerates neural nets, etc. They enable running moderately complex models on-device in real time (think of face recognition in the camera app, or voice assistants). These mobile accelerators are why we can have things like real-time video filters or AR effects using neural nets on a phone without killing the battery immediately.

In essence, hardware has co-evolved with model scaling: bigger models needed better hardware, and better hardware enabled even bigger models. Without GPUs and TPUs, we simply could not have trained models like ResNet-152 or GPT-3 in any reasonable timeframe.

\subsection{Inference-Time Optimization Techniques}
Once a model is trained, using it effectively is crucial. A model that is too slow or too large to deploy is of little practical use, no matter how accurate. Therefore, a lot of effort goes into making models lean and fast during inference, often via \textbf{model compression} and \textbf{optimized implementations}.

Here are some key techniques:
\begin{itemize}
\item \textbf{Pruning:} Neural network pruning involves removing weights or neurons that are not important. Early work (Han et al., 2015) on the “lottery ticket hypothesis” and pruning showed that you can zero-out a large fraction of weights in a trained network (e.g., 90\% of them) and fine-tune the model to recover accuracy, resulting in a much sparser model. If implemented correctly on hardware (sparse matrix multiplication), a pruned model can run faster and use less memory. There are different granularity: one can prune individual connections, entire neurons/filters, or even whole layers (if you determine they aren’t needed). Structured pruning (removing whole filters) has the advantage that it results in a smaller dense model that is directly faster on standard hardware. Unstructured pruning (removing arbitrary weights) yields a sparse model that might require specialized support to see speed gains.
\item \textbf{Quantization:} This is about reducing the numeric precision of the model’s parameters and computations. Instead of 32-bit floats, we might use 8-bit integers to represent weights and activations. That immediately gives 4x reduction in memory and can vastly speed up compute on hardware that supports integer arithmetic (which is most hardware). Quantization can be done post-hoc (post-training quantization), sometimes with a slight accuracy drop, or during training (quantization-aware training) where you simulate low precision during training so the model learns to be robust to it. Many CNNs can be quantized to int8 with minimal loss in accuracy. For example, an int8 quantized inference of ResNet-50 can be 2-3x faster than float16 inference on CPUs. Even more extreme, there are research works on 4-bit and 2-bit networks, or even binary neural networks (1-bit weights). Those often see larger accuracy hits, but for some applications they can work. Modern toolkit: TensorFlow Lite, PyTorch Mobile, etc., all have quantization support to help deploy models in low precision.
\item \textbf{Knowledge Distillation:} This technique, introduced by Hinton et al. (2015) \cite{HintonDistillation}, involves training a smaller \textbf{student} model to replicate the behavior of a larger \textbf{teacher} model. During training, instead of (or in addition to) the ground truth labels, the student is trained on the teacher model’s output probabilities (the “soft targets”). These soft targets contain richer information than hard labels (they tell you not just the correct class but also that e.g. class A was 0.2 probability, class B 0.1, etc., which encodes some of the teacher’s knowledge about similarities). By imitating the teacher, the student can often achieve a level of performance that’s surprisingly close to the teacher, despite having far fewer parameters. Distillation has been used in e.g. compressing BERT-like models: the original BERT base had 110M parameters, and people distilled it down to a “TinyBERT” or “DistilBERT” with around 40M or less, with maybe only a small drop in accuracy on downstream tasks. Distillation can also improve the student’s generalization by transferring teacher’s general knowledge. It’s a powerful method to get a production-ready model after doing research with a giant model.
\item \textbf{Efficient Libraries and Operator Fusions:} On the deployment side, frameworks like \textbf{TensorRT} (NVIDIA) or \textbf{ONNX Runtime} optimize the computation graph of the model. They can fuse multiple operations into one for efficiency (e.g., combine a convolution, bias addition, and activation into a single GPU kernel), they choose fast algorithms for conv or matmul based on the hardware, and apply optimizations like removing redundancies or constant folding. These compilers can give big speedups without changing the model architecture at all. Similarly, for mobile, \textbf{Core ML} for iOS or \textbf{NNAPI} for Android take a model and run it on the phone’s CPU/GPU/NPU with optimizations.
\item \textbf{Batching and Parallelization:} If you have to run inference on many inputs (like an ML service handling thousands of requests per second), you can often batch multiple inputs together to make use of the hardware more efficiently. GPUs, for instance, are great at throughput if you provide a large batch, although that introduces some latency. Balancing batch size for throughput vs. latency is an engineering concern. But for offline inference (processing large datasets), batching can drastically reduce total compute time by amortizing overheads.
\item \textbf{Dynamic Inference and Cascades:} Another idea is not all inputs require the full model complexity. For instance, one can use a small model to quickly route or filter easy cases, and only send harder cases to a bigger model (cascaded models or early-exit models). Some networks are designed with early exit points: if the model is confident at an intermediate layer, it can output early without executing the rest of the layers, saving compute.
\item \textbf{Hardware-specific tricks:} On some hardware, you might leverage specific instructions. For example, ARM CPUs have NEON vector instructions for mobile nets, or use of Winograd algorithms for convolution (trading off some precision for speed), etc. These are lower-level but collectively can be significant.
\end{itemize}

AI accelerators and these optimizations are why, despite the explosion in model sizes, we are still able to use AI in real time in many cases. For example, running a large Transformer for text prediction on a server might be slow if done naively, but with quantization and a GPU or TPU, you can get responses in fractions of a second, enabling interactive AI like chatbots or translation services.

It’s a continuous battle though: as models get bigger (e.g., from 1B to 10B to 100B parameters in NLP), each step forces new innovations in compression or serving. For instance, there’s work on quantizing transformers to 8-bit or even 4-bit to fit them on a single GPU for inference, or using distillation to make “portable” versions of GPT-3 that are only few billions of parameters.

In hardware terms, what’s also interesting is how these accelerators have changed the research: Knowing that you have hardware that excels at certain operations might bias architecture choices. For instance, the vision community moved away from certain operations that weren’t as GPU-friendly (like unpooling or certain normalization across batch at test time) and more towards ones that map well to accelerators (matrix multiplies, etc.). There’s also specialized hardware like content-addressable memory for fast retrieval that could integrate with neural nets in future.

To summarize this section:
\begin{itemize}
\item Specialized hardware (GPUs, TPUs, NPUs, etc.) has been critical to training and using scaled models. Without GPUs, deep learning might have hit a wall much earlier.
\item Inference-time optimizations ensure that models can be deployed in practice: through pruning, quantization, distillation, efficient runtimes, etc., we shrink and speed up models to meet real-world constraints.
\item These techniques often enable a model that was trained large to be used in a smaller footprint, bridging the gap between cutting-edge research models and production systems (like compressing a 1GB model to 100MB and making it run on a phone).
\item The co-evolution of models, algorithms, and hardware is a hallmark of deep learning progress. As a practitioner, awareness of these tools is important to maximize the impact of any model you build.
\end{itemize}

\section{Summary and Outlook}
In this lecture, we’ve taken a grand tour of model scaling in deep learning, from the early days of simply making networks deeper, to the modern era of billion-parameter models and everything in between. The key themes can be summarized as follows:

\begin{itemize}
\item \textbf{Bigger models tend to perform better}: We’ve seen numerous examples (VGG, ResNet, GPT-3, etc.) where increasing the size of the model (depth, width, or total parameters) led to leaps in performance on benchmarks. This has been a consistent trend across vision and NLP.
\item \textbf{But naive scaling isn’t enough}: Just piling on layers or parameters can fail without the right architecture and training methods. Innovations like residual connections and normalization were crucial to enable effective depth scaling. Efficient architectures (Inception, DenseNet, MobileNet) made it possible to use parameters more effectively. Neural architecture search and compound scaling gave systematic ways to scale models without wasting computation.
\item \textbf{Scaling is multi-dimensional}: We discussed depth, width, input size, and also data and compute. Truly scaling a solution might mean more training data or more inference steps, not just more layers. The success of Chinchilla emphasized balancing model size and dataset size. The idea of test-time compute scaling opened another axis: reasoning steps.
\item \textbf{Engineering and hardware matter}: Distributed training on many GPUs/TPUs made it possible to even consider training huge models. AI-specific hardware accelerated both training and inference. Model compression and optimization techniques ensure that these large models can actually be used in practice and are not just academic curiosities.
\item \textbf{Efficiency and optimization are key for the future}: As we reach toward ever larger models, the cost and environmental impact become serious considerations. The community is responding with techniques for efficiency (both algorithmic and hardware-level). There’s an increasing focus on getting more out of each parameter (e.g., through better architecture or reuse like MoE models) and each FLOP (through quantization, better algorithms).
\end{itemize}

What might the future hold for model scaling? On one hand, we might continue to see growth in model size, especially as organizations compete to build more capable general AI systems. It’s possible we’ll see trillion-parameter models become more common (some already exist in sparse form). With improved algorithms, those could be trained efficiently (for example, using mixture-of-experts to activate only parts of the model per input, so not all trillion parameters are used every time).

We may also discover new \textbf{paradigms that break the current scaling mold}. For instance, there’s interest in \textbf{sparsity} and \textbf{conditional computation}: instead of a monolithic dense model, have a very large network but only a small subset is used for any given task or input (expert networks, dynamic routing, etc.). This way, capacity grows without linear growth in computation. Another direction is \textbf{neurosymbolic or hybrid systems} that combine neural networks with explicit reasoning or search components, which ties into our discussion on test-time compute. These could achieve better performance without needing exponentially more parameters.

The role of \textbf{data} is also becoming more prominent. If truly massive models are under-trained, one bottleneck might be high-quality data. We may see efforts to create larger and more diverse training corpora, or synthetic data generation to feed these models.

From a research perspective, a fascinating question is: \emph{How far can scaling take us}? Some in the field (inspired by results like GPT-3) suspect that simply making models bigger and training on more data will eventually yield very powerful general AI. Others believe algorithmic advances will be needed beyond a point, because some aspects of intelligence might not emerge just from scale. The likely reality is a combination: scaling will continue to produce gains, but we’ll also augment models with new ideas to make them more efficient and robust.

For you, as future practitioners or researchers, understanding model scaling gives you a powerful lens. If your model isn’t performing well enough, consider if making it larger or giving it more data might help, and know the techniques to do so effectively (like adding normalization or using a known scalable architecture). Conversely, if you need to deploy a model, know how to compress and optimize it, and consider if you can achieve the same with a smaller model for that context.

In conclusion, model scaling has been a driving force in the progress of deep learning. We’ve gone from relatively shallow nets that could recognize handwritten digits, to networks tens of layers deep mastering ImageNet, to networks so large they can write coherent essays or have conversational ability. Each jump required not just more computation, but also ingenuity in design to make that computation count. As we continue to scale, issues of efficiency, cost, and sustainability will be paramount, but so will the potential for truly remarkable AI capabilities. The journey of scaling is not just about making things bigger; it’s about learning how to \emph{grow} our models wisely and responsibly to reach new heights of performance.
\chapter{Model Compression and Cost Optimization}

\section{Introduction}
As AI models become increasingly powerful, their computational and storage costs have grown dramatically. For example, the language model GPT-3 has 175 billion parameters, requiring hundreds of gigabytes of memory and massive compute power to train and run \cite{Brown2020GPT3}. Such large models are challenging to deploy on everyday devices or within limited data centers. Model compression and cost optimization techniques address this challenge by making models smaller, faster, and more efficient, without significantly sacrificing accuracy. These methods are crucial for:
\begin{itemize}
    \item \textbf{Deployment on edge devices:} Reducing model size and energy use so that AI can run on smartphones, IoT sensors, and embedded hardware.
    \item \textbf{Faster inference:} Decreasing the latency and increasing the throughput of model predictions, which is vital for real-time applications.
    \item \textbf{Lower resource costs:} Cutting down memory footprint and CPU/GPU usage, thus saving cloud computing costs and energy consumption in data centers.
    \item \textbf{Continued innovation:} Enabling researchers and practitioners to experiment with advanced models without requiring prohibitive computational resources.
\end{itemize}

In the following sections, we will explore several key techniques for model compression and efficiency. Each section introduces a concept in accessible terms and highlights why it matters, along with references to seminal works or widely-used methods in that area.

\section{Knowledge Distillation}
One fundamental approach to model compression is *knowledge distillation*. This technique involves transferring the “knowledge” from a large, cumbersome model (often called the *teacher*) to a smaller model (the *student*). The smaller student model is trained to replicate the behavior of the teacher model, typically by trying to match the teacher’s output probabilities (or feature representations) on a training set \cite{Hinton2015Distillation}. 

The idea, introduced by Hinton et al. (2015) \cite{Hinton2015Distillation}, is that the teacher’s outputs (for example, the soft probabilities it assigns to various classes) contain rich information about how the teacher generalizes. By training the student to mimic these outputs rather than the hard true labels alone, the student can learn to perform nearly as well as the teacher. This process effectively compresses the knowledge of the larger model into a more compact form. 

Knowledge distillation is important because it allows us to deploy a lightweight model that achieves performance close to a heavy-duty model. In practical terms, one might train a very large neural network or an ensemble of networks to achieve high accuracy, and then distill it into a single small network. The smaller network requires far less memory and computation, making it feasible to use in production (for instance, on a mobile app or a web service) while still benefiting from the teacher’s high accuracy.

\section{Dropout and Sparse Networks}
Regularization techniques like *dropout* can also contribute indirectly to model efficiency and cost optimization. Dropout, introduced by Srivastava et al. (2014) \cite{Srivastava2014Dropout}, involves randomly “dropping out” (i.e., setting to zero) a subset of neurons during each training batch. This forces the network to not rely too heavily on any one neuron, thereby reducing overfitting. While the primary purpose of dropout is to improve generalization, one useful side effect is that it encourages the network to develop redundant, distributed representations of features. In practice, this means the network’s effective capacity is used more efficiently, and at inference time (when we typically remove dropout), we can often get away with smaller models or sparser activations without losing much accuracy.

Beyond dropout, research has shown that neural networks often contain much smaller *sparse sub-networks* that can be trained to achieve performance comparable to the full model. This is sometimes referred to as the *“lottery ticket hypothesis,”* which suggests that within a large random-initialized network, there exist sparse winning sub-networks that, when trained in isolation, can reach nearly the original accuracy \cite{Frankle2019Lottery}. This finding implies that many weights in large models are redundant. If we can identify and extract these efficient sub-networks, we have an opportunity to drastically reduce model size and computation.

In summary, techniques that promote sparsity (whether through regularization like dropout or through explicit discovery of sparse structures) hint that we can slim down models. They lay the groundwork for methods like pruning, which we discuss next, by indicating which parts of a network are less necessary.

\section{Model Pruning}
If a significant fraction of a model’s parameters are unnecessary or redundant, we can remove them – a process known as *model pruning*. Pruning algorithms trim a neural network’s weights or neurons that contribute the least to model predictions. Early work by LeCun et al. on “Optimal Brain Damage” in the 1990s demonstrated that pruning weights based on their impact on the loss can dramatically reduce network complexity with minimal loss in accuracy. More recently, Han et al. (2015) showed that modern deep networks can be pruned by 80-90\% of their parameters while maintaining accuracy, by iteratively removing small-magnitude weights and fine-tuning the model \cite{Han2015Learning}.

By eliminating redundant connections, pruning yields a smaller model that requires less memory and computation. For example, Han et al.’s method first trains a large network, then prunes it, and finally retrains the remaining connections to fine-tune the model. The result is a much sparser network. In a follow-up work, *Deep Compression*, Han and colleagues combined pruning with quantization and even Huffman coding to compress neural networks by an order of magnitude or more \cite{Han2015Learning}. 

Pruned models are especially useful for deployment on resource-limited hardware. With far fewer active weights, these models can execute faster on CPUs and GPUs (since there is less work to do) and can be stored in smaller flash or disk storage. Pruning also benefits energy efficiency: skipping unnecessary calculations means less power is consumed. As an added benefit, some hardware and libraries can exploit sparsity by only computing the needed operations.

In practice, pruning can be applied at different levels of granularity:
\begin{itemize}
    \item \textbf{Weight pruning:} Remove individual connections (weights) that are low in magnitude.
    \item \textbf{Neuron or filter pruning:} Remove entire neurons or convolutional filters that have little impact (this results in smaller layers and can directly reduce computation in those layers).
    \item \textbf{Structured pruning:} Remove structured parts of the network (like whole channels or attention heads) so that the resulting network can still be efficiently implemented without irregular memory access patterns.
\end{itemize}
After pruning, the model is usually fine-tuned or retrained, allowing the remaining weights to adjust and sometimes recover any lost accuracy.

\section{Mixture of Experts}
A *Mixture of Experts (MoE)* is an approach that uses multiple sub-models (experts) and a gating mechanism to decide which expert(s) to use for a given input. One of the motivations for MoE is to increase model capacity significantly while keeping the computation per input relatively constant. In an MoE layer, you might have dozens or even thousands of expert subnetworks, but for each input, only a sparse selection of those experts are activated and computed \cite{Shazeer2017MOE}. 

Shazeer et al. (2017) demonstrated a sparsely-gated MoE that allowed a model to have an extremely large number of parameters (hundreds of billions) but only use a small fraction of them for each data sample, thereby not dramatically increasing the computation required for each inference \cite{Shazeer2017MOE}. The gating network learns to route each input to the most appropriate experts. Because only a few experts process any given input, the effective computation (and cost) per input remains manageable, even though the total parameter count is very high.

From a cost-optimization perspective, MoEs are appealing because they offer a way to scale model size (and therefore potential accuracy or capacity) without a proportional scaling of computational cost. This is a form of conditional computation: the model “decides” where to allocate resources for each example. In deployed systems, MoEs could be used to save computation by, for instance, only running a complex portion of a model when needed. If certain inputs are easy, the gate might route them to a simple expert, whereas only challenging inputs invoke the full capacity of larger experts.

It’s worth noting that while MoEs reduce the average computation, they introduce complexity in training and load-balancing the use of experts. However, as shown by recent large-scale implementations of MoE in natural language processing, these challenges can be managed to achieve state-of-the-art results efficiently \cite{Shazeer2017MOE}.

\section{Parameter-Efficient Fine-Tuning}
Large pre-trained models (such as BERT or other transformer models) are often fine-tuned for specific tasks. Traditional fine-tuning updates all the parameters of the model for each new task, which becomes very costly when models have hundreds of millions or billions of parameters. *Parameter-efficient fine-tuning* strategies address this by only training a small subset of parameters for each task, keeping the majority of the model unchanged.

One popular approach is to add small adapter modules to the network and only train these new parameters, leaving the original model weights fixed. Houlsby et al. (2019) introduced Adapter layers for transformers, demonstrating that by training just a few million extra parameters (the adapters) one can achieve performance close to full fine-tuning on a variety of tasks \cite{Houlsby2019}. This approach drastically reduces the number of parameters that need to be learned and stored per task. Instead of having a full copy of the model for each task, you only need to keep the tiny adapter parameters for each one.

Another method, called *LoRA (Low-Rank Adaptation)*, injects trainable low-rank matrices into each layer of the model \cite{Hu2021LoRA}. During fine-tuning, only these low-rank matrices are updated. LoRA has shown that it can match the performance of full fine-tuning while training a tiny fraction of the parameters \cite{Hu2021LoRA}. This again means huge memory savings, especially when one model must be fine-tuned to many tasks.

Parameter-efficient fine-tuning is important for multi-task and continual learning scenarios. If you have a single big model that needs to serve dozens of different tasks or domains, using these techniques allows you to adapt to each task without the overhead of a full model copy. It also often speeds up training (fewer parameters to update) and reduces the risk of catastrophic forgetting by limiting how much of the model is altered for each new task.

\section{Hardware Optimization}
Optimizing AI models isn’t only about the algorithms themselves; it’s also about how they run on hardware. AI practitioners often design models with an awareness of the target hardware to achieve better speed and efficiency. This includes leveraging specialized hardware (like GPUs, TPUs, FPGAs, and AI accelerators) and optimizing how computations are carried out on these devices.

Specialized hardware can vastly accelerate certain operations. For instance, Google’s Tensor Processing Unit (TPU) is an ASIC (application-specific integrated circuit) custom-built for neural network operations. The first-generation TPU focused on accelerating inference with 8-bit integer arithmetic and achieved an order-of-magnitude higher performance per watt than general-purpose CPUs for neural net workloads \cite{Jouppi2017TPU}. The key was designing the chip to multiply matrices and vectors (common in neural nets) very quickly, and using lower-precision arithmetic to save on silicon area and power \cite{Jouppi2017TPU}. This example shows that by tailoring hardware to the properties of neural networks, we can drastically cut cost and energy for running AI models.

Even on existing hardware like CPUs/GPUs, there are many low-level optimizations:
\begin{itemize}
    \item Using vectorized operations and tensor cores: Modern CPUs have SIMD instructions, and GPUs have tensor cores (as on NVIDIA devices) that can multiply small matrices very fast. Ensuring your model’s computations line up with these units (for example, using matrix sizes that are multiples of certain values) can improve efficiency.
    \item Memory access patterns: Structuring the model to reuse data in caches and avoid unnecessary data transfer can speed up inference. For example, merging layers or fusing operations means the processor can do more work while data is already loaded, instead of reading and writing intermediate results repeatedly.
    \item Pipeline parallelism and batching: When running many inferences, using batch processing can amortize overheads. If the use case allows, processing inputs in batches can keep hardware utilization high and reduce per-sample cost.
\end{itemize}

In summary, hardware optimization involves understanding and exploiting the strengths of the target device. Sometimes it even involves designing new hardware for the demands of AI. The end goal is the same as other compression techniques: achieve the desired model performance with as little computational cost and energy usage as possible.

\section{Number Representation}
The way numbers are represented and processed in a model can have a big impact on efficiency. Deep learning models traditionally use 32-bit floating-point numbers (float32) for weights, activations, and gradients. However, not all those 32 bits of precision are necessary for neural network computations. Using lower-precision representations can significantly speed up computation and reduce memory usage.

One straightforward step is to use 16-bit floats (float16 or bfloat16) instead of 32-bit floats. Many hardware accelerators support 16-bit arithmetic which is typically twice as fast and uses half the memory. There is evidence that neural networks can be trained successfully with 16-bit precision if done carefully \cite{Gupta2015Precision}. Gupta et al. (2015) showed that using 16-bit fixed-point numbers with stochastic rounding can train deep networks to nearly the same accuracy as 32-bit floats \cite{Gupta2015Precision}. This kind of result suggests a lot of those extra bits in float32 aren’t crucial for the learning process.

There are also even more aggressive reductions:
\begin{itemize}
    \item 8-bit integers (INT8) for inference: After training a model in float32, one can convert weights (and even activations) to 8-bit integers for inference. This shrinks the model size by 4x and often allows using fast integer math pipelines in hardware.
    \item 1-bit or 2-bit (binary/ternary networks): In research, some extreme approaches train networks with binary weights or activations (1-bit values) \emph{[not a specific citation given here, but see e.g. Hubara et al., 2016 on binary nets]}. These networks trade a lot of precision for massive efficiency gains, though often with a drop in accuracy.
\end{itemize}

Changing number representation can affect the model’s accuracy if not done well, because lower precision introduces quantization error (rounding error). The challenge is to keep the model robust to this loss of precision. Often, techniques like scaling (adjusting the range of values to use the limited bits effectively) or calibration are used to mitigate precision loss. The next section on mixed precision training discusses how one can even train with lower precision.

\section{Mixed Precision Training}
Mixed precision training is a technique that uses high-precision and low-precision numbers in combination to speed up training. The typical recipe is to keep a master copy of weights in 32-bit precision for accuracy, but use 16-bit (half precision) for forward and backward computations. By using 16-bit floats for most operations, memory usage and bandwidth are roughly halved, and arithmetic operations can be much faster on hardware that supports fast half-precision math \cite{Micikevicius2018Mixed}.

Micikevicius et al. (2018) demonstrated that mixed precision training can train a variety of neural networks without loss of accuracy, given some care (such as scaling the loss to avoid underflow in gradients) \cite{Micikevicius2018Mixed}. The approach typically looks like this:
\begin{itemize}
    \item Maintain weights in full 32-bit precision (to accumulate small updates accurately).
    \item During each training step, cast weights to 16-bit and compute forward pass and gradients in 16-bit.
    \item Compute weight updates (like gradient accumulations) in 32-bit to preserve precision, then update the 32-bit master weights.
    \item Use a technique called *loss scaling* to counteract the limited range of 16-bit floats: multiply the loss by a scale factor before backpropagation and divide gradients by the same factor later. This prevents tiny gradient values from becoming zero in half precision.
\end{itemize}

By doing this, one can often get nearly a 2x speedup in training and reduce memory usage, all while maintaining model quality \cite{Micikevicius2018Mixed}. Mixed precision has quickly become a standard in training large models because it provides significant efficiency gains with very little downside.

In summary, mixed precision training exploits the insight that not all calculations need full 32-bit precision. Many parts of neural network computation are tolerant to lower precision, as long as critical parts (like weight updates) retain accuracy. This technique directly cuts down training time and resource usage, enabling us to train bigger models or train models faster on existing hardware.

\section{Quantization}
Quantization is the process of converting a neural network’s parameters and computations from high precision (e.g., 32-bit float) to lower precision (e.g., 8-bit integer). Unlike the mixed precision approach which still kept some high precision around, quantization often focuses on making a model purely low-bit for inference after it has been trained. The main goal is efficiency: an 8-bit model uses a quarter of the memory of a 32-bit model, and integer arithmetic can be executed much faster on many processors (especially those with specialized DSP or vector units).

A simple form is *post-training quantization*: after training a model with full precision, you convert the weights to 8-bit. In many cases, you also quantize activations (the intermediate results) to 8-bit as the model runs. Modern libraries and frameworks have support for this, often with minimal changes needed from the user. The challenge is ensuring the quantized model still performs well. Some networks can lose accuracy when naively quantized because the 8-bit approximation of their weights/activations is not exact.

One influential work by Jacob et al. (2018) described a quantization scheme used for efficient inference on mobile CPUs, where both weights and activations are int8, and the model achieves nearly the same accuracy as the float version \cite{Jacob2018Quantization}. The approach involves choosing appropriate scaling factors for each layer so that 8-bit values cover the range of the 32-bit tensors as effectively as possible. For example, if a layer’s weights range from -2.5 to 2.5 in float32, you might map that to the -128 to 127 range of int8. During inference, operations are carried out in integer math, and results are scaled back to normal ranges at the end.

Quantization can provide tremendous speed-ups. Many CPUs and accelerators have special instructions for 8-bit matrix multiplication which runs much faster than 32-bit math. Also, the reduced memory bandwidth is a big win: reading 8-bit values from memory is 4 times faster than reading 32-bit values, which often is a bottleneck. Thus, quantization not only shrinks model size on disk, but often yields real-time speed improvements and energy savings during inference.

However, aggressive quantization (like going below 8 bits) can hurt accuracy significantly if not done carefully. That’s where techniques like quantization-aware training come in.

\section{Quantization-Aware Training}
Quantization-aware training (QAT) is a strategy to prepare a model during the training phase to handle quantization, so that when you later convert it to a lower bit width, it still performs well. Instead of training the model in full precision and then hoping it works with int8, QAT actually simulates the effects of quantization (usually 8-bit) during training. In each training iteration, the weights and activations are “fake-quantized” — that is, rounded to the nearest value that would exist in int8 — so that the forward and backward pass experience the quantization errors. By doing this, the model can adjust its parameters to compensate for any loss in precision.

Jacob et al.’s paper on integer inference also touches on the training procedure for quantization \cite{Jacob2018Quantization}. Additionally, a whitepaper by Krishnamoorthi (2018) provides an overview of techniques for quantizing networks and highlights the importance of incorporating quantization in the training loop to maintain accuracy \cite{Krishnamoorthi2018}. The process of QAT typically involves:
\begin{itemize}
    \item Starting with a pre-trained float model (or training from scratch with quantization awareness).
    \item During each forward pass, before computing each layer’s output, insert a simulated quantization step (rounding to 8-bit representation).
    \item Compute the loss as usual and do backpropagation. The gradients flow through these simulated quantization operations (often using a straight-through estimator for the non-differentiable rounding).
    \item The model learns to tolerate or correct for the quantization. For instance, weight values might shift slightly to better align with representable 8-bit numbers.
\end{itemize}

After such training, when the weights are finally quantized and the model is deployed in 8-bit, the accuracy is typically much closer to the original full-precision model compared to naive post-training quantization. This is especially useful for networks that are sensitive to quantization (e.g., those with very small or very large weight distributions, or very tight tolerance in certain layers like last-layer classifiers).

Quantization-aware training requires more effort during the training phase and sometimes a specialized training pipeline, but it pays off by enabling the efficiency of quantized models without sacrificing predictive performance. It’s a critical technique for pushing models to low bit-widths in scenarios where every bit and every joule of energy matters (like on-device speech recognition or vision on smartphones).

\section{Deployment Frameworks}
Finally, to put all these techniques into practice, it helps to use deployment frameworks and tools designed for model compression and optimization. These are software frameworks or libraries that take a trained model and produce an optimized version for a target platform. They often implement many of the techniques discussed above under the hood.

Some notable examples include:
\begin{itemize}
    \item TensorFlow Lite – a framework by Google for deploying neural networks on mobile and IoT devices. It can convert models into a special efficient format and supports optimizations like quantization and pruning.
    \item ONNX Runtime – an inference engine for the Open Neural Network Exchange (ONNX) format that can perform optimizations and target different hardware (from mobile CPUs to GPUs).
    \item PyTorch Mobile – a set of tools within PyTorch to package models for mobile, including support for quantized models.
    \item TensorRT – NVIDIA’s optimization toolkit that takes in neural network models and produces highly optimized inference code for NVIDIA GPUs (doing transformations like layer fusion, precision lowering, etc.).
    \item Apache TVM – an open source deep learning compiler that automatically optimizes models for a given hardware target \cite{Chen2018TVM}. TVM treats the model computation graph like a program and applies compiler optimizations to generate fast low-level code for CPUs, GPUs, or specialized accelerators.
\end{itemize}

Using these frameworks, developers can often get significant speed-ups without having to manually implement all optimizations. For example, TVM can autotune the implementation of each layer of a network (such as the loop order and memory layout for a convolution) to best utilize the device’s caches and cores, yielding performance close to hand-tuned kernels \cite{Chen2018TVM}. TensorFlow Lite can take a full TensorFlow model and convert it into an integer-only version that runs efficiently on ARM processors.

These deployment frameworks encapsulate a lot of engineering best practices. They ensure that when you compress a model (via quantization, pruning, etc.), the resulting model actually runs faster and uses less power on the target device. In essence, they bridge the gap between the model design and the hardware execution.

\section*{Conclusion}
Model compression and cost optimization techniques are essential for bringing the power of AI to practical applications. By distilling knowledge, pruning unneeded parts, using clever architectures like mixtures of experts, tuning only what we need, leveraging hardware capabilities, and reducing numerical precision, we can create models that are both powerful and efficient. This means AI systems can be deployed more widely — from cloud servers to tiny edge devices — and operate under real-world constraints.

The continued development of these techniques, along with supportive frameworks and hardware advances, will allow us to scale AI models further while keeping them affordable and energy-efficient. As you explore these methods, remember that the goal is to achieve the right balance between model complexity and operational efficiency for your specific task and constraints.

\chapter{Manifold Learning}

\section{Introduction to Manifold Learning}
Many real-world datasets (such as images, audio, and text) lie in high-dimensional spaces but often exhibit lower-dimensional underlying structure. We call this structure a \textbf{manifold}. Intuitively, a manifold is a smooth, curved surface on which data points reside. Learning to represent data in terms of these underlying manifolds (rather than raw, high-dimensional coordinates) can lead to powerful techniques for visualization, dimensionality reduction, and \emph{generative} modeling (creating new data samples).

Classic manifold learning methods include:
\begin{itemize}
    \item \textbf{Isomap} and \textbf{Locally Linear Embedding (LLE)} \cite{Tenenbaum2000,Roweis2000}, which aim to preserve local or global distances while unrolling the manifold into a lower-dimensional space.
    \item \textbf{t-SNE}, which preserves local neighborhoods for visualization in 2D or 3D.
    \item Linear methods like \textbf{PCA} capture variance directions but cannot handle nonlinear manifolds as effectively.
\end{itemize}
Deep learning extends these ideas, using neural networks to automatically learn complex, nonlinear embeddings that capture manifold structure. 

Throughout this chapter, we will:
\begin{enumerate}
    \item Introduce \textbf{autoencoders} and variants, which learn to encode data into lower-dimensional latent spaces and decode back to reconstructions.
    \item Connect loss functions to \textbf{probability and likelihood}, building toward \textbf{variational autoencoders (VAEs)} that allow sampling from learned manifolds.
    \item Explore \textbf{GANs} (Generative Adversarial Networks) and their variants for high-fidelity data generation and image-to-image translation.
    \item Present \textbf{normalizing flows} as invertible transformations with exact likelihood.
    \item Conclude with modern \textbf{diffusion models}, which iteratively denoise data from noise, achieving state-of-the-art generative performance.
\end{enumerate}

\section{Autoencoders: The First Step}
\textbf{Autoencoders} (AEs) \cite{Hinton2006} are neural networks designed to learn a compressed \emph{latent representation} of data without any labels. They consist of two parts:
\begin{itemize}
    \item \textbf{Encoder}: a function $f_\phi(\mathbf{x})$ that maps input $\mathbf{x}\in \mathbb{R}^D$ to a latent vector (or feature map) $\mathbf{h}\in \mathbb{R}^d$ with $d<D$.
    \item \textbf{Decoder}: a function $g_\theta(\mathbf{h})$ that reconstructs $\mathbf{x'}$ in $\mathbb{R}^D$ from the latent code $\mathbf{h}$.
\end{itemize}
The network is trained to minimize a \textbf{reconstruction loss} $L(\mathbf{x},\mathbf{x'})$, e.g. mean squared error (MSE):
\[
\mathcal{L}_{\text{AE}}(\phi,\theta) = \frac{1}{N}\sum_{i=1}^N \left\|\mathbf{x}^{(i)} - g_\theta(f_\phi(\mathbf{x}^{(i)}))\right\|^2.
\]
By forcing a narrow bottleneck $d<D$ (or other regularization), the network must learn the essential factors of variation, effectively discovering a manifold representation of the data.

\subsection{Why Autoencoders?}
\begin{itemize}
    \item \textbf{Dimensionality Reduction}: AEs compress data into a latent space. Like PCA, but can learn nonlinear manifolds.
    \item \textbf{Feature Extraction}: The encoder can serve as a learned feature extractor for classification or other tasks.
    \item \textbf{Denoising and Missing Data Recovery}: Some AE variants explicitly learn to remove noise or fill in missing parts.
\end{itemize}

\subsection{Simple PyTorch-like Example}
\begin{lstlisting}
encoder = Encoder(input_dim=784, hidden_dim=64)
decoder = Decoder(hidden_dim=64, output_dim=784)

for epoch in range(num_epochs):
    for x in data_loader:  # x: batch of images shaped [batch_size, 784]
        h = encoder(x)
        x_recon = decoder(h)
        loss = ((x_recon - x)**2).mean()  # MSE
        optimizer.zero_grad()
        loss.backward()
        optimizer.step()
\end{lstlisting}

After training, we can:
\begin{itemize}
    \item Extract $h=f_\phi(\mathbf{x})$ as a \emph{latent feature}.
    \item Generate reconstructions $\mathbf{x'}=g_\theta(\mathbf{h})$ for analysis or data augmentation.
\end{itemize}
However, plain autoencoders do not provide an explicit way to \emph{sample new data} from the manifold. If we sample random $\mathbf{h}$, there's no guarantee it corresponds to valid data. In the next sections, we'll see how to add probabilistic structure to fix this.

\section{Understanding Loss Functions}
A \textbf{loss function} tells the model ``how wrong it is'' and thus drives learning by backpropagation. Different choices yield different behaviors:

\subsection{L1 Loss (Mean Absolute Error)}
\[
L_1(\mathbf{x},\mathbf{x'}) = \|\mathbf{x} - \mathbf{x'}\|_1 = \sum_j |x_j - x'_j|.
\]
This is robust to outliers but can lead to sparser or blocky reconstructions.

\subsection{MSE Loss (Mean Squared Error)}
\[
L_2(\mathbf{x},\mathbf{x'}) = \|\mathbf{x} - \mathbf{x'}\|^2 = \sum_j (x_j - x'_j)^2.
\]
This heavily penalizes large errors, sometimes causing ``blurry'' reconstructions if the model averages multiple possibilities.

\subsection{Cross-Entropy Loss}
\[
-\sum_j \left[ x_j \log x'_j + (1-x_j)\log(1-x'_j) \right]
\]
Commonly used when $x_j\in \{0,1\}$ or $[0,1]$; interprets $x'_j$ as a Bernoulli parameter. Especially popular in classification or binary data reconstructions.

\subsection{Link to Probability}
Choosing, e.g., MSE corresponds to assuming Gaussian noise in reconstructions. Cross-entropy often corresponds to a Bernoulli or categorical likelihood model. This connection becomes explicit when we discuss \textbf{probability and likelihood} in generative models.

\section{Probability and Likelihood}
In statistical modeling, we treat data $\mathbf{x}$ as generated from some distribution $p(\mathbf{x})$. A generative model with parameters $\theta$ tries to match this distribution, i.e. $p_\theta(\mathbf{x}) \approx p(\mathbf{x})$. \textbf{Maximum Likelihood Estimation} (MLE) chooses $\theta$ to maximize:
\[
\log p_\theta(\mathbf{x}) \quad \text{(or the sum over all training samples).}
\]
Why? Because if $p_\theta(\mathbf{x})$ is large, the model is assigning high probability to the observed data. 

When we have a simple parametric form, $p_\theta(\mathbf{x})$, maximizing $\sum \log p_\theta(\mathbf{x}^{(i)})$ is straightforward. But with complex latent-variable models like autoencoders or neural nets, it is not always trivial to compute or differentiate $\log p_\theta(\mathbf{x})$. This leads to \textbf{variational methods} and \textbf{adversarial approaches} we’ll see later.

In practice:
\begin{itemize}
    \item A \emph{reconstruction loss} can be interpreted as the negative log-likelihood under a specific noise model (e.g., MSE $\leftrightarrow$ Gaussian).
    \item Directly modeling $p_\theta(\mathbf{x})$ is what VAEs and flows attempt; GANs implicitly learn it by generating samples, and we measure sample realism with a discriminator.
\end{itemize}

\section{Convolutional Autoencoders}
For image data, \textbf{convolutional autoencoders} \cite{Masci2011} replace dense layers with convolutions:
\begin{itemize}
    \item \textbf{Encoder}: uses conv/pooling or strides to reduce spatial size and build up feature channels.
    \item \textbf{Decoder}: uses upsampling or transposed convolutions to reconstruct the original resolution.
\end{itemize}
This architecture exploits local spatial correlation in images (edges, textures, etc.), substantially reducing parameters compared to fully connected layers on all pixels.

\begin{lstlisting}
class ConvEncoder(nn.Module):
    def __init__(self, latent_dim=128):
        super().__init__()
        self.conv1 = nn.Conv2d(3, 16, 4, stride=2, padding=1)  # 64->32
        self.conv2 = nn.Conv2d(16, 32, 4, stride=2, padding=1) # 32->16
        self.conv3 = nn.Conv2d(32, 64, 4, stride=2, padding=1) # 16->8
        self.fc = nn.Linear(8*8*64, latent_dim)

    def forward(self, x):
        x = F.relu(self.conv1(x))
        x = F.relu(self.conv2(x))
        x = F.relu(self.conv3(x))
        x = x.view(x.size(0), -1)
        z = self.fc(x)
        return z

class ConvDecoder(nn.Module):
    def __init__(self, latent_dim=128):
        super().__init__()
        self.fc = nn.Linear(latent_dim, 8*8*64)
        self.deconv1 = nn.ConvTranspose2d(64, 32, 4, 2, 1)
        self.deconv2 = nn.ConvTranspose2d(32, 16, 4, 2, 1)
        self.deconv3 = nn.ConvTranspose2d(16, 3, 4, 2, 1)

    def forward(self, z):
        x = F.relu(self.fc(z))
        x = x.view(-1, 64, 8, 8)
        x = F.relu(self.deconv1(x))
        x = F.relu(self.deconv2(x))
        x = torch.sigmoid(self.deconv3(x))
        return x
\end{lstlisting}

In training, we feed images $[batch, 3, 64, 64]$ to the encoder, get a latent $z$, and decode back to reconstruct the image. Convolutional AEs excel at:
\begin{itemize}
    \item \textbf{Image denoising or restoration}.
    \item \textbf{Learning efficient low-dimensional embeddings for vision tasks}.
\end{itemize}
But still, we have no direct way to \emph{sample} new images unless we sample $z$ from an unknown distribution. This leads us to \textbf{probabilistic autoencoders}.

\section{Denoising Autoencoders}
A \textbf{Denoising Autoencoder (DAE)} \cite{Vincent2010} explicitly trains on noisy inputs $\tilde{\mathbf{x}}$ and aims to reconstruct the clean $\mathbf{x}$. Formally:
\[
\tilde{\mathbf{x}} = \text{Corrupt}(\mathbf{x}), \quad
\mathbf{x'} = g_\theta\big(f_\phi(\tilde{\mathbf{x}})\big),
\]
and we minimize $L(\mathbf{x},\mathbf{x'})$ where $\mathbf{x'}$ tries to match the \emph{original}, uncorrupted $\mathbf{x}$.

\begin{lstlisting}
for x in data_loader:
    noise = torch.randn_like(x) * 0.1
    x_noisy = x + noise
    h = encoder(x_noisy)
    x_recon = decoder(h)
    loss = ((x_recon - x)**2).mean()
    ...
\end{lstlisting}

\subsection{Why Denoising Helps}
By forcing the network to remove noise, the autoencoder must learn a mapping that projects noisy samples back onto the data manifold. This acts like a regularizer, preventing trivial copying. Denoising AEs can:
\begin{itemize}
    \item \textbf{Recover underlying signals} in images, audio, etc.
    \item \textbf{Learn robust representations} (less sensitive to small perturbations).
    \item \textbf{Establish the foundation} for iterative denoising processes, as in \emph{diffusion models}.
\end{itemize}

\section{Variational Autoencoders (VAE)}
A major limitation of standard AEs: no well-defined \emph{latent distribution} for generating new samples. \textbf{Variational Autoencoders (VAEs)} \cite{Kingma2014} fix this by:
\begin{itemize}
    \item Assuming a latent variable $\mathbf{z}$ is drawn from a prior $p(\mathbf{z})$ (often $\mathcal{N}(0,I)$).
    \item Defining $p_\theta(\mathbf{x}|\mathbf{z})$ as the decoder distribution.
    \item Introducing an \emph{encoder/inference} network $q_\phi(\mathbf{z}|\mathbf{x})$ to approximate the true posterior $p_\theta(\mathbf{z}|\mathbf{x})$.
\end{itemize}
The training objective (ELBO) is:
\[
\mathcal{L}(\theta,\phi) = \mathbb{E}_{q_\phi(\mathbf{z}|\mathbf{x})}\big[\log p_\theta(\mathbf{x}|\mathbf{z})\big]
- D_{\mathrm{KL}}\big(q_\phi(\mathbf{z}|\mathbf{x}) \parallel p(\mathbf{z})\big).
\]
Maximizing $\mathcal{L}$ encourages:
\begin{enumerate}
    \item Good \textbf{reconstruction} ($\log p_\theta(\mathbf{x}|\mathbf{z})$).
    \item Latent codes that match the prior distribution ($D_{\mathrm{KL}}$ penalty).
\end{enumerate}

\subsection{Reparameterization Trick}
To backprop through $q_\phi(\mathbf{z}|\mathbf{x})$, we sample $\mathbf{z}$ by:
\[
\mathbf{z} = \boldsymbol{\mu}_\phi(\mathbf{x}) + \boldsymbol{\sigma}_\phi(\mathbf{x}) \odot \boldsymbol{\epsilon}, 
\quad \boldsymbol{\epsilon}\sim \mathcal{N}(0,I).
\]
This separates randomness $\epsilon$ from parameters, enabling gradient flow into $\mu,\sigma$. A simple training loop:
\begin{lstlisting}
mu, logvar = encoder(x)       # outputs of size [batch, latent_dim]
sigma = torch.exp(0.5 * logvar)
eps = torch.randn_like(mu)
z = mu + sigma * eps
x_recon_params = decoder(z)
recon_loss = ...
kl_loss = 0.5 * torch.sum(sigma**2 + mu**2 - 1.0 - logvar)
loss = recon_loss + kl_loss
\end{lstlisting}

\subsection{Why VAEs Matter}
\begin{itemize}
    \item They define a proper generative model: To sample new $\mathbf{x}$, draw $\mathbf{z}\sim \mathcal{N}(0,I)$, decode with $p_\theta(\mathbf{x}|\mathbf{z})$.
    \item They balance \emph{reconstruction quality} vs. \emph{latent space regularity}, preventing overfitting.
    \item They allow \emph{interpolation} in latent space: smoothly morph between two data points.
\end{itemize}
However, VAEs sometimes produce blurry samples due to the pixel-wise likelihood objective. We’ll see alternative approaches (GANs) or improvements (e.g., \emph{Vector Quantization}) to mitigate this.

\section{Vector Quantization and VQ-VAE}
\textbf{VQ-VAE} \cite{Oord2017} replaces continuous latents with \emph{discrete} codes drawn from a learned \textbf{codebook} $\{\mathbf{e}_1,\ldots,\mathbf{e}_K\}$. 

\begin{enumerate}
    \item Encoder outputs a continuous $\mathbf{h}$.
    \item Quantization: find the nearest code $\mathbf{e}_k$ in the codebook. The latent is $\mathbf{e}_k$ (discrete index $k$).
    \item Decoder reconstructs from $\mathbf{e}_k$.
\end{enumerate}
Gradients bypass the non-differentiable nearest-neighbor step via a \emph{straight-through} or \emph{stop-gradient} trick. The loss has:
\[
\text{Recon loss} + \|\text{sg}[\mathbf{h}] - \mathbf{e}_k\|^2 + \beta \|\mathbf{h} - \text{sg}[\mathbf{e}_k]\|^2.
\]
This \emph{commitment loss} updates both codebook vectors and encoder output. 

\textbf{Why discrete latents?}
\begin{itemize}
    \item Natural for \emph{compression} or symbolic domains (speech, text).
    \item Can train a separate discrete \emph{prior model} (like PixelCNN) over latent indices, enabling powerful generation.
    \item Often avoids \emph{posterior collapse} seen in continuous VAEs.
\end{itemize}

\section{Generative Adversarial Networks (GANs)}
\textbf{GANs} \cite{Goodfellow2014} are a different family of generative models that \emph{do not} explicitly define $p_\theta(\mathbf{x})$. Instead, they set up a \textbf{minimax game} between:
\begin{itemize}
    \item A \textbf{generator} $G(\mathbf{z})$ mapping noise $\mathbf{z}$ to fake samples $\mathbf{x'}$.
    \item A \textbf{discriminator} $D(\mathbf{x})$ outputting real vs. fake probabilities.
\end{itemize}
Objective:
\[
\min_G \max_D\, \Big[\mathbb{E}_{\mathbf{x}\sim p_{\text{data}}}[\log D(\mathbf{x})] + 
\mathbb{E}_{\mathbf{z}\sim p(\mathbf{z})}[\log(1 - D(G(\mathbf{z})))]\Big].
\]
\textbf{Discriminator} tries to distinguish real from fake. \textbf{Generator} tries to fool the discriminator. This adversarial setup can yield very sharp, realistic samples.

\subsection{GAN Training Loop (Pseudocode)}
\begin{lstlisting}
for real_data in data_loader:
    # Update D
    z = torch.randn(batch_size, latent_dim)
    fake_data = G(z).detach()
    d_loss = -( torch.log(D(real_data) + eps).mean() 
              + torch.log(1 - D(fake_data) + eps).mean() )
    optimizerD.zero_grad(); d_loss.backward(); optimizerD.step()

    # Update G
    z = torch.randn(batch_size, latent_dim)
    gen_data = G(z)
    g_loss = -torch.log(D(gen_data) + eps).mean()  # non-saturating
    optimizerG.zero_grad(); g_loss.backward(); optimizerG.step()
\end{lstlisting}

GANs can generate extremely realistic images but are \emph{trickier to train}:
\begin{itemize}
    \item \emph{Mode collapse}: generator repeatedly produces similar outputs.
    \item \emph{No explicit latent inference}: we cannot easily get $z$ for a given real $\mathbf{x}$.
    \item \emph{Hyperparameter sensitivity}: balancing $G$ and $D$ is delicate.
\end{itemize}

\section{Advanced GAN Models}
\subsection{Deep Convolutional GAN (DCGAN)}
\cite{Radford2016} Provided stable architectural guidelines:
\begin{itemize}
    \item Use \textbf{strided convolutions} in $D$ and \textbf{transposed convolutions} in $G$.
    \item Use \textbf{batch normalization} except in the final layer.
    \item Use ReLU in $G$, LeakyReLU in $D$.
\end{itemize}
DCGAN was a breakthrough for generating $64\times64$ color images with better stability than earlier fully connected GANs.

\subsection{Conditional GAN (CGAN)}
\cite{Mirza2014} Adds a condition $\mathbf{y}$ (class label, etc.) to both $G$ and $D$. This allows \emph{controlled} generation (e.g., generating a digit of a specified class). 

\subsection{Pix2Pix}
\cite{Isola2017} Uses CGAN for \emph{image-to-image translation} with paired data: $(A,B)$ pairs. The generator learns $A\to B$, and the discriminator sees $(A,B)$ vs. $(A,\hat{B})$. Adding an $L_1$ or $L_2$ term ensures overall structure is preserved, while the GAN loss provides sharp details.

\subsection{CycleGAN}
\cite{Zhu2017} Handles \emph{unpaired} translation across domains $A$ and $B$ by training two generators $G: A\to B$ and $F: B\to A$ plus two discriminators. Enforces \textbf{cycle consistency}: $F(G(a))\approx a$ and $G(F(b))\approx b$. This allows, e.g., turning horse images into zebra images without one-to-one pairs.

\subsection{StarGAN}
\cite{Choi2018} Extends multi-domain image translation in a single model. Condition the generator on a target domain label. Discriminator must classify domain and authenticity. Reduces complexity from training many pairwise models.

GANs remain popular for tasks needing crisp visuals or domain translation. But they lack a likelihood function and can be unstable. Now we turn to \textbf{normalizing flows}, which do have tractable likelihoods.

\section{Normalizing Flow}
\textbf{Normalizing flows} \cite{Dinh2017} build complex distributions by applying a sequence of \emph{invertible} transformations $f_i$ to a simple base distribution (e.g. Gaussian). If $\mathbf{z}_0\sim p_0(\mathbf{z}_0)$, we define:
\[
\mathbf{z}_K = f_K \circ \cdots \circ f_1(\mathbf{z}_0).
\]
Then the density is computed via the change of variables:
\[
p_X(\mathbf{z}_K) = p_0(\mathbf{z}_0)\,\prod_{i=1}^K \left|\det \frac{\partial f_i}{\partial \mathbf{z}_{i-1}}\right|^{-1}.
\]
With careful design (e.g. \emph{coupling layers}), $\det$ is easy to compute. Flows allow:
\begin{itemize}
    \item Exact log-likelihood evaluation.
    \item Direct sampling by drawing $\mathbf{z}_0$ and applying the transformations.
    \item Inversion to find latent codes of real data (unlike vanilla GAN).
\end{itemize}

\subsection{RealNVP and Glow}
\cite{Dinh2017} introduced \textbf{RealNVP}, using \emph{affine coupling} layers. \textbf{Glow} refined it with $1\times1$ learned convolutions. Both can generate decent images and provide tractable likelihood. However, invertible constraints can be heavy on memory and sometimes less visually sharp than GANs.

\section{Diffusion Models: From Noise to Clarity}
\textbf{Diffusion models} \cite{SohlDickstein2015,Ho2020} have emerged as top-tier generative methods, often surpassing GANs in image quality. The idea:
\begin{enumerate}
    \item A \textbf{forward process} gradually adds noise to data $x_0$ over $t=1,\dots,T$ steps, producing $x_T$ close to pure noise.
    \item A \textbf{reverse process} learns to remove noise step-by-step until it recovers an $x_0$ sample from the data distribution.
\end{enumerate}
Mathematically, $q(x_t|x_{t-1})=\mathcal{N}(x_t;\sqrt{1-\beta_t}x_{t-1},\beta_t I)$ in the forward process. The model $p_\theta(x_{t-1}|x_t)$ approximates the \emph{reverse} distribution. 

\subsection{Connection to Denoising Autoencoders}
At each step $t$, the model is effectively a \emph{denoiser} for the noisy $x_t$. Training uses a noise-prediction loss. The final generation starts from random Gaussian noise $x_T$ and iteratively applies the learned reverse steps.

\section{Denoising Diffusion Models (DDPM, DDIM)}
\subsection{DDPM (Ho et al.)}
\cite{Ho2020} formalized a simple training objective that predicts the noise added at each step:
\[
L_{\text{simple}}=\mathbb{E}_{t,x_0,\epsilon}\big[\|\epsilon - \epsilon_\theta(x_t,t)\|^2\big],
\]
where $x_t$ is a noisy version of $x_0$, and $\epsilon_\theta$ is the network output. At inference, sampling each step $t$ yields high-fidelity images but can be slow (hundreds of steps).

\subsection{DDIM (Song et al.)}
\cite{Song2021} introduced a deterministic variant and faster sampling with fewer steps. The same trained model can produce decent samples in 50--100 steps instead of 1000, bridging speed concerns.

Diffusion models excel at mode coverage and yield strikingly detailed images. They circumvent adversarial training pitfalls. They do, however, require iterative sampling.

\section{Guided Diffusion}
To control generation (e.g., class labels, text prompts), \emph{guided diffusion} modifies each reverse step based on a \emph{classifier} or \emph{conditioning vector}. 
\begin{itemize}
    \item \textbf{Classifier-based guidance}: a separate classifier $C(y|x_t)$ helps push $x_t$ toward a desired label $y$ by adjusting gradients.
    \item \textbf{Classifier-free guidance}: train the diffusion model to handle both conditioned/unconditioned inputs, then blend the two predictions at sampling time to direct the generation.
\end{itemize}
This approach underpins state-of-the-art text-to-image models (e.g. Stable Diffusion, Imagen) which produce images aligned with user prompts.

\section{Latent Diffusion Model (LDM)}
\textbf{Latent Diffusion} \cite{Rombach2022} applies diffusion in a lower-dimensional latent space rather than pixel space:
\begin{enumerate}
    \item First train an autoencoder ($E,D$) to compress images into latent codes $z = E(x)$.
    \item Perform diffusion on $z$, not on $x$, drastically reducing computation for high-res images.
    \item Decode final latent $\hat{z}$ to image $\hat{x} = D(\hat{z})$.
\end{enumerate}
This two-stage approach underlies \textbf{Stable Diffusion}: it’s far more efficient than pixel-based diffusion while maintaining image fidelity.

Applications go beyond text-to-image:
\begin{itemize}
    \item \textbf{Text-to-Video} (e.g., Make-A-Video) extends the idea to 3D/temporal latents.
    \item \textbf{Conditional tasks} (inpainting, super-resolution) become simpler with the same pipeline.
\end{itemize}

\section{Summary and Final Thoughts}
We have surveyed a broad spectrum of \textbf{manifold learning} and \textbf{generative modeling} techniques:

\begin{itemize}
    \item \textbf{Autoencoders} learn a compressed representation via reconstruction. Denoising variants push data back onto the manifold.
    \item \textbf{VAEs} add a probabilistic latent space for generation, balancing reconstructions and prior regularization.
    \item \textbf{GANs} use an adversarial game to produce very realistic samples, featuring many powerful variants for conditional or unpaired tasks.
    \item \textbf{Normalizing flows} maintain exact likelihood via invertible transforms.
    \item \textbf{Diffusion models} iteratively denoise from random noise, combining the best of stable training and high fidelity generation.
\end{itemize}

Each method tackles the core challenge of learning high-dimensional data distributions, effectively discovering the \emph{manifold} on which real data reside. In practice, the choice depends on your goals: explicit likelihood (Flows/VAEs), extreme realism (GANs/Diffusion), or fast sampling vs. interpretability trade-offs. As the field evolves, hybrid approaches are emerging (e.g., VAE-GAN combos, diffusion with normalizing flows for faster sampling). Understanding these fundamental building blocks equips you to navigate and innovate within modern generative AI research.

\chapter{Transformers}
\section{Introduction to Transformers and Motivation}
Transformers are a type of neural network architecture that have revolutionized natural language processing. They were introduced to address limitations of earlier sequence models. Before Transformers, the dominant approach for sequence-to-sequence tasks (like translating a sentence from English to French) was to use recurrent neural networks (RNNs) such as LSTMs. An encoder RNN would read the input sequence into a fixed-size vector, and a decoder RNN would generate an output sequence from that vector \cite{sutskever2014}. While effective for short sequences, this approach struggled with long sentences because the fixed vector bottleneck could not retain all information.

A major breakthrough was the \emph{attention mechanism}, first introduced in the context of RNN-based translation \cite{bahdanau2015}. Instead of compressing an entire sentence into one vector, the decoder RNN learned to attend to different parts of the input sequence at each step. This means the model could dynamically focus on relevant words (e.g., focusing on the French word for “bank” when translating the English word “bank” depending on context). Attention significantly improved translation quality by allowing the model to look back at the encoder's outputs for important information, rather than relying solely on a single context vector.

The Transformer architecture \cite{vaswani2017} took the attention idea to the extreme: it eliminated RNNs entirely and relied only on attention mechanisms (and simple feed-forward networks) to handle sequence data. Transformers have an \textbf{encoder-decoder} structure similar to previous seq2seq models, but each layer in a Transformer uses a technique called \textbf{self-attention} to look at other positions in the same sequence. This design enables much more parallelization during training (since you don't have to process tokens one-by-one as in RNNs) and can capture long-range dependencies more directly. When \cite{vaswani2017} introduced Transformers in the paper \emph{“Attention Is All You Need”}, they achieved state-of-the-art results in translation, and the model has since become the foundation for many advanced language models. In summary, Transformers were motivated by the need for models that handle long sequences with flexible, learned alignments (attention) and that can be efficiently trained with parallel computation.

\section{Attention Mechanism: Queries, Keys, Values}
At the heart of Transformers is the \textbf{attention mechanism}, often described in terms of \emph{queries}, \emph{keys}, and \emph{values}. This can be understood with an analogy: imagine a librarian (the “query”) trying to find information in a library. The library catalog entries are “keys,” and the books on the shelf are “values.” The librarian (query) compares his query to all catalog cards (keys) to see which ones are relevant, and then he pulls out the corresponding books (values) to read the needed information. In a Transformer, each position in a sequence (each word, for example) can play the role of a query, and all positions also have associated keys and values (which are learned vector representations).

Concretely, the model learns three matrices to project an input token's embedding into three vectors: a query vector $\mathbf{q}$, a key vector $\mathbf{k}$, and a value vector $\mathbf{v}$. For a given query $\mathbf{q}$ (for example, the query might represent a particular word's embedding seeking context), the attention mechanism computes a score against each key $\mathbf{k}_j$ in the sequence to measure relevance. These scores (often dot products between $\mathbf{q}$ and each $\mathbf{k}_j$) determine how much weight to give to each value $\mathbf{v}_j$ when producing the attention output.

Intuitively, you can think of the query as asking a question: “How relevant are each of the other words to me?” Each key is like a description or tag of a word, and the dot product $\mathbf{q}\cdot \mathbf{k}_j$ indicates how well the query matches that description. A high score means that word $j$ has information important to the queried word. We then take a weighted sum of all value vectors, using these scores (after normalization) as weights. The resulting vector is the new representation of the query word, enriched by information from the words it found relevant.

This mechanism allows the model to flexibly blend information from different parts of the sequence. For example, if the word “it” is the query, its query vector might match strongly with the key of another word like “bank” or “animal” depending on context, and thus pull in the value (features) of those words. The attention output for “it” will then incorporate clues from those related words, helping the model figure out what “it” refers to. All of this is done with learned vectors and is differentiable, so the model learns to set queries, keys, and values in a way that useful connections get high attention weights.

\section{Self-Attention: Contextual Interpretation and Word Disambiguation}
When we talk about \textbf{self-attention}, we mean the attention mechanism applied to a sequence of elements (like words in a sentence) where the query, keys, and values all come from the same sequence. In other words, each word is attending to other words in the \emph{same} sentence to gather context. This is a powerful idea because it lets the model dynamically decide which other words are important to understanding the current word.

One big advantage of self-attention is in resolving ambiguities and providing context. Consider the word “bank” in two different sentences:
\begin{itemize}
    \item \emph{“I went to the bank to deposit money.”}
    \item \emph{“The river bank was slippery after the rain.”}
\end{itemize}
In the first sentence, “bank” refers to a financial institution; in the second, it refers to the side of a river. How does a model know which is which? With self-attention, the representation of the word “bank” will be influenced by other words in the sentence. The query for “bank” will look at keys for words like “deposit” and “money” in the first sentence, assigning high attention weight to them, and thus the value (meaning) from those words will flow into “bank.” The model can thereby infer that “bank” in this context is something related to money. In the second sentence, the query for “bank” will focus on words like “river” or “slippery,” leading to a different interpretation (landform by a river). In essence, self-attention allows each word to \textit{adapt its representation based on the other words present}, giving the model a way to disambiguate words with multiple meanings using context.

Self-attention is also useful for understanding relationships like pronoun references. For example, in the sentence \emph{“Alice gave her keys to Bob because she trusted him,”} the pronoun “she” should be linked to “Alice” and “him” to “Bob.” A Transformer can handle this by having the query vector for “she” strongly attend to the key of “Alice,” and similarly the query for “him” attend to “Bob.” This means the model’s representation of “she” will incorporate information from “Alice” (e.g., gender or identity cues), helping it keep track of who is who. Traditional left-to-right models (like standard RNNs) would have to carry along context in a fixed-size hidden state, which can be challenging over long distances. Self-attention, by contrast, creates direct connections between relevant words, no matter how far apart they are in the sentence.

Another benefit is that self-attention is computed in parallel for all words. The model doesn’t have to process word by word sequentially; it can look at the whole sentence at once. This means it can capture long-range dependencies (like a connection between the first and last word of a sentence) without difficulty. In summary, self-attention gives Transformers the ability to read a sentence and dynamically decide, for each word, which other words to pay attention to. This leads to rich, context-dependent representations that make tasks like translation, comprehension, and summarization much more accurate.

\section{Scaled Dot-Product Attention: Formula Walkthrough with Example}
The most common implementation of attention in Transformers is called \textbf{scaled dot-product attention}. We can express it with a simple formula. Given a set of queries $Q$, keys $K$, and values $V$ (often these are matrices containing multiple query, key, and value vectors for a sequence), the attention output is:
\[ \text{Attention}(Q, K, V) = \text{softmax}\!\Big(\frac{Q K^T}{\sqrt{d_k}}\Big)\, V~. \]
Let’s break this down:
\begin{itemize}
    \item $Q K^T$: This computes the dot product between each query vector and each key vector. If $Q$ has dimension $(n \times d_k)$ and $K$ is $(m \times d_k)$ (where $n$ could be the number of query positions and $m$ the number of key positions, and $d_k$ is the dimensionality of keys/queries), then $QK^T$ is an $n \times m$ matrix of raw attention scores. The entry at position $(i,j)$ is basically $q_i \cdot k_j$, the similarity between the $i$th query and $j$th key.
    \item $\frac{1}{\sqrt{d_k}}$: We divide the dot products by the square root of the key dimension. This is a scaling factor. Without it, when $d_k$ is large, the dot products tend to have large magnitude, pushing the softmax into very small gradients (because softmax would produce extremely peaked distributions). Scaling by $\sqrt{d_k}$ keeps the variance of the dot products more normalized, which empirically leads to more stable training \cite{vaswani2017}.
    \item $\text{softmax}(\cdot)$: We apply a softmax to each row of the scaled score matrix. This turns each row of scores (for a given query) into a probability distribution over the keys. The softmax emphasizes the larger scores and diminishes the smaller ones, but ensures the weights sum to 1. So now we have attention weights.
    \item Finally, we multiply by $V$. If $V$ is an $m \times d_v$ matrix of values, the result of $\text{softmax}(QK^T/\sqrt{d_k}) V$ is an $n \times d_v$ matrix. Each output row is essentially a weighted sum of the value vectors, where the weights are those softmax attention weights.
\end{itemize}

In simpler terms, for each query $q_i$, the mechanism computes a weight $\alpha_{ij}$ for every key $k_j$:
\[ \alpha_{ij} = \frac{\exp(q_i \cdot k_j / \sqrt{d_k})}{\sum_{j'=1}^{m} \exp(q_i \cdot k_{j'} / \sqrt{d_k})}~, \] 
and then the output for query $i$ is $o_i = \sum_{j} \alpha_{ij} v_j$. Here $\alpha_{ij}$ is the attention weight (a number between 0 and 1) that says how much attention query $i$ pays to the value $v_j$.

\paragraph{Example:} Suppose we have a single query and two key-value pairs for illustration. Let the dot products (after scaling) between the query and the two keys be $[2.0,~1.0]$. Applying softmax, we get weights approximately $[0.73,~0.27]$ (since $e^{2.0}=7.39$, $e^{1.0}=2.72$, and normalized these give $7.39/(7.39+2.72)\approx0.73$, $2.72/(7.39+2.72)\approx0.27$). This means the first key is considered about three times more relevant than the second for answering the query. The attention output will be $0.73 \cdot v_1 + 0.27 \cdot v_2$. In other words, it’s mostly leaning on the information from the first value vector, but also mixing in a bit of the second. If $v_1$ represented, say, the context “Paris” and $v_2$ represented “London,” the output would be a vector that is closer to the content of “Paris.” Essentially, the mechanism decided “Paris” was more relevant to our query, and thus the resulting representation emphasizes that content.

We can outline the computation in pseudocode for clarity:

\begin{lstlisting}[language=Python]
# Given: list of queries Q, list of keys K, list of values V (aligned by index)
for each query_index i:
    for each key_index j:
        score[i][j] = dot_product(Q[i], K[j]) / sqrt(d_k)
    weights[i] = softmax(score[i])        # softmax over j
    output[i] = sum_j(weights[i][j] * V[j])
\end{lstlisting}

This process happens for every query (in practice we compute it as matrix operations for efficiency). The outcome is that each query vector $q_i$ is transformed into a new vector $o_i$ that is a blend of the values, with more weight coming from those values whose keys matched $q_i$ closely. This is the core operation that allows Transformers to route information flexibly around the sequence.

\section{Multi-Head Attention: How Multiple Heads Help}
Scaled dot-product attention as described uses a single set of queries, keys, and values to compute a single weighted sum for each position. \textbf{Multi-head attention} extends this idea by running multiple attention operations in parallel, called “heads,” each with its own query, key, and value projections. The Transformer splits the model's dimensionality into $h$ smaller subspaces and each head operates on a subspace.

Specifically, for each attention head $i = 1 \dots h$, we have separate learned projection matrices $W_i^Q$, $W_i^K$, $W_i^V$ that project the input embeddings into $d_k$-dimensional queries, keys, and values for that head. We compute attention for each head:
\[ \text{head}_i = \text{Attention}(Q W_i^Q,\; K W_i^K,\; V W_i^V)~, \] 
using the scaled dot-product attention formula. Each head will produce its own output vector for each position (of size $d_v$, which is often chosen to be $d_k$ for simplicity). Then, the $h$ outputs for each position are concatenated and projected again with another weight matrix $W^O$:
\[ \text{MultiHead}(Q,K,V) = \text{Concat}(\text{head}_1,\ldots,\text{head}_h)\; W^O~. \]

Why do we do this? Because each head can learn to focus on different types of relationships or aspects of the input. For example, in a translation model, one attention head might learn to pay special attention to the next word in the sequence (capturing local context or adjacency), while another head might focus on a related noun somewhere else in the sentence, and yet another might look at the verb tense from a distant part of the sentence. By having multiple heads, the model can attend to multiple things at once. The information from these different heads is then combined, giving the model a richer understanding at that position.

An analogy is to imagine a team of readers each highlighting a text with different colored highlighters, where each color corresponds to looking for a particular pattern or dependency. One reader might highlight pronoun-antecedent relations, another highlights subject-verb relations, another looks for an object related to the verb, etc. At the end, if you overlay all the highlights, you get a comprehensive annotation of the text. Multi-head attention achieves a similar effect: each head is like a specialist focusing on a certain kind of interaction, and by combining them, the model can capture a complex mixture of relationships.

In practice, multi-head attention also allows the model to have a larger “attention capacity.” Instead of one attention weight distribution, the model effectively has $h$ different distributions it can assign, which can be useful when there are several relevant pieces of information for a given word. Importantly, each head operates on a smaller dimensional subspace (since the input is split among heads), so each head can learn to attend based on different features of the embeddings. The result is often more stable and expressive than a single-head attention with the full dimensionality.

To summarize, multi-head attention:
\begin{itemize}
    \item Projects the input into multiple sets of queries, keys, values (one set per head).
    \item Each head performs attention independently (looking at the sequence in possibly different ways).
    \item Their outputs are concatenated and merged, allowing the model to integrate these different perspectives.
\end{itemize}
This was a key innovation of the Transformer, allowing it to simultaneously consider different types of similarities between words. Empirically, using multiple heads was found to improve performance significantly \cite{vaswani2017}, as each head can capture unique aspects of the data.

\section{Positional Encoding: Sinusoidal Encoding and Intuition}
One challenge with the Transformer’s architecture is that, unlike an RNN, it doesn’t have an inherent sense of sequence order. An RNN processes tokens sequentially, so order is implicit. A Transformer processes all tokens in parallel, and the self-attention mechanism by itself is permutation-invariant – if you jumbled the words, the set of queries, keys, and values would be the same, and attention wouldn't know the difference. To give the Transformer a notion of word order, we add a \textbf{positional encoding} to the input embeddings.

The original Transformer used a fixed sinusoidal positional encoding \cite{vaswani2017}. The idea was to create a set of oscillating values (sine and cosine waves) of different frequencies that encode the position index. Specifically, for a token at position $pos$ (starting from 0 for the first position), and for each dimension $i$ of the positional encoding vector:
\[
PE(pos, 2i) = \sin\!\Big(\frac{pos}{10000^{2i/d_{\text{model}}}}\Big), \qquad
PE(pos, 2i+1) = \cos\!\Big(\frac{pos}{10000^{2i/d_{\text{model}}}}\Big)~,
\] 
where $d_{\text{model}}$ is the model’s embedding size, and $i$ runs over the dimensions. This generates a deterministic vector of length $d_{\text{model}}$ for each position $pos$. Each dimension of this positional encoding corresponds to a sinusoid with a different wavelength. Very low $i$ (like $i=0$) yields a high-frequency sinusoid, which means it changes rapidly between positions (capturing fine positional differences), whereas higher $i$ yields slower oscillations (capturing coarse positional information).

These positional encoding vectors are added to the token’s word embedding vectors at the input of the Transformer. By addition, each word embedding is slightly shifted in a unique way depending on its position. The model can then use these signals to infer the relative or absolute position of words. For example, the difference $PE(pos_2) - PE(pos_1)$ between two position encodings has meaningful information about how far apart pos$_2$ is from pos$_1$. The sinusoidal scheme has a nice property: it’s periodic and continuous, so the model can potentially generalize to sequence lengths longer than those seen in training (though in practice, there's a limit), and it can learn to attend to relative positions by combining sin and cos values (like learning to detect phase differences between positions).

Let's build some intuition. Think of each position encoding as a kind of unique fingerprint for that position, created by mixing different frequencies of sine and cosine waves. No two positions (within a reasonable range) will have the exact same encoding because all the sinusoids align differently for each index. The model doesn’t “know” the math of these encodings, but it can learn to interpret them. For instance, it might learn that certain patterns in the positional encoding correspond to the token being early in the sentence vs late, or that if you subtract one token’s positional encoding from another’s, the resulting difference vector might correlate with how far apart they are.

In practice, other approaches to positional encoding exist (like learned positional embeddings where you just have a trainable vector for each position up to a maximum length). The sinusoidal method has the advantage of not adding any new parameters and providing a kind of generalization for very long sequences (you could extrapolate beyond the positions seen in training, in principle). Another form of positional encoding used in some Transformer variants is “relative positional encoding,” which encodes relative distances rather than absolute positions, but the basic need remains: give the model information about order.

To visualize sinusoidal position encoding, imagine one dimension is a sine wave that completes one full cycle every 100 positions, another completes a cycle every 1000 positions, another every 10,000, etc. At position 0, all sinusoids start at well-defined values (sin(0)=0, cos(0)=1). As position increases, each sinusoid oscillates. Any specific position will have a unique combination of sine and cosine values across these frequencies, which the model can use as a signature of that position.

In summary, positional encoding injects order information into the otherwise order-agnostic Transformer. The sinusoidal formulation is a clever, continuous way to do this, ensuring that each position has a unique code and that the model can learn to attend to specific relative positions if useful (since, say, shifting the query by $k$ positions causes a predictable phase shift in the encoding).

\section{Feed-Forward Network (FFN): Structure and Purpose}
Aside from the attention sublayers, each Transformer block contains a simple \textbf{feed-forward network (FFN)} that processes each position independently. This is often called a \emph{position-wise} feed-forward network because it consists of the same MLP (multilayer perceptron) applied to each token's representation separately.

In the original Transformer design, the FFN has two linear transformations with a non-linear activation in between. If the model’s hidden size (embedding size) is $d_{\text{model}}$, typically the FFN first expands the dimensionality to some larger $d_{\text{ff}}$ (for example, $d_{\text{ff}} = 2048$ when $d_{\text{model}} = 512$ in the original), applies a ReLU (or in some models, a GELU) activation, then projects back down to $d_{\text{model}}$. Formally, for each position with input vector $\mathbf{x}$, the FFN computes:
\[ \text{FFN}(\mathbf{x}) = W_2\,\max(0,\,W_1 \mathbf{x} + b_1) + b_2~, \] 
where $W_1$ (dimensions $d_{\text{ff}} \times d_{\text{model}}$), $b_1$, $W_2$ ($d_{\text{model}} \times d_{\text{ff}}$), and $b_2$ are learned parameters, and $\max(0,\cdot)$ is the ReLU activation applied elementwise. The same $W_1, b_1, W_2, b_2$ are used for every position, but since there's no mixing between positions in this operation, it treats each token independently.

Why is this feed-forward network needed? It provides additional transformation capacity for each token’s representation after the attention step. The attention sublayer allows tokens to exchange information, but after that, we want to further process the combined information for each token. The FFN can be thought of as analogous to the fully connected layers in a CNN that come after convolutional layers: once the interaction (convolution or attention) has happened, you use an FFN to compute higher-level features. The FFN can create new features out of the attended output for each token. For example, if after attention a token’s representation has information from itself and related words, the FFN might compute something like “given this combined info, what is a good higher-level feature representation for this token?” It's like a little neural network “brain” at each position that further refines the representation.

Moreover, by increasing the dimensionality to $d_{\text{ff}}$ (like 4 times larger) in the middle, the FFN can capture complex combinations of features in a higher-dimensional space, then project them back down. This expansion gives the model more capacity at each layer to model relationships or patterns that are local to the token.

A complete Transformer layer (in the encoder or decoder) typically goes: multi-head attention $\to$ add \& normalize $\to$ feed-forward $\to$ add \& normalize. The “add \& normalize” refers to the residual connection and layer normalization. The residual connection means that the input of each sublayer (attention or FFN) is added to its output, and then a layer normalization is applied. For instance, if $\mathbf{h}$ is the input to the FFN sublayer and $\mathbf{h}'$ is the output of FFN, we actually output $\text{LayerNorm}(\mathbf{h} + \mathbf{h}')$. This residual addition helps preserve the original information and makes training easier (a technique inspired by ResNets in vision \cite{vaswani2017}). The layer normalization helps stabilize and smooth the training by normalizing the output at each layer.

From a beginner’s perspective, one can view each Transformer layer as doing two things:
\begin{enumerate}
    \item Mix information across the sequence with attention (so each word learns something from other words).
    \item Further transform each word’s representation in isolation with a mini neural network (FFN) to compute new features.
\end{enumerate}
The combination of these is powerful: the attention ensures the model has a rich soup of information at each position, and the FFN then turns that soup into something useful for the next layer or final prediction. Without the FFN, the model would be linear combinations of values only; the FFN introduces non-linearity and interactions among the combined features.

In summary, the feed-forward network in Transformers gives each position a chance to independently process the information it gathered from others and increase the model’s overall expressiveness. It’s a simple but crucial component that works with attention to build deep representations of sequences.

\section{Look-Ahead Mask: Preventing Use of Future Information}
When Transformers are used for tasks like language modeling or text generation (such as in GPT models or the decoder part of the original Transformer for translation), they must not peek at future tokens that they are supposed to predict. We achieve this by using a \textbf{look-ahead mask} (also called a \textbf{causal mask} or \textbf{future mask}) in the self-attention mechanism of the decoder.

The idea is straightforward: when predicting the word at position $i$, the model should only attend to positions $\leq i$ (itself and the past), not any positions $> i$ (the future). During training, we typically have the full sequence, but we mask out the future positions so the model cannot use them in attention. This ensures that the model predicts each token using only the tokens before it, exactly as it would at inference time when generating text step by step.

In practice, the look-ahead mask is implemented by modifying the attention scores before the softmax. We set the scores for any illegal attention links (i.e., query position $i$ to key position $j > i$) to $-\infty$ (or a very large negative number) so that after the softmax they become effectively 0. For example, consider a sequence of length 4. The attention mask matrix (allowed=1, disallowed=0) for a single attention head would look like:
\[
M = \begin{pmatrix}
1 & 0 & 0 & 0\\
1 & 1 & 0 & 0\\
1 & 1 & 1 & 0\\
1 & 1 & 1 & 1
\end{pmatrix},
\] 
where row $i$ corresponds to query position $i$ and column $j$ to key position $j$. Row 3 (0-indexed) has [1,1,1,0] meaning the 4th token can attend to positions 1,2,3 (and itself position 4 if we include it as <= i) but not position 4+1 (which doesn't exist) or beyond. In this matrix, 1 indicates “can attend” and 0 indicates “masked out.” When applying this mask to the computed attention scores, any position with 0 will get $-\infty$ before softmax, resulting in 0 probability assigned.

What this means conceptually is that when the model is computing the representation for the 5th word in a sequence, it will only be allowed to incorporate information from the 1st through 5th words. It cannot sneak a look at the 6th word. This property is crucial for generative tasks so that the model doesn’t cheat by looking ahead. During training, we present the model with the full target sentence (for example, in translation the decoder is given the shifted target sentence), but thanks to masking, each position’s prediction is made using only earlier target words.

If we didn’t apply a look-ahead mask, the self-attention in the decoder could attend to the future words, and the model would essentially see the answer before predicting it, making training meaningless for generation purposes. The mask forces the decoder to behave in an autoregressive manner. At inference time, we generate one word at a time: feed the first word, get the next, then append it, feed the first two, get the third, etc., which naturally ensures we never see future words. The training-time mask just mirrors this process in a parallelized way.

To sum up, the look-ahead mask is a simple trick to enforce causality in sequence generation. By zeroing out attention to future tokens, it ensures the model can be used to generate coherent text one token after another without inadvertently using information that should not be known yet. In the Transformer implementation, this masking is often done by adding a mask matrix $M$ (with $-\infty$ for masked positions) to the $QK^T/\sqrt{d_k}$ scores before softmax. The result is that attention weights for future positions become 0. Thus, the decoder can be trained on full sequences while maintaining the principle that each position predicts the next token using only past context.

\section{Byte-Pair Encoding (BPE): Subword Tokenization Method}
Transformers and other modern language models operate on tokens, which are often not individual characters or fixed-size bytes, but not necessarily full words either. \textbf{Byte-Pair Encoding (BPE)} is a popular method for tokenizing text into subword units, and it was used in models like GPT-2. The goal of BPE (in NLP) is to strike a balance between character-level modeling (which can handle any input but is very long sequence length) and word-level modeling (which can’t handle out-of-vocabulary words). BPE builds a vocabulary of common subword chunks so that frequent words are usually one or a few tokens, while rare or unseen words can be constructed from smaller pieces.

The BPE algorithm for text works roughly as follows \cite{sennrich2016}:
\begin{enumerate}
    \item Start with all words in the training corpus broken down into individual characters (plus a special end-of-word symbol so that we know where one word ends).
    \item Count the frequency of every pair of symbols that appear next to each other. A “symbol” is initially a character, but as we merge, symbols can become sequences of characters.
    \item Find the most frequent adjacent pair of symbols in the corpus.
    \item Merge that pair into a single new symbol (effectively, add a new token to the vocabulary which is the combination of those two).
    \item Replace all occurrences of that pair in the text with the merged symbol.
    \item Repeat steps 2-5 until we have reached the desired vocabulary size or there are no more pairs to merge.
\end{enumerate}

The result is a vocabulary that includes single characters, some common two-letter combinations, some longer subwords, and eventually whole words if they were very frequent. For example, suppose our corpus includes the words “low”, “lowest”, and “newer”. Initially:
\[
\text{low} \to l~o~w~\_ , \quad
\text{lowest} \to l~o~w~e~s~t~\_ , \quad
\text{newer} \to n~e~w~e~r~\_ ,
\] 
(where “\_” represents a end-of-word marker). The most frequent pair might be “l” + “o” (because “lo” appears in “low” and “lowest”). So we merge “lo” into a single token “lo”. Now we have:
\[
\text{low} \to lo~w~\_, \quad
\text{lowest} \to lo~w~e~s~t~\_, \quad
\text{newer} \to n~e~w~e~r~\_.
\]
Next, maybe “lo” + “w” is frequent (in “low” and “lowest”), merge to get “low”. Now:
\[
\text{low} \to low~\_, \quad
\text{lowest} \to low~e~s~t~\_, \quad
\text{newer} \to n~e~w~e~r~\_.
\]
Next, perhaps “e” + “r” is frequent (in “newer”), merge into “er”. Now:
\[
\text{low} \to low~\_, \quad
\text{lowest} \to low~e~s~t~\_, \quad
\text{newer} \to n~e~w~er~\_.
\]
And then “low” + “est” might merge depending on frequencies, etc. Eventually, we might end up with vocabulary tokens like “low”, “est”, “new”, “er”, etc., such that “lowest” is tokenized as “low” + “est” and “newer” as “new” + “er”. We achieved an open vocabulary: even if we encounter a new word like “newest” later, it can be tokenized as “new” + “est” which are in our vocabulary.

The reason this is helpful is that the model doesn’t have to learn from scratch that “est” is a suffix meaning something like “most” in superlatives, or that “low” is a root – it sees those as separate tokens. It also drastically reduces the number of unknown or out-of-vocabulary words. Almost any word can be expressed as a sequence of BPE tokens from a good vocabulary. The merges effectively incorporate frequent letter combinations (including whole words for very common ones) so the model can treat them as a single token, which is more efficient and often more meaningful.

Byte-Pair Encoding was originally a compression algorithm \cite{gage1994} (it replaced common byte pairs in data with shorter codes). In NLP, \cite{sennrich2016} adapted it for word segmentation. One advantage of BPE is that it’s deterministic given the learned merges: any new text will be tokenized in a consistent way by greedily applying the longest possible merge rules (so it always prefers longer known subwords over splitting into characters).

In modern Transformers:
\begin{itemize}
    \item GPT models use variants of BPE (GPT-2 used BPE on byte sequences, treating text as bytes to include any Unicode).
    \item BERT uses a similar approach called WordPiece, which is conceptually similar to BPE (builds a vocabulary of subwords based on frequency and likelihood).
\end{itemize}
Both ensure that common words are usually one token (“the”, “apple”), slightly less common words might be two tokens (“ap@@” + “ple” in WordPiece notation, or “ap”, “\#\#ple”), and rare words break into several pieces or characters.

For a beginner, think of BPE as teaching the model a “syllabary” or “alphabet” of word pieces. Instead of single letters (too slow to spell everything out) or whole words (can’t cover all words, especially misspellings or names), it learns common chunks. This way, the model sees “nationalization” broken into “national”, “ization” for example, understanding those parts, and it can recombine parts for new words (“internationalization” would share “national” and “ization”). BPE makes training faster (fewer time steps than character-level) and generalization better (no out-of-vocab errors).

\section{GPT Architecture: Decoder-Only Transformers and Autoregressive Modeling}
GPT (Generative Pre-trained Transformer) is a line of models developed by OpenAI that exemplifies the use of Transformers for language generation. The core idea of GPT is to use a \textbf{decoder-only Transformer} in an \textbf{autoregressive} fashion to model text. Let's unpack what that means.

The original Transformer from \cite{vaswani2017} had two parts: an encoder and a decoder. The encoder read input (e.g., an English sentence) and the decoder produced output (e.g., the translated French sentence), attending to encoder states as well as previous decoder states. For language modeling, however, we typically don’t have an external input sequence; the task is to predict the next token in a sequence given all previous tokens. GPT simplifies the architecture by using just the Transformer decoder stack (without an encoder stack). Essentially, GPT is a Transformer decoder with multiple layers of masked self-attention and feed-forward networks, trained to predict the next word in a sequence.

Key characteristics of GPT architecture:
\begin{itemize}
    \item It uses \textbf{masked self-attention} (as described with the look-ahead mask) so that at training time, the model is trained on sequences and cannot see beyond the current position.
    \item It is trained in an \textbf{autoregressive} manner, meaning it maximizes the likelihood of each token given the previous tokens. If we have a sequence of tokens $x_1, x_2, \ldots, x_n$, it trains to estimate $P(x_n | x_1, \ldots, x_{n-1})$ for each position. The product of these conditional probabilities gives the joint probability of the sequence.
    \item There is no encoder input; the model generates text from a start-of-sequence prompt or from scratch. You can feed in some initial text (a prompt), and then it will continue generating additional text.
    \item Architecturally, GPT uses multi-head self-attention and feed-forward layers just like the standard Transformer decoder. It also includes residual connections and layer norms. The final output of GPT is produced by taking the Transformer’s output at the last layer and feeding it to a softmax layer over the vocabulary to get probabilities for the next token.
\end{itemize}

\textbf{GPT-1} \cite{radford2018} was the first such model, demonstrating that a Transformer language model pre-trained on a large corpus (BooksCorpus) and then fine-tuned on specific tasks could yield good results. It had on the order of 117 million parameters and was a proof of concept that pre-training on unlabeled text can help downstream tasks (this was around the same time the BERT model came out, which is different in approach but also based on Transformers).

\textbf{GPT-2} \cite{radford2019} scaled this idea up dramatically. GPT-2 had up to 1.5 billion parameters in its largest version and was trained on a very large dataset (around 8 million web pages, called WebText). GPT-2 showed remarkable ability to generate fluent and coherent paragraphs of text, perform rudimentary reading comprehension, translation, and question answering in a zero-shot way (without task-specific fine-tuning) just by being prompted with an example. For instance, if prompted with an English sentence followed by its French translation a couple of times, GPT-2 could continue and translate the next sentence (even though it was not explicitly trained for translation, it picked up some ability from its huge training corpus). The release of GPT-2 was staged carefully due to concerns about misuse (like generating fake news), highlighting how powerful the approach is.

\textbf{GPT-3} \cite{brown2020} pushed the scale to an unprecedented level: 175 billion parameters, trained on an even larger corpus of internet text. Instead of fine-tuning for each task, the focus with GPT-3 was on \emph{few-shot learning} via prompting. Users could give GPT-3 a prompt with a few examples of a task (like a couple of math problems and solutions, or a question and its answer) and then ask a new question; GPT-3 often could continue the pattern and produce a correct or at least plausible answer. This showed that with enough capacity and data, a language model could learn to perform many tasks implicitly. The GPT-3 architecture was similar to GPT-2 (decoder-only Transformer with masked self-attention), just much larger and with some improvements in training techniques and initialization. It uses Byte-Pair Encoding for tokenization with a 50,000 token vocabulary, and it’s so large that it captures a lot of world knowledge and linguistic patterns.

An example of using GPT model (autoregressive generation) in pseudocode:
\begin{lstlisting}[language=Python]
prompt = "Once upon a time"
output = prompt
for i in range(100):  # generate 100 tokens
    logits = model(output)              # get raw probabilities for next token
    next_token = sample_from_softmax(logits[-1])  # take the last token's logits
    output = output + next_token        # append the generated token
print(output)
\end{lstlisting}
Here, the model is a trained GPT-like Transformer. We start with a prompt and iteratively sample the next token and append it. The look-ahead mask during training ensured that at generation time this works correctly (the model always conditions only on what’s already generated).

In summary, GPT’s architecture is a stack of Transformer decoder layers that use self-attention (with masking) and feed-forward networks. It is trained as a language model on massive text data. Over the iterations from GPT-1 to GPT-3 (and beyond), the trend has been: bigger models + more data = better performance and emergent abilities. The “decoder-only” design is extremely effective for generating text, and it forms the backbone of many state-of-the-art systems, including the famous ChatGPT (which is a further fine-tuned version of a GPT model). GPT models highlight the power of the Transformer architecture for generative tasks and have opened the era of large-scale “foundation models” that can be adapted to many tasks.

\section{BERT and RoBERTa: Masked Language Modeling and Improvements}
While GPT is a Transformer for generation (one-directional, autoregressive), \textbf{BERT} (Bidirectional Encoder Representations from Transformers) takes a different approach suitable for understanding and non-generative tasks. BERT \cite{devlin2019} uses the Transformer \emph{encoder} architecture and is trained with a technique called \textbf{masked language modeling (MLM)}. Instead of predicting the next word, BERT is trained to predict randomly masked words in a sentence using both left and right context.

Here’s how BERT’s training works:
\begin{itemize}
    \item You take an input sentence (or pair of sentences). You randomly choose some of the tokens (e.g., 15\% of them) and replace them with a special \texttt{[MASK]} token. For example: “The \texttt{[MASK]} was very hungry.” 
    \item The Transformer encoder reads the whole sequence. Because it’s an encoder, it has no restriction on attention direction — each token can attend to all others (this is why we call it bidirectional; the model can use both left and right context).
    \item The model is trained to output the correct identity of the masked tokens. In our example, if the original sentence was “The cat was very hungry,” the model should produce “cat” in place of the mask.
\end{itemize}
By doing this, BERT learns deep bidirectional representations of language. It’s as if it’s doing a cloze task (fill-in-the-blanks). This MLM objective forces BERT to understand the context around a word from both sides, which is very useful for tasks like understanding the meaning of a sentence, answer extraction, etc. In addition to MLM, BERT was also trained on a \textbf{next sentence prediction (NSP)} task: the model would see two sentences and predict if the second sentence actually follows the first in the original text. They did this by sometimes feeding pairs of consecutive sentences from the corpus as positive examples, and random pairings as negative examples. NSP was intended to help BERT understand sentence relationships (important for QA or natural language inference). However, later research found NSP might not be crucial.

BERT’s architecture uses the encoder Transformer: multiple self-attention layers without any masking (except the trivial mask to ignore padding tokens). It introduces special tokens \texttt{[CLS]} at the start (whose output embedding is used for classification tasks) and \texttt{[SEP]} to separate sentences. During fine-tuning, the \texttt{[CLS]} token’s output is often fed into a classification layer for tasks like sentiment analysis or QA (with further processing to extract answers). BERT achieved state-of-the-art on many NLP tasks after fine-tuning, showcasing the power of pre-trained Transformers for language understanding.

\textbf{RoBERTa} (Robustly Optimized BERT) \cite{liu2019} is an improved version of BERT introduced by Facebook AI. RoBERTa didn’t change the architecture, but rather the training procedure and hyperparameters to get more out of the model:
\begin{itemize}
    \item They trained on much more data (including longer training with bigger batches). BERT was trained on BookCorpus and Wikipedia (16GB of text); RoBERTa used those plus news, web text, etc., totaling over 160GB.
    \item They removed the Next Sentence Prediction objective. RoBERTa found that NSP was not helpful and that one can just train on the MLM objective alone and still get great results. In fact, removing NSP allowed them to use uninterrupted text sequences and vary input length.
    \item They used \textbf{dynamic masking}: Instead of deciding once which words to mask and keeping that fixed for an example throughout training (as BERT did), they would change which tokens are masked on different passes. This means the model sees more variety – e.g., in one epoch the word “dog” might be masked in a sentence, in another epoch maybe “barks” is masked for the same sentence. This leads to more robust learning of token representations.
    \item Other optimizations: larger batch size, different learning rate schedule, etc., to make training more effective.
\end{itemize}
The result was that RoBERTa outperformed BERT on many benchmarks, essentially by “training the heck out of it” with better practices. The name “Robustly Optimized” reflects that they did a thorough job of finding how to get the most out of the BERT architecture.

From a beginner’s perspective, BERT and RoBERTa show another paradigm of Transformer use: not for generating text, but for encoding text into a deep understanding. They produce contextual embeddings – each token’s output is an embedding that reflects the meaning of that token in context. These embeddings can then be used for downstream tasks: classification, span extraction, etc. For example, to do sentiment analysis, you can take the \texttt{[CLS]} token’s embedding from BERT (which is like a summary of the sentence) and put a classifier on top. Or for QA, you can take two sentences (passage and question) fed together with a \texttt{[SEP]} separator, and train the model to mark the start and end tokens of the answer span in the passage.

RoBERTa’s improvements illustrate how important the training setup is. It wasn’t that BERT’s idea was flawed; it was that you could push it further with more data and some simplifications. Indeed, RoBERTa removed NSP entirely and still did better, implying the bidirectional MLM was doing most of the heavy lifting in BERT’s learning.

In summary, BERT introduced the concept of masked language modeling to get bidirectional context in Transformer encoders, and RoBERTa fine-tuned that approach to yield even better pre-trained models. These models are not used to generate free text (they're not typically asked to continue a story), but rather to provide powerful language understanding that can be specialized to many tasks with a bit of fine-tuning.

\section{Vision Transformer (ViT): Transformers for Images as Patches}
Transformers have also made a leap from text to vision. The \textbf{Vision Transformer (ViT)} \cite{dosovitskiy2020} is an application of the Transformer encoder architecture to image classification (and by extension, other vision tasks). The challenge was how to feed an image to a Transformer, which expects a sequence of discrete tokens (like words). ViT’s solution: split the image into patches and treat each patch as a “word” or token.

Here’s how ViT works:
\begin{itemize}
    \item You take an input image (say 256x256 pixels). You choose a patch size, e.g., 16x16 pixels. You divide the image into non-overlapping 16x16 patches. In this example, you’d get $(256/16) \times (256/16) = 16 \times 16 = 256$ patches.
    \item Flatten each patch into a vector. A 16x16 patch with 3 color channels has $16\times16\times3 = 768$ pixel values. You then linearly project this 768-dimensional vector to the model’s embedding dimension (let’s say 768 as well for convenience). This gives a patch embedding, analogous to a word embedding in NLP.
    \item Similar to BERT’s use of a \texttt{[CLS]} token, ViT prepends a learnable embedding called the \texttt{[CLS]} token at the beginning of the sequence of patch embeddings. This token will serve as a summary representation of the whole image, which we’ll use for classification.
    \item Also add positional encodings to each patch embedding, because we want to give the model information about where each patch is located in the image (top-left, bottom-right, etc.). They used learned positional embeddings or fixed ones (the ViT paper used learned position embeddings since image positions are fixed grid locations).
    \item Now you have a sequence of tokens: [CLS] + patch 1 + patch 2 + ... + patch N. This sequence is fed into a Transformer encoder. The Transformer layers then perform self-attention among all these patches (and the CLS token).
\end{itemize}

The self-attention mechanism will allow the model to globally reason about the image. Each patch can attend to any other patch. For instance, if part of the image contains an eye of a cat and another part contains an ear of the cat, the model can make connections between those patches via attention, which might help it realize the overall object is a cat. This global receptive field is a stark contrast to convolutional neural networks, which typically only mix information locally and need many layers to achieve global interaction. A single Transformer attention layer is global (any patch can directly look at any other patch's features).

After passing through multiple Transformer encoder layers, we take the final hidden state corresponding to the \texttt{[CLS]} token. We then attach a simple feed-forward neural network (like a single linear layer or a small MLP) on top of that \texttt{[CLS]} representation to produce a classification (for example, which category the image belongs to). We train the whole system end-to-end on image classification loss (like cross-entropy for the correct label).

Some important notes and intuition:
\begin{itemize}
    \item Each patch is like a “word” describing a part of the image. At first, this might seem lossy (we threw away spatial resolution within each patch by flattening), but if the patches are small (like 16x16), the network can still reconstruct spatial information by looking at neighboring patches and their relations.
    \item The positional encoding ensures the model knows patch \#5 is, say, top right corner and patch \#200 is bottom left, etc. Otherwise, it would treat the image as a bag of patches with no order.
    \item ViT doesn’t inherently know that patches near each other are also spatially related unless it learns that via attention. This means it has less built-in inductive bias than a CNN (which assumes locality and translation invariance), so ViT typically needs a lot of data to train from scratch successfully. In the ViT paper, they pre-trained on very large datasets (like ImageNet-21k or JFT-300M) to get good results, and then fine-tuned to smaller ones.
    \item One advantage of this approach is scalability and flexibility. If you have a bigger image or different resolution, you can divide into patches and still feed into the same model (though position embedding might need slight adaptation). Also, self-attention can potentially capture long-range dependencies (like parts of an object on opposite sides of the image) in one layer.
\end{itemize}

What about performance? ViT, when sufficiently pre-trained, matched or exceeded the performance of convolutional networks on image classification tasks \cite{dosovitskiy2020}. It opened the door to purely Transformer-based vision models. After ViT, many variants and improvements came out (like Swin Transformer, which introduces some locality, etc.), but the core idea of treating image patches as tokens was a watershed moment.

For a student, you can think of ViT as “treating an image like a sentence of patches.” The model doesn’t inherently know what a pixel is or what a contiguous shape is—it figures out relevant patterns through attention. Early layers might attend to very adjacent patches to detect simple shapes (like edges, similar to what CNN filters do), and later layers might connect distant parts to recognize whole objects.

In summary, Vision Transformer applies the Transformer to images by chopping images into fixed-size patches. Each patch becomes a token embedding, and then a Transformer encoder processes a sequence of those patch tokens (plus a special classification token). The elegance of this approach is that it shows Transformers are a general and flexible architecture not just for language, but for any data that can be turned into a sequence of embeddings. The success of ViT demonstrates that with enough data, you can drop traditional convolution completely and still have a high-performing vision model, leveraging the same attention mechanisms that proved powerful in NLP.

\section{CLIP: Contrastive Training of Image and Text Encoders}
\textbf{CLIP} (Contrastive Language-Image Pretraining) is a model by OpenAI that connects vision and language by training an image encoder and a text encoder together in a joint multimodal space \cite{radford2021}. The goal is for the model to learn to associate images with their correct descriptions. Once trained, CLIP can understand images and captions in a versatile way (for example, you can use it for zero-shot image classification).

How does CLIP work? It uses a \emph{contrastive learning} approach:
\begin{itemize}
    \item There are two networks: one takes an image and produces an image embedding (a vector representation), and another takes a text (for example, a caption) and produces a text embedding.
    \item The model is trained on a large dataset of image-text pairs. For example, one training example might be an image of a dog and the caption “a dog jumping over a log.”
    \item During training, CLIP takes a batch of, say, $N$ image-text pairs. It processes all images through the image encoder (like a ResNet or a Vision Transformer) and all texts through the text encoder (which can be a Transformer encoder) to get $N$ image vectors and $N$ text vectors.
    \item Then it computes similarity (like cosine similarity or dot product) between every image embedding and every text embedding. This forms an $N \times N$ matrix of similarities.
    \item The diagonal of this matrix (pair $i$ with text $i$) are the “correct” image-text pairs, and off-diagonals are mismatched pairs. CLIP uses a contrastive loss (often InfoNCE loss) that basically says: each image should be most similar to its own caption and less similar to other captions, and vice versa for each caption.
    \item Concretely, the loss can be implemented by doing a softmax on each row (images) to predict which column (caption) is the match, and a softmax on each column (caption) to predict the matching image. The model is trained to maximize the probability of the correct matches (diagonals) in these softmax distributions. This is equivalent to maximizing the dot product of true image-caption pairs while minimizing it for incorrect pairs.
\end{itemize}

Through this training, the image encoder learns to produce embeddings that capture the semantic content of the image (because it needs to align with the text that describes that image), and the text encoder learns to produce embeddings that capture the meaning of the text in a way that aligns with images. They meet in the middle, so to speak, in a shared embedding space where matching images and texts have high similarity.

What is special about this approach is that it doesn’t require explicit labels like “dog” or “cat” for images; it uses natural language descriptions as supervision. The supervision is weaker (language is rich and varied), but also much more flexible (it can describe anything you have data for). OpenAI collected (or used) a huge dataset of 400 million (image, caption) pairs from the internet for training CLIP, which is why it works so well.

After training, CLIP can be used in cool ways:
\begin{itemize}
    \item You can do \textbf{zero-shot image classification}. Suppose you have a standard classification task with labels like {dog, cat, airplane, ...}. Normally, you’d train a classifier on labeled images. With CLIP, you can avoid that by providing the text encoder with the possible labels in a prompt like “a photo of a dog”, “a photo of a cat”, “a photo of an airplane”, etc. CLIP’s text encoder will produce an embedding for each label prompt. Then you embed the image using the image encoder, and you compare the image embedding to each label text embedding to see which is most similar. Whichever is highest, you predict that as the class. In experiments, CLIP was surprisingly good at this zero-shot classification, often matching or exceeding the performance of fully supervised models on many datasets. For example, CLIP can take an image it’s never seen and tell with good accuracy which of 1000 ImageNet categories it likely belongs to, just by using the category names and “a photo of” prompt.
    \item CLIP can also be used for image search (retrieval). If you encode a bunch of images and a query text, you can find which image embedding has the highest similarity to the query text embedding—essentially finding images that best match a description. Conversely, you can encode an image and search through a corpus of text embeddings for the best matching description.
    \item Because CLIP has a rich joint space, it has even been used in generative tasks. For instance, models like DALL-E 2 use CLIP’s image encoder as part of a feedback mechanism to generate images that match a text prompt.
    \item CLIP’s image encoder (often a ViT) can be used as a general vision feature extractor for other tasks, and its text encoder (a Transformer) is like a language model tuned for image descriptions.
\end{itemize}

Why do we call it “contrastive training”? Because the model isn’t directly told “this is the correct caption for this image” via a classification loss; instead, it is given both correct and incorrect pairings and has to sort them out. It contrasts the correct pair against all the incorrect ones. The training tries to maximize the similarity of the true pairs and minimize similarity of the false pairs. This technique is powerful for representation learning, as it doesn’t require one-hot labels, only pairings or similarities.

From an intuitive standpoint, think of CLIP’s training as a game: it sees a jumble of images and captions and must figure out which goes with which. To do that well, it needs to develop an understanding that, e.g., in an image of a dog on a beach, the embedding for that image should somehow encode “dog” and “beach” so that it will align with a caption talking about a dog on a beach, and not align with a caption about “an airplane in the sky.” Similarly, the text “dog on a beach” must produce features that highlight dogness and beachness to match the image.

The result is a model that “knows” about both modalities. It has effectively learned a lot of visual concepts and their names from natural language supervision. This approach leveraged the abundance of image-text data on the internet (like pictures with alt text or user-provided descriptions) instead of needing curated labeled data.

In summary, CLIP is a multimodal model that learns to connect images and text by bringing matching image-caption pairs closer in a shared embedding space while pushing non-matching pairs apart. It demonstrates that Transformers (and similar architectures) can bridge different data types (vision and language) effectively. CLIP’s representations turned out to be extremely useful and have been used in various applications where understanding the link between visuals and language is required.

\section{InstructGPT and Reward Models: Fine-Tuning with Human Preference}
Large language models like GPT-3 are very powerful but they are trained purely to predict the next token, not necessarily to follow human instructions or produce helpful, correct answers. \textbf{InstructGPT} \cite{ouyang2022} is an approach by OpenAI to fine-tune GPT-3 models so that they follow instructions better and align with what humans expect. The core of InstructGPT is using human feedback to train a \textbf{reward model} and then using reinforcement learning (specifically RL from human feedback, RLHF) to optimize the language model with that reward model.

The InstructGPT process can be summarized in three phases:
\begin{enumerate}
    \item \textbf{Supervised fine-tuning (SFT) with human demonstrations:} First, they ask human annotators to demonstrate ideal model behaviors for a variety of instructions/prompts. For example, if the prompt is “Explain the moon landing in simple terms,” a human might write a good answer. A dataset of such prompt-answer pairs is collected. The base GPT-3 model is then fine-tuned on this dataset via supervised learning (just standard next-token prediction on the human-written answers). This gives an initial model that is already somewhat aligned with human intent because it mimics human-written responses.
    \item \textbf{Reward model training:} Next, they gather human preference data. They have the current model (from step 1) generate multiple answers to the same prompt, or they use some variants of the model to generate different responses. Human labelers then rank these responses from best to worst according to which one follows the instruction best, is most helpful, correct, and so forth (taking into account things like truthfulness, lack of toxicity, completeness). Using these rankings, they train a \textbf{reward model} $R$ that, given a prompt and a candidate response, outputs a scalar score (a “reward”) indicating how good that response is. Essentially, the reward model is trained so that it assigns higher scores to responses that humans ranked higher. Technically, one can train this as a pairwise comparison: for two responses $A$ and $B$ to the same prompt, if the human said $A$ is better than $B$, then train the reward model to have $R(\text{prompt}, A) > R(\text{prompt}, B)$ by a margin. This reward model is typically a Transformer (often the same architecture as the language model, but smaller) with an output head that produces a scalar value instead of predicting language.
    \item \textbf{Reinforcement learning fine-tuning (RLHF with PPO):} Now they have a reward model that can judge responses. They then fine-tune the original language model (from step 1) using reinforcement learning to maximize the reward model’s score. A common algorithm for this is \textbf{Proximal Policy Optimization (PPO)} \cite{schulman2017}, which is a stable policy gradient method. In this context, the “policy” is the language model: given a prompt (state), it produces a response (action) one word at a time, and after the response is complete, the reward model provides a reward signal. However, we can't wait until an entire essay is written to give a reward for each token choice, so PPO is used in an episode-wise fashion, optimizing the expected reward of the final output. The loss function also often includes a term to keep the fine-tuned model’s outputs close to the original model’s outputs (this prevents it from diverging too far and saying random stuff just to game the reward model). That term is typically implemented as a Kullback-Leibler (KL) divergence penalty between the new policy (fine-tuned model) and the original model’s distribution, which ensures the model doesn’t forget its base knowledge or become unstable.
\end{enumerate}

After this process, the resulting model is InstructGPT: it is much better at following instructions and producing answers that humans prefer, compared to the base model. For example, if you ask the base GPT-3 “How do I make a peanut butter sandwich?”, it might give a reasonable answer or it might ramble or not directly address that it should give instructions. The InstructGPT model will more likely produce a step-by-step, clear answer because it was tuned to do exactly that when users ask for it.

The role of the reward model here is crucial. It serves as a proxy for human judgment so that the reinforcement learning algorithm can have a signal to optimize. The reward model essentially captures “human preference” in a number. In reinforcement learning terms, the language model is the agent, the prompt is the state, it generates a sequence of actions (words) to form a response, then the reward model (plus maybe some baseline to reduce variance) gives a reward at the end of the episode. PPO then updates the model weights to increase the probability of word sequences that yield higher reward.

This combination of techniques is known as RLHF (Reinforcement Learning from Human Feedback). The concept was inspired by earlier work in which an agent could be trained to satisfy human preferences \cite{christiano2017}. In instructGPT, it’s applied to conversational AI. The success of instructGPT was clear: the fine-tuned models were much more aligned – they were less likely to produce irrelevant or harmful answers, and users found them more useful. In fact, ChatGPT as known publicly is essentially this kind of model.

It’s important to note that the reward model can have flaws or biases depending on the data. The model will optimize for whatever the reward model (and thus the human labelers) favor. If not careful, it can learn to trick the reward model (model outputs that look good to the reward model but might be nonsensical to a human if the reward model has weaknesses). To mitigate that, the iterative approach and careful checks are used.

In summary, InstructGPT augments a pre-trained language model with an extra round of training that involves humans in the loop:
- Humans provide example responses (to prime the model).
- Humans compare model outputs to train a reward model.
- The model is then fine-tuned using reinforcement learning to maximize the reward (i.e., human satisfaction).
This process produces a model that more reliably does what you want when you prompt it – essentially making the model follow instructions and align with human preferences for response quality. It demonstrates how we can steer large models to be more useful and safe by defining what we want (via the reward model) and optimizing for it.

\section{Reinforcement Learning with PPO: Fine-Tuning Language Models}
Reinforcement Learning (RL) comes into play in fine-tuning language models when we have a reward signal instead of direct supervised targets. In the case of InstructGPT and similar, once we have a reward model that can score an output, we use an RL algorithm to adjust the language model’s behavior. \textbf{Proximal Policy Optimization (PPO)} \cite{schulman2017} is a popular choice for this because of its stability and efficiency in handling large policy networks (like a big Transformer).

Let’s break down how PPO is used in this context, conceptually:
- We treat the language model as a \textbf{policy} $\pi_\theta$ that, given a state (which is the prompt or conversation history), outputs an action (a probability distribution over the next token, and sequentially a whole response).
- Generating a full response is like the policy producing a sequence of actions until a termination (perhaps a special end-of-sequence token or reaching a length).
- After the model produces a response, we can compute a \textbf{reward} $R$ for that output using the reward model (or other criteria). For simplicity, think of $R$ as a single number evaluating the whole response.
- Now, we want to adjust the policy $\pi_\theta$ to increase the expected reward $E[R]$ over the distribution of prompts (and the model’s own stochastic outputs). This is a reinforcement learning problem.

Vanilla policy gradient methods would tell us to nudge the model’s output probabilities to make high-reward outputs more likely. However, directly doing that can be unstable if the changes are too large (the model might diverge and drop performance, or the language might become repetitive or collapse). PPO offers a solution by using a clipped objective:
\[ L(\theta) = E_{\text{prompt}, \text{response}} \Big[ \min\big( r(\theta) \, A, \; \text{clip}(r(\theta), 1-\epsilon, 1+\epsilon) \, A \big) \Big], \]
where $r(\theta)$ is the probability ratio $\frac{\pi_\theta(\text{response}|\text{prompt})}{\pi_{\theta_\text{old}}(\text{response}|\text{prompt})}$, and $A$ is the advantage (which in simplest form can be the reward minus a baseline). The idea is:
- If an output has higher reward than expected (positive advantage), we want to increase its probability. If it has lower reward (negative advantage), decrease its probability.
- But the \textbf{clip} limits how much we change $r(\theta)$. For instance, if $\epsilon = 0.2$, it prevents the new policy from going beyond 20\% change in probability for the action compared to the old policy, in a single update step. This means we take small, cautious steps.
- The min between the unclipped and clipped term means if the improvement $r(\theta)A$ is too large (beyond the clip), we only take as much improvement as the clip allows. This keeps updates from blowing up.

In practice, they also maintain a value function (a model that predicts expected reward) to compute the advantage more accurately and reduce variance. But conceptually, the above is fine.

How does this apply to our language model?
- We have an initial policy (the supervised fine-tuned model from step 1 in the previous section, or even the base model) as $\pi_{\theta_\text{old}}$.
- We generate responses from it (often using sampling to explore different outputs).
- We get rewards for each.
- Then we update the model parameters a little in the direction that would increase the probability of good responses and decrease that of bad responses, but limited by the PPO clipping to avoid going too far.

We also often include the KL divergence penalty as part of the reward or as a separate term. This effectively acts like a penalty if the new policy $\pi_\theta$ drifts too far from the original distribution $\pi_{\theta_\text{old}}$. It’s like saying “don’t forget how to speak fluent language or start outputting nonsense just to tweak the reward.” It keeps the language style and general knowledge anchored.

Imagine the reward model strongly prefers very enthusiastic answers (maybe it learned humans like exclamation marks). If unpenalized, the policy might start outputting exclamation-laden answers everywhere. The KL penalty would discourage it from straying too far from the more neutral base distribution unless it truly improves reward.

The end result of using PPO is a balanced update that improves the desired behavior incrementally without wrecking the model’s language ability. PPO is favored because:
- It’s relatively straightforward to implement on top of existing policy gradient code.
- It doesn’t require second-order derivatives or complex math (like TRPO, an older method, does).
- It’s been found to be stable across many domains.
- The clipping mechanism is a heuristic that works well to prevent oscillations and ensure the training doesn’t collapse even if the reward signal is sometimes noisy or imperfect.

For a beginner: you can think of PPO as a careful teacher for the model. Instead of just saying “this output was good, so massively boost it,” PPO says “this output was good, let’s make it a bit more likely next time, but not too much, and this other output was bad, let’s make it a bit less likely.” Over many iterations, these nudges add up to a noticeable behavior change. PPO just ensures the nudges are not too big each time (proximal = nearby, meaning the new policy stays close to the old one after each update).

In code or algorithm form, each iteration of PPO for language model fine-tuning might look like:
1. Sample a batch of prompts from a dataset.
2. For each prompt, have the model generate a response (maybe multiple responses) by sampling.
3. For each (prompt, response), compute reward via the reward model.
4. Compute the advantage $A = R - b$ (where $b$ could be a baseline from a value function).
5. Compute gradients of the PPO loss with respect to $\theta$ and update $\theta$ (possibly with multiple mini-batches from the collected data, known as epochs in PPO).
6. Optionally update the baseline (value function) to better predict reward.
7. Rinse and repeat with new data from the updated policy.

Using PPO in this way was pivotal to making InstructGPT work. Without RL, if they had tried to directly fine-tune on a scalar reward with supervised learning (which doesn’t make sense directly because there’s no target output), or if they tried to treat the highest-ranked output as a “correct answer” for cross-entropy, that would ignore the nuanced feedback. RLHF with PPO uses all the information (relative preferences) and finds an optimal policy that maximizes expected reward.

In summary, PPO is the reinforcement learning algorithm that takes the feedback from the reward model and turns it into a policy improvement for the language model, doing so in a stable, controlled fashion. It ensures that the fine-tuning process doesn’t ruin the model while trying to align it with human preferences, leading to a high-quality tuned model that behaves better according to those preferences.

\section{Direct Preference Optimization (DPO): A New Approach Without RL}
While RL with PPO has been successful, it is complex and can be unstable or resource-intensive. \textbf{Direct Preference Optimization (DPO)} \cite{rafailov2023} is a more recent method that aims to fine-tune language models from human preference data \emph{without using RL algorithms} at all. Instead, DPO finds a way to directly incorporate the preference information into a supervised-style objective.

The key idea of DPO is to derive an objective that the language model can be optimized on which, in theory, leads to the same optimum as the RLHF setup. DPO starts from the insight that if we have a reward model $R$ learned from human preferences, we can consider an “optimal” policy $\pi^*(y|x)$ that would be proportional to the base model policy $\pi_0(y|x)$ times an exponential of the reward:
\[ \pi^*(y|x) \propto \pi_0(y|x) \exp\{ \beta R(x,y)\}, \] 
for some scale factor $\beta$ (like an inverse temperature). Intuitively, this $\pi^*$ is a reweighted version of the original language model $\pi_0$ where outputs with higher reward (according to $R$) are boosted. If we could achieve $\pi^*$, that would be the ideal aligned model (assuming $R$ perfectly captures human preferences).

DPO then tries to directly train $\pi_\theta$ to approximate $\pi^*$ without doing a step-by-step RL procedure. They derive a loss that looks like a binary logistic regression on pairs of outputs:
- For a given prompt $x$, suppose we have two responses: $y^+$ (the human-preferred one) and $y^-$ (the dispreferred one).
- We want $\pi_\theta(y^+|x)$ to be higher than $\pi_\theta(y^-|x)$. Specifically, from the formula of $\pi^*$, the ratio should relate to the exponent of the reward difference:
\[ \frac{\pi^*(y^+|x)}{\pi^*(y^-|x)} = \frac{\pi_0(y^+|x)}{\pi_0(y^-|x)} \exp\{\beta (R(x,y^+) - R(x,y^-))\}. \]
If $y^+$ is better, $R(x,y^+) > R(x,y^-)$, we expect the optimal policy to put higher relative probability on $y^+$.
- DPO leads to a loss like:
\[ L_{\text{DPO}} = - \log \sigma( \beta(\log \pi_\theta(y^+|x) - \log \pi_\theta(y^-|x)) ), \] 
or conceptually 
\[ -\log \frac{\pi_\theta(y^+|x)}{\pi_\theta(y^+|x) + \pi_\theta(y^-|x)}, \] 
which is basically saying “we want $\pi_\theta$ to give the pair $(y^+, y^-)$ the correct ordering, with as high confidence as possible.”

In simpler terms, DPO treats the preference comparison as a training example: “For prompt $x$, output $y^+$ should have a higher score than $y^-$. Make the model’s logits reflect that.” This is reminiscent of how we might train a binary classifier to pick which output is better, except here the “classifier” is embedded in the generative model’s probabilities.

The surprising result from the DPO authors was that optimizing this kind of loss (plus some regularization to not drift too far from the original model) yields a model that is as good as the PPO-trained model in following human preferences, but it’s much simpler to implement. It doesn’t require sampling from the model and calculating advantages in a loop; you just need pairs of outputs with a preference label.

Some advantages of DPO:
- It doesn’t explicitly use a reward model during training (though in practice you need one to get the comparisons, or you already have the comparisons from human data).
- It avoids the instability of RL and can be done with standard gradient descent (it’s like doing a form of pairwise logistic regression).
- It’s computationally simpler: you can assemble a dataset of preferred vs dispreferred output pairs and just fine-tune the model on that dataset.

We can also see DPO as treating the problem as “the model should classify which output is better.” Because the model is generative, making $\pi_\theta(y^+|x)$ higher than $\pi_\theta(y^-|x)$ across all such pairs essentially tunes the model to generate $y^+$-like outputs more frequently.

It’s interesting to note that DPO’s derivation leverages the assumption that the original model’s distribution $\pi_0$ can act like a prior. DPO doesn’t throw away the original knowledge; it’s adjusting it by the human preferences. In a way, it’s doing the same thing as RLHF (which also often includes a term to keep the new policy close to the original), but baked into a single loss function.

For a beginner: imagine you have many examples of “In this situation, humans liked Response A more than Response B.” DPO says: “Okay, I will fine-tune the model so that it gives Response A a higher score than Response B for that situation.” Doing this for many examples teaches the model generally what humans prefer.

Comparing to PPO:
- PPO indirectly achieves the same effect but requires fiddling with reward scaling, advantage estimation, many iterations of sampling and optimizing.
- DPO just requires a dataset of comparisons (which we typically have from the same process as training the reward model).
- DPO doesn’t need to maintain a value function or sample multiple times; it’s a direct supervised learning approach to an RL problem.

The research found DPO to be competitive with PPO on tasks like summarization and dialogue alignment, suggesting it’s a promising alternative. However, it’s fairly new and being studied further. It simplifies alignment work because practitioners can avoid the RL part, which is often the trickiest and most finicky stage.

In summary, Direct Preference Optimization is a method that bypasses reinforcement learning by converting preference data into a direct training objective for the language model. It seeks to combine the strengths of the original model with the feedback data in a single, stable training phase. If PPO was like using a trial-and-error loop to gradually reinforce good behavior, DPO is like jumping straight to fitting the model to what’s considered good vs bad. It’s a great example of how insights from one approach (RL) can inform a simpler approach that achieves a similar outcome.

\section{Language Models as Reward Models: Concept and Future Implications}
In the context of alignment and fine-tuning, an intriguing idea has emerged: using large language models themselves as reward models or judges. We’ve seen that to align a model with human preferences, we often train a separate reward model on human data. But what if the language model could directly model the reward? Or put differently, what if a language model could be used to evaluate other outputs?

There are a few angles to “language models as reward models”:
\begin{itemize}
    \item \textbf{Implicit reward in the base model:} The DPO approach suggested that the original language model $\pi_0$ contains a lot of information that can act like a reward prior. Essentially, $\log \pi_0(y|x)$ could be seen as a part of the reward (language models assign probabilities to outputs – those probabilities reflect how “natural” or likely the output is). If we combine that with a learned human preference term, we get something like $\log \pi_\theta(y|x) \approx \log \pi_0(y|x) + \text{(preference term)}$. In other words, a well-trained base LM might already “know” that certain outputs are generally better (grammatically, factually) via its likelihood. So one could imagine using the language model’s own scoring as part of the reward signal.
    \item \textbf{Using a strong language model to judge:} Recently, researchers have experimented with using GPT-4 or other very advanced models to serve as a proxy for human judgment. For instance, one can prompt a model: “Here is a question and two answers. Which answer is better?” If the model is sufficiently aligned and powerful, it might make similar judgments as a human. This is known as \textbf{AI feedback} or using AI as a reward model. The advantage is that it’s cheaper and faster than gathering human labels for every new fine-tuning task. If you trust the model’s ability to evaluate, you can use it to generate a lot of comparison data. This has been tried in some works (like using GPT-4 to fine-tune a smaller model by labeling responses, sometimes called “reinforcement learning from AI feedback, RLAIF”).
    \item \textbf{Language model self-critiquing:} Approaches like Anthropic’s Constitutional AI \cite{bai2022} let a model generate outputs and then critique them based on a set of principles (essentially the model is used to measure how well outputs follow the principles). That critique can be seen as a reward signal. In a sense, the language model is being used to approximate a reward function (the degree to which the output violates or follows the written constitution).
    \item \textbf{Unification of policy and reward:} The phrase “Your Language Model is Secretly a Reward Model” from the DPO paper hints at a future where the line between the policy (the one generating answers) and the reward model might blur. Perhaps large models can internally estimate human preference as they generate text. If a model can predict “what would a human think of this response?” accurately, it could steer itself toward better responses without needing an external reward model.
\end{itemize}

What are future implications of treating language models as reward models?
- It could greatly streamline the alignment process. If we don’t have to train a separate reward network and run RL, and can instead rely on large models’ evaluation capabilities, we can more quickly fine-tune or even prompt models to behave well.
- One day, a language model might be able to improve itself by evaluating its own outputs. For example, it could generate several candidate answers to an instruction, then internally “think” or “vote” on which is best (using an internal reward heuristic), and output that. In some sense, this is already happening in techniques like “chain-of-thought” prompting where the model is asked to critique or refine its answer.
- Using language models as reward models also ties into safety considerations: We must ensure the model’s notion of “reward” aligns truly with human values and not some proxy that can be exploited. If a model learns to please an AI judge that isn’t perfectly aligned with humans, it might still output things humans don’t actually want (but the AI judge does). This is called the alignment of the reward model. If the AI judge is a large model that itself was aligned using human data (like GPT-4, presumably aligned), then we are stacking alignment on alignment.
- Another implication: scalability. Human feedback is a bottleneck. If AI models can serve as automatic judges, we can create massive amounts of training data for preferences without direct human labor. For instance, you could generate thousands of summaries and have a model pick the best ones to train a better summarizer.
- There’s also research indicating that as models get more capable, they might develop a form of “knowledge” of ethics or human values just from their training data. If that’s the case, perhaps future language models could have a built-in “conscience” scoring mechanism (hard to measure, but conceptually) that we could tap into instead of training separate classifiers.

One example of using a model as a reward model is the idea of “GPT-4 as a judge for fine-tuning ChatGPT.” If GPT-4 consistently gives high ratings to certain types of responses, we train ChatGPT to produce those. This is essentially using GPT-4’s complex understanding as the reward model. This approach has to be careful – we don’t want to oversimplify alignment to just mimic a bigger model, but it’s a powerful tool.

In reinforcement learning terms, language models as reward models is like having the critic be another model. If the critic is good, the actor (policy model) can learn quickly. If the critic is flawed, the actor will learn bad habits. So a lot of future work will likely focus on how to ensure these AI-based reward models truly match human intent (maybe by periodically auditing them with real human feedback or mixing the two).

Looking further ahead: if we manage to align models well and they truly understand our preferences, one could envision that the distinction between “model doing task” and “model checking task” goes away – a sufficiently advanced model might do both in one go. For example, a single model could be prompted with a request and, using its knowledge, just directly give a response it knows is helpful and aligned (because it knows what we'd prefer without needing explicit feedback). That’s the ideal scenario: the model inherently acts as if it has an internal reward for helping us, rather than us externally imposing it.

In conclusion, the concept of language models as reward models is about leveraging the models’ own intelligence and knowledge to evaluate outputs, thereby reducing reliance on separate systems or human input. It’s an active research area and an exciting one: it suggests a future where AI can align AI, under human guidance – a sort of bootstrap of alignment. This could make developing helpful and safe AI assistants much more efficient. But careful oversight will be needed to ensure the “AI feedback” remains grounded in actual human values. It’s a bit like training a student to grade their own exam correctly; if you succeed, the student can largely guide their own learning, but you need to verify that their grading criteria are truly correct and not self-serving.

\chapter{Object-Oriented AI Development Based on MCP}

\section{Introduction}

In recent years, artificial intelligence (AI) systems have become increasingly modular and interconnected. This shift has emerged because people want AI models not only to solve singular tasks but also to coordinate multiple capabilities, access external tools or data sources, and support complex workflows. Creating such systems can be challenging if the design is not organized from the start, which is why an object-oriented mindset is so valuable.

In this chapter, we will investigate three core ideas:

\begin{itemize}
  \item \textbf{Model Context Protocol (MCP).} This is a newly introduced standard that defines how AI models can talk to external resources in a reliable and consistent manner.
  \item \textbf{Hierarchical Ontologies.} These help manage different types of data, tasks, or concepts in a structured way, particularly when dealing with multimodal inputs like text, images, and audio.
  \item \textbf{LLMs as AI Engines.} Large Language Models can serve as central orchestrators of tool usage and data flow, much like a game engine coordinates graphics, physics, and audio in video games.
\end{itemize}

In the sections that follow, we delve into each concept in detail, always paying attention to how they reinforce each other in an object-oriented approach.

\section{Object-Oriented Programming (OOP)}
Learning to code involves understanding fundamental concepts that guide how we structure, organize, and collaborate on software projects. In particular, Object-Oriented Programming (OOP) helps us break down complex problems into manageable parts, and version control systems like Git enable teams (and individuals) to work together smoothly. Let us walk through these ideas step by step.

\subsection{Classes and Objects}
In OOP, \textbf{classes} act like blueprints that define a type of object. These blueprints outline the properties (data) and behaviors (functions or methods) that the objects created from that class will have. When we create an actual, usable instance of a class, we call it an \textbf{object}.

\paragraph{Example}
Imagine a ``Car'' class specifying attributes like color and speed, and methods such as \texttt{accelerate()} or \texttt{brake()}. Any individual car on the road is an object, created using this blueprint.

\subsection{Data Abstraction and Encapsulation}
Two foundational OOP concepts that help keep code organized and secure are data abstraction and encapsulation.

\paragraph{Data Abstraction}
Data abstraction focuses on exposing only the essential features of an object, hiding the unnecessary inner details. You see only what you need, which keeps things simpler.

\paragraph{Encapsulation}
Encapsulation bundles the data (attributes) and the methods (functions) that operate on that data. By placing data inside the class, we ensure only those methods in the class can directly access and modify it. This shields the data from unintended changes and makes the code more robust.

\subsection{Inheritance}
Inheritance allows us to create a new class based on an existing class, so we can reuse code rather than writing it from scratch.

\paragraph{Example}
If you have a ``Vehicle'' class with shared characteristics (wheels, seats), you can create a ``Car'' class that inherits from ``Vehicle'' and adds features (such as a trunk). This makes the code easier to maintain and expand.

\subsection{Polymorphism}
\textbf{Polymorphism} enables a function or method to behave differently depending on the context or the type of data it is handling. In simpler terms, one function name can perform different tasks.

\paragraph{Example}
A function named \texttt{draw()} could display a circle if it is drawing a circle object, or a rectangle if it is drawing a rectangle object. Each object can respond in its own way.

\subsection{Benefits of Object-Oriented Programming}
Object-oriented programming simplifies both the creation and maintenance of software projects:
\begin{itemize}
    \item It makes development more organized and easier to manage.
    \item It supports data hiding, which is good for security.
    \item It is well-suited to solving real-world problems by breaking them into logical objects.
    \item It encourages code reuse, reducing repetitive tasks.
    \item It allows for more generic code that can handle various data without rewriting.
\end{itemize}

\subsection{Example: Linking to Real-World Projects}
One way to strengthen your understanding of these concepts is by exploring real project code. For instance, the \href{https://github.com/karpathy/minGPT}{minGPT repository by Karpathy} demonstrates how object-oriented principles and clean coding can come together in practice. Browsing open-source projects like this illustrates how classes, objects, inheritance, and polymorphism appear in actual programs.

\subsection{Why an Object-Oriented Mindset?}

Object-oriented design emphasizes modularity, encapsulation, and inheritance. In AI, these principles prove extremely useful because:

\begin{itemize}
  \item \textbf{Interchangeability of Components.} We can swap out an existing model with a new one or replace a tool with an updated version, provided we maintain the same interface.
  \item \textbf{Clear Lines of Responsibility.} One module (or class) can handle data retrieval, another can handle data transformations, and a third can execute a specialized AI inference task. Debugging and maintenance become simpler when each module has a clear purpose.
  \item \textbf{Security and Access Control.} By defining interfaces precisely, we can ensure only authorized modules can invoke certain actions, which is essential for real-world enterprise or production settings.
\end{itemize}

Developers familiar with classical software engineering will see parallels between standard object-oriented patterns and AI systems. The difference is that now we integrate large language models or advanced neural networks as part of these components, which introduces new design considerations and complexities.

\section{Model Context Protocol (MCP)}

\subsection{Overview and Motivation}

As AI usage grows, we frequently want models to incorporate external or “live” information. Historically, this might have been handled by building custom integrations for each tool or data source. However, such an approach is not scalable. For example, a chatbot that needs stock market data, weather data, or access to the user’s documents would require separate code to handle each integration. This leads to repetitive development and maintenance overhead.

Anthropic introduced the Model Context Protocol (MCP) to address this issue. MCP is an attempt to provide a universal interface that AI models can use to request data or actions from an external service. One might consider it analogous to the way TCP/IP standardized networking, except that here the focus is on bridging AI clients with specialized services.

\subsection{Core Concepts of MCP}

\begin{itemize}
  \item \textbf{Client–Server Interaction.} In MCP, an AI model or AI application acts as the client, and each external resource (such as a database, a search engine, or a specialized API) is an MCP server. The server publishes the available actions or data endpoints, and the client calls them when needed.
  \item \textbf{Contextual Calls.} The protocol encourages sending along relevant context with each request. For instance, you might include the user’s query, a short description of the data needed, or a reference to prior steps in a conversation, so the server can respond accurately and keep track of relevant session information.
  \item \textbf{Modular Connectors.} An MCP server can be written once, for a particular service, and then reused by multiple AI agents. This is comparable to how device drivers in an operating system can be shared, which reduces duplication.
\end{itemize}

\subsection{Benefits for AI Systems}

\textbf{Modularity} is the primary benefit. By adopting MCP, you are decoupling the AI model’s logic from the specifics of how it interacts with external data. This decoupling means:

\begin{itemize}
  \item \textit{Less code duplication.} If many AI applications need to fetch weather data, a single MCP server can be used repeatedly.
  \item \textit{Greater consistency.} Standardizing the interface ensures that requests and responses follow a predictable format, making the system easier to debug.
  \item \textit{Stronger security.} Access rules can be enforced at the protocol layer. For instance, if the AI should not be allowed to modify certain documents, the MCP server can reject any write operations to that document, regardless of how the model was prompted.
\end{itemize}

\subsection{Illustrating an MCP Workflow with Text}

Since we are not using figures, we can walk through a hypothetical scenario in text form. Suppose a user says to an AI assistant, “Please find the most recent articles on renewable energy in my academic database, summarize them, and send me a final combined document.”

The AI assistant can break this instruction into steps, each implemented by calls to MCP servers:

\begin{enumerate}
  \item \textbf{Database Access:} The AI assistant calls an MCP server that connects to the user’s academic database. It requests all articles tagged with “renewable energy,” and the server returns the relevant references or article abstracts.
  \item \textbf{Summarization Service:} The AI assistant then calls a different MCP server that specializes in text summarization. It provides the abstracts or document texts as input. The summarization server returns concise summaries of each article.
  \item \textbf{Document Assembly:} The AI assistant may have another MCP server for file management (for example, generating a PDF or text file). It calls that server to assemble the collected summaries into a single aggregated document.
  \item \textbf{Final Action:} The AI assistant notifies the user that the task is complete and provides them with a link or copy of the assembled document.
\end{enumerate}

In this scenario, each external service is developed and maintained independently of the AI assistant. The assistant does not care how the database is structured internally or how the summarizer algorithm works. It only knows the MCP calls that are available and can be made, which fosters a clean separation of concerns.

\subsection{Example: Minimal Pseudocode for MCP Calls}

We can imagine a simple Python-like interface for making MCP requests. Suppose we have a client class:

\begin{lstlisting}[language=Python]
class MCPClient:
    def __init__(self, server_url):
        self.server_url = server_url

    def call_method(self, method_name, parameters):
        # This function might send an HTTP request or use another protocol
        # The server interprets the request, performs the action, and returns JSON
        response = send_request(self.server_url, method_name, parameters)
        return response

# In practice, you'd expand this to handle authentication, error handling, etc.
\end{lstlisting}

A typical call might look like:

\begin{lstlisting}[language=Python]
# Example usage
if __name__ == "__main__":
    # Suppose we want a summarization service
    summarizer = MCPClient("https://summarize.example.com/mcp")
    
    text_data = {
        "text": "A long text about renewable energy and new findings..."
    }
    summary = summarizer.call_method("summarize_text", text_data)
    print("Summary of the text:")
    print(summary)
\end{lstlisting}

Although this is a simplified illustration, it captures the essence of how a client might make MCP calls to a server.

\section{Hierarchical Ontology for Multimodal Systems}

\subsection{What is an Ontology?}

In AI, an ontology is a structured representation of concepts and their relationships. Traditional ontologies can be complex, involving formal logic, taxonomies, and metadata. For many practical AI applications, we want something slightly simpler but still beneficial: a hierarchical structure that helps us categorize data, tasks, and domain concepts.

\subsection{Why Hierarchy Matters for Multimodality}

Modern AI often deals with multiple data types, such as text, images, audio, and even videos or sensor readings. Each data type carries unique metadata and requires different preprocessing steps. By placing them in a well-defined hierarchy, we can:

\begin{itemize}
  \item \textbf{Keep track of shared properties.} If we consider audio data in a single parent class, then properties like sample rate, duration, or waveform type can be centralized and inherited by more specific audio subclasses.
  \item \textbf{Integrate multiple modalities.} Data that represents the same concept, such as “cat,” might appear in text, images, or audio (like meowing). In a hierarchical ontology, we can unify them under one node labeled “Cat” and link relevant sub-nodes or attributes.
  \item \textbf{Add tasks systematically.} Tasks like “translation” or “summarization” might fall under a “language tasks” branch, whereas “image classification” and “object detection” would be under “vision tasks.” This structure helps an AI system decide which modules or methods to use for a given request.
\end{itemize}

\subsection{Illustrative Textual Example: Animal Hierarchy}

Suppose we need to categorize animals in a system that receives both text references and images. We might define a small ontology:

\noindent
\textit{Animal}\\
--- \textit{Mammal}\\
------ Dog\\
------ Cat\\
--- \textit{Bird}\\
------ Eagle\\
------ Sparrow

If a user provides an image of a cat, the system can store it under “Mammal $\rightarrow$ Cat.” If a user writes a sentence about dogs, that text is stored under “Mammal $\rightarrow$ Dog.” By doing so, we can unify or cross-reference them if the user later asks for “all mammal data” in the system.

\subsection{Class-Based Ontology in Simple Code}

We can represent a piece of data (for instance, text, image, or audio) with base and derived classes:

\begin{lstlisting}[language=Python]
class Data:
    def __init__(self, content):
        self.content = content

class TextData(Data):
    def __init__(self, text):
        super().__init__(text)

class ImageData(Data):
    def __init__(self, image_path):
        super().__init__(image_path)

class AudioData(Data):
    def __init__(self, audio_path):
        super().__init__(audio_path)

# Example usage
text_item = TextData("A cat is purring softly.")
image_item = ImageData("images/cat_photo.jpg")
audio_item = AudioData("audio/cat_sound.mp3")
\end{lstlisting}

These classes are intentionally minimal but illustrate a hierarchical relationship. More advanced versions might track metadata such as shape, dimensions, or source.

\subsection{Task Ontologies}

Similarly, we can categorize tasks:

\begin{lstlisting}[language=Python]
class Task:
    def __init__(self, name):
        self.name = name

class LanguageTask(Task):
    def __init__(self, name):
        super().__init__(name)

class VisionTask(Task):
    def __init__(self, name):
        super().__init__(name)

class TranslationTask(LanguageTask):
    def __init__(self):
        super().__init__("Translation")

class SummarizationTask(LanguageTask):
    def __init__(self):
        super().__init__("Summarization")

class ImageClassificationTask(VisionTask):
    def __init__(self):
        super().__init__("Image Classification")
\end{lstlisting}

This structure allows the system to more easily route tasks to appropriate components. For instance, if you receive an \texttt{ImageClassificationTask} object, you know it is a vision-related problem that expects image data as input and returns categorical labels.

\subsection{Real-World Examples of Ontologies}

\begin{itemize}
  \item \textbf{ImageNet and WordNet.} ImageNet organizes images based on the WordNet ontology, enabling neural networks to recognize thousands of object categories. WordNet’s structure clarifies how different nouns relate (is-a relationships, synonyms, etc.).
  \item \textbf{AudioSet.} Google’s AudioSet classifies hundreds of audio events and arranges them in a tree. Examples include broad classes like “Music” with subcategories for “Guitar” or “Piano,” and “Animal sounds” with subcategories for “Dog bark,” “Bird song,” and others.
  \item \textbf{Domain-Specific Hierarchies.} Many industrial applications (healthcare, finance) rely on custom ontologies that reflect domain knowledge, such as patient diagnosis codes, financial instruments, or regulatory compliance categories.
\end{itemize}

\subsection{Why This Matters for Object-Oriented AI}

When we discuss “Object-Oriented AI,” we are not just talking about code structure, but also conceptual design. A well-defined hierarchy of data and tasks forms the backbone of how your AI system conceptualizes the world. Each class or node in the hierarchy corresponds to something the system can recognize, process, or generate. This approach fosters consistency and scalability, because additions or modifications can slot neatly into the existing hierarchy.

\section{LLMs as AI Engines}

\subsection{Parallel with Game Engines}

Large Language Models (LLMs), such as GPT-4 or other advanced neural networks, can do more than just produce text. They can interpret instructions, reason about tasks, and coordinate a variety of operations in a chain of thought. This is reminiscent of a game engine, which manages various modules (graphics, physics, sound) to render a cohesive experience.

In an AI application:

\begin{itemize}
  \item The \textbf{LLM} acts as the central decision-maker, interpreting user queries and deciding the steps to solve them.
  \item A set of \textbf{tools} (APIs, specialized machine learning models, databases) are attached to the LLM. The LLM can call upon these tools when needed.
  \item The \textbf{context} or “world state” is a place to store memory or relevant information about the ongoing session or environment.
\end{itemize}

\subsection{Tool-Oriented LLM Workflows}

Frameworks such as \textbf{LangChain} have popularized the idea of letting an LLM select which tool to use on the fly. When the user’s question or task arrives, the LLM essentially creates a plan: it might call a search tool first, then parse results, and then respond with a final answer. Alternatively, it might do calculations with a calculator tool before finalizing its output.

\subsection{Function Calling in Modern LLMs}

OpenAI’s function calling interface exemplifies how an LLM can dynamically invoke a set of predefined functions. Each function has a name and a schema describing its parameters. The model decides, based on context, which function to call, if any, and how to populate the arguments:

\begin{itemize}
  \item If the user says, “Retrieve the weather in Berlin next Tuesday,” the model can output something like: \texttt{\{"name": "get\_weather", "arguments": \{"location":"Berlin","date":"2025-04-01"\}\}}.
  \item The application code sees this JSON, calls the actual function \texttt{get\_weather} with the specified arguments, obtains the result, and feeds it back to the model. The model then composes the final response to the user based on that data.
\end{itemize}

This structured approach is easier to manage than purely text-based prompting, as the function calls are clearly delineated and typed.

\subsection{How an AI Engine Might Loop Internally}

Even without figures, we can describe the cycle in words. Each time the user interacts:

\begin{enumerate}
  \item The user’s request enters the system. Possibly it includes text, images, or other data. 
  \item The AI engine (where the LLM and orchestrator logic resides) checks if the request can be answered directly. If not, the LLM identifies which tool or service must be called.
  \item The engine calls that service, possibly using MCP if the service is external, or function calling if it is local. 
  \item The tool returns the result, which is fed back into the LLM. 
  \item The LLM updates its internal reasoning or “chain of thought” and then either calls another tool or produces a final answer. 
  \item That final answer is returned to the user, possibly along with logs of what steps were taken.
\end{enumerate}

\subsection{Code Snippet: LangChain-like Workflow}

A small sketch of how a chain might look in pseudo-code:

\begin{lstlisting}[language=Python]
class LLMEngine:
    def __init__(self, language_model, tools):
        """
        language_model: the LLM (e.g., GPT-4) interface
        tools: a dict of function_name -> function_callable
        """
        self.language_model = language_model
        self.tools = tools

    def handle_request(self, user_input):
        # Possibly a loop here:
        # 1. Send user_input and any relevant context to the LLM
        # 2. LLM decides if it needs to call a tool
        # 3. If yes, parse which tool to call and with what arguments
        # 4. Execute tool, gather result, feed it back to LLM
        # 5. Continue until final answer is generated
        return "Result or answer from the LLM"

# Example usage
if __name__ == "__main__":
    # Suppose we have a mock language model
    engine = LLMEngine(
        language_model="SomeLLM",
        tools={
            "web_search": lambda query: "Fake search results for " + query,
            "calculator": lambda expr: eval(expr)
        }
    )

    user_query = "What is 52 * 2?"
    final_answer = engine.handle_request(user_query)
    print(final_answer)
\end{lstlisting}

In a real application, the LLM’s logic would produce something like, “I need to call the calculator tool with the expression '52 * 2'.” The rest of the system would interpret that as a request to run \texttt{eval("52*2")}, return 104, feed it back to the model, and so forth.

\subsection{Benefits of the Engine Approach}

\begin{itemize}
  \item \textbf{Organization.} It is clear which component does what. The LLM is for reasoning and text generation, while specific tasks are delegated to specialized tools.
  \item \textbf{Modularity.} New tools can be added by simply registering them with the engine, without major changes to the rest of the code.
  \item \textbf{Explainability.} If the user wants to know how the system arrived at an answer, you can show them the sequence of tool calls. Each step is traceable.
  \item \textbf{Scalability.} Multiple LLMs or sub-agents can be introduced, each specialized in a domain. The engine coordinates them, merging their outputs.
\end{itemize}

\subsection{Multiple Agents in One Engine}

Sometimes we want more than one AI agent operating in the same environment. For instance, one agent might handle knowledge retrieval, another might handle creative writing, and a third might handle code generation. A higher-level engine can orchestrate these agents, passing intermediate results between them as needed. Each agent might also have its own set of tools or ontological knowledge relevant to its domain.

This setup resembles multi-agent systems in robotics or simulations. The key difference is that these agents exchange text or data rather than physically interacting, although one could integrate real-world sensors or actuators if desired.

\section{Implementation Tips and Considerations}

\subsection{Define Clear Interfaces}

Both MCP and any local function-calling system rely on consistent, well-documented interfaces. Each tool or MCP endpoint should have:

\begin{itemize}
  \item A concise name and a short description.
  \item A parameter list explaining input fields.
  \item An explanation of the returned data format.
\end{itemize}

Following these guidelines reduces confusion and errors, especially if different developers are building different parts of the system.

\subsection{Use Ontologies Early}

Designing an ontology from the start, even a small one, helps avoid ad-hoc categories that can become inconsistent. Think carefully about the major data types or tasks that appear in your application, and place them in a hierarchy. Over time, you can refine or extend this structure rather than restarting from scratch.

\subsection{Permission and Security Boundaries}

When an AI can call external tools, especially tools with write access (like sending emails or modifying files), it is important to define safety measures. For instance, you might require explicit user confirmation for certain actions. This prevents the AI from taking unwanted or harmful steps if it is given a misleading or malicious prompt.

\subsection{Controlling Context Size}

LLMs have context window limits. If your AI agent constantly accumulates new information, you may run into token length constraints. Techniques like retrieval augmentation, chunking, or summarization can help manage this. Ensuring the most relevant details are available, while discarding or summarizing older context, is often necessary in long-running systems.

\subsection{Iterative Development and Testing}

Complex AI setups can become difficult to debug. It helps to record each request and response in the chain of actions. For example, keep track of:

\begin{itemize}
  \item The user’s query.
  \item The LLM’s internal reasoning step or plan (if accessible through chain-of-thought or a partial logging approach).
  \item Tool calls made (method name, parameters, results).
  \item Final answer or action.
\end{itemize}

These logs help you quickly diagnose which step might have failed or which tool returned unexpected data.

\section{Conclusion and Future Directions}

We have explored three interlocking pillars for building object-oriented AI systems:

\begin{itemize}
  \item \textbf{Model Context Protocol (MCP).} A standardized way for AI models to communicate with external services. This fosters modularity, reduces duplication, and simplifies security.
  \item \textbf{Hierarchical Ontologies.} A structured approach for organizing data and tasks, especially useful in multimodal settings where text, images, and audio must be seamlessly integrated.
  \item \textbf{LLMs as AI Engines.} A perspective that treats large language models as central orchestrators that can call tools, manage states, and cooperate with other models or agents. This design parallels how game engines coordinate various subsystems.
\end{itemize}

Putting these ideas together results in systems that are significantly more capable than a simple stand-alone AI model. The object-oriented paradigm provides boundaries, clarity, and extensibility. Developers can add new modules (such as a speech recognizer, an image classifier, or a robotics controller) without overhauling the entire architecture. Each module is treated as an object or component, with the LLM deciding how and when to use it.

Future directions in this domain may include:

\begin{itemize}
  \item Wider adoption of standardized protocols like MCP across different cloud services, so AI can quickly interface with popular platforms.
  \item More advanced hierarchical ontologies that incorporate temporal or causal relationships, allowing AI to reason about sequences of events or evolving data over time.
  \item Increased focus on multi-agent frameworks, where multiple specialized LLMs or AI modules interact within one environment. 
  \item Research on human-AI collaboration mechanisms that incorporate these object-oriented principles to ensure transparency, trust, and accountability.
\end{itemize}

By learning and applying these approaches, you will be better equipped to build, maintain, and extend AI systems that must operate in real-world contexts where data, tasks, and integrations can constantly change.

\chapter{AI and the Metaverse:
Digital Twins, Egocentric Multimodal AI, and Decentralized GPU Clusters}
\section{Introduction}

The term \emph{metaverse} is commonly used to describe a network of immersive virtual environments that blend digital and physical realities. It promises opportunities for interactive experiences and social connectivity beyond traditional screens. The arrival of advanced artificial intelligence (AI) and augmented reality (AR) technologies means these virtual worlds are not only visually compelling, but also deeply intelligent and context-aware. This chapter explores three key themes central to the convergence of AI and the metaverse: digital twin-based physical AI, egocentric multimodal AI agents for AR glasses, and decentralized GPU clusters for energy-efficient training and inference.

Digital twins are virtual representations of physical assets or systems that continuously synchronize with real-world data. When enhanced by AI, digital twins become powerful tools for monitoring, predicting, and optimizing the behavior of their real-world counterparts. Egocentric multimodal AI agents refer to intelligent assistants embedded in AR glasses or similar wearable devices. These agents perceive the world from the user's point of view and help facilitate natural, context-driven interactions. Decentralized GPU clusters address the growing demand for massive compute power by distributing AI workloads across many nodes, often located near the data source, to achieve efficiency and privacy gains.

After reading, one should understand how digital twins simulate real-world environments in virtual spaces, how AR glasses can benefit from AI that sees and hears from a user's perspective, and why decentralization of GPU resources might be essential for the next phase of AI-driven immersive applications.

\section{Digital Twin-Based Physical AI}

\subsection{What is a Digital Twin?}
A digital twin is a virtual representation of a real-world object, system, or environment. It uses sensor data, simulation models, and analytics to replicate the state and behavior of its physical counterpart in real time. For example, a manufacturing plant may have a digital twin that reflects the status of machines and assembly lines, allowing engineers to monitor performance, predict failures, and simulate improvements. 

In the metaverse context, digital twins provide a bridge between physical and virtual worlds. A city or factory can be cloned inside a virtual space where AI can run simulations to optimize traffic flow, predict machinery breakdown, or schedule maintenance. Once confident with the simulation, the results can be applied back to the real system. This cyclical process accelerates innovation and reduces risk, because mistakes in the virtual domain do not harm the physical world.

\subsection{Why Combine AI with Digital Twins?}
AI augments digital twins in several ways. First, it interprets and learns from continuous data streams that update the twin. This allows real-time detection of anomalies or issues. Second, AI supports predictive analytics. By modeling trends and patterns in the twin, AI can estimate when a part will fail or how a design change will affect performance. Third, AI enables rapid simulation of \emph{what-if} scenarios. A city planning department could test new road layouts in a virtual model before investing in costly infrastructure changes. These benefits increase efficiency and unlock the potential of \emph{continuous optimization} in many industries, including manufacturing, healthcare, and energy.

\subsection{Examples of Digital Twin Use}
\textbf{Manufacturing.} Large industrial companies are already using digital twins to monitor equipment health. A turbine might have sensors for temperature, vibration, and load, all feeding into its digital twin. An AI system processes those data to identify anomalies or predict maintenance schedules. This leads to less downtime and smoother operations.

\textbf{Smart Cities.} Urban planners and local governments build city-scale digital twins by integrating live traffic data, population movement, and environmental sensors. AI-driven simulations can then optimize traffic lights, reduce congestion, and plan routes for emergency services.

\textbf{Healthcare.} A hospital might create digital twins of its ICU beds, tracking patient status in near real time. AI systems predict patient deterioration before it becomes critical and suggest interventions. On an individual level, researchers are exploring the concept of patient digital twins, which could help doctors personalize treatment by simulating outcomes for each patient.

\textbf{Energy and Infrastructure.} Power grids and wind farms employ digital twins for each turbine or substation. AI continuously adjusts operating conditions to match demand and maximize efficiency. When a sensor indicates an unusual vibration, the twin and real turbine are flagged for inspection. This predictive maintenance prevents costly repairs and downtime.

\section{Egocentric Multimodal AI Agents for AR Glasses}

\subsection{First-Person Perspective AI}
As wearable devices and AR headsets become more widespread, AI is moving literally in front of our eyes. \emph{Egocentric AI} refers to intelligent systems that perceive the world from a user's own viewpoint. Traditional AI might analyze data from external, fixed-position cameras. By contrast, egocentric AI sees and hears what the user experiences in real time through the wearable device itself. This gives an assistant in AR glasses a deep understanding of context, including what objects the user looks at, what environment they are in, and what they are trying to do.

When a user asks, ``Where did I put my keys,'' the AI can look back at first-person video (if recording is enabled and privacy settings allow it) to find the moment the user placed the keys somewhere. If the user says, ``What is that painting?'' the AI can analyze the camera feed of whatever the user is currently viewing. This new perspective drastically changes how AI can respond to everyday tasks and information queries.

\subsection{Sensors and Multimodal Inputs}
AR glasses integrate multiple sensors that feed into an AI system:

\textbf{Eye-Tracking.} Cameras facing the wearer's eyes detect gaze. This lets the AI know exactly where the user is looking. Combined with voice commands, gaze tracking disambiguates references like ``What is that?'' The system identifies the object currently in focus.

\textbf{Hand-Tracking and Gestures.} Outward cameras track hand movements. If the user reaches for a knob, the AI can overlay instructions or warnings. If the user uses specific gestures, the system translates them into commands, like pinching to click or swiping to scroll virtual menus.

\textbf{Front-Facing Camera.} The wearable device captures a real-time view of the environment. Computer vision models detect people, objects, text, and events in the scene. If the user is in a kitchen, the AI might notice a stove is turned on or recognize cooking utensils.

\textbf{Microphones.} Voice commands are an important interface in AR glasses. The device also listens to ambient sounds, which can signal context such as alarms, music, or conversation. This may help the AI provide transcription, translation, or real-time assistance, such as clarifying what was just said in a busy environment.

\textbf{Spatial Mapping.} Depth sensors like LiDAR measure the geometry of surroundings. This helps place virtual overlays and also prevents collisions with real objects. AI can use these data to understand where the user is in a room and label important objects or pathways.

All these modalities combine, allowing the assistant to interpret user intent, environment, and objects in a holistic way. This sensor fusion represents the core of egocentric AI for AR, since no single data type can capture the full context.

\subsection{AI Models for Egocentric Intelligence}
Bringing this to life involves various AI components:

\textbf{Computer Vision.} Deep learning models recognize objects, faces, and text from first-person footage. They must be efficient enough to run in real time. This often involves specialized hardware or partial cloud offloading.

\textbf{Natural Language Understanding.} The device interprets user voice commands and dialogues, possibly aided by large language models. It can combine textual understanding with visual context to answer queries such as, ``Does this snack contain peanuts?'' while analyzing the label the wearer is looking at.

\textbf{Contextual Reasoning.} The system maintains an internal representation of the user's environment, tasks, and recent actions. If the user was assembling furniture, the AI might recall the user's last step and propose the next step, placing holographic instructions in the AR view.

\textbf{Interaction Design.} AR glasses must balance helpfulness with minimal intrusion. The system decides when to proactively offer information (for safety or assistance) or wait for a prompt. The user typically interacts via speech, gaze, or gestures instead of mouse and keyboard.

These techniques are under development by major companies like Meta (formerly Facebook), Microsoft, Apple, and various startups. Their shared objective is to make AR wearables into natural extensions of human capability rather than mere displays. 

\subsection{Industry Examples}
\textbf{Meta.} Meta's Reality Labs is researching \emph{egocentric perception} and has released the Ego4D dataset for first-person AI. Their smart glasses already support voice commands and limited scene understanding. Future releases will likely integrate advanced context awareness.

\textbf{Microsoft.} HoloLens 2 provides industrial use cases like remote assistance and step-by-step training. It uses hand-tracking, eye-tracking, and voice to let workers interact with digital overlays. AI is used for object recognition and for guiding users through tasks like machine repair.

\textbf{Apple.} Vision Pro is a mixed reality headset that uses eye and hand tracking as primary inputs. While the focus is on user interface, future iterations or rumored AR glasses may include real-time object recognition and an on-board AI assistant. 

\textbf{Others.} Magic Leap focuses on enterprise solutions. Snap Spectacles experiment with first-person video for social media. Google has shown prototypes of real-time translation glasses. Startups explore specialized uses like sports training, medical guidance, and more. 

\section{Decentralized GPU Clusters for Training and Inference}

\subsection{The Need for Scalable AI Computation}
As AI models grow in complexity, the hardware required to train them can be enormous. Traditional approaches rely on centralized cloud data centers that host thousands of graphics processing units (GPUs) in one location. While effective for certain tasks, centralized clouds can introduce high latency when users or data sources are geographically distant. They also create single points of failure and raise issues of data sovereignty when sensitive data must remain local.

\subsection{What is a Decentralized GPU Cluster?}
A decentralized GPU cluster distributes computing resources across many nodes in different locations. These nodes, which can be edge servers, on-premise servers, or individual personal computers with spare GPU capacity, are linked via the internet. An overarching scheduling or coordination layer assigns workloads so that multiple nodes cooperate to train AI models or serve inference. This approach is sometimes compared to older concepts like distributed computing (SETI@home) or peer-to-peer networks, but updated with modern GPU hardware and AI-specific frameworks.

\subsection{Benefits of Decentralization}
\textbf{Latency Reduction.} Placing AI services nearer to users means faster response times, critical for real-time interactions such as AR overlays or autonomous vehicles. If every inference request had to be sent to a remote data center, latency might be unacceptable in high-speed tasks.

\textbf{Bandwidth Efficiency.} Instead of sending large volumes of raw data to a single cloud, local edge devices can pre-process or perform inference directly. Only relevant summaries or model updates need to go back to any central server, reducing network loads.

\textbf{Better Resource Utilization.} Many GPUs sit idle when gamers are not using them or when research labs have off-peak hours. A decentralized approach could turn these idle resources into a global AI compute pool. This also helps smaller organizations collaborate to match large-scale compute capabilities.

\textbf{Data Privacy and Compliance.} With federated learning, data can remain on local nodes. Users or companies can contribute model updates rather than raw data. This aligns with regulations that mandate data residency in particular regions.

\textbf{Resilience.} If one node or data center fails, others can continue working. This distributed structure avoids single points of failure. 

\subsection{Key Technologies}
\textbf{Edge Computing.} Telecommunication companies and cloud providers are setting up mini-data centers or edge servers close to end-users. AR and VR data can be processed with minimal latency. Applications range from gaming to remote surgery.

\textbf{Blockchain and Distributed Ledgers.} Some networks use blockchain-based protocols to coordinate and incentivize participants who share their GPU power. Smart contracts may track contributions and distribute rewards. This aims to create a self-sustaining ecosystem that does not require trust in one central authority.

\textbf{Federated Learning.} Instead of moving data, federated learning sends model training tasks to each node. Only partial updates or gradients are aggregated, allowing multi-party collaboration without sharing private information. This method has been piloted in healthcare, where hospitals collaborate to improve diagnostic models.

\textbf{Zero Trust Security.} As participants may be geographically and institutionally diverse, a zero trust approach ensures each part of the network is authenticated and authorized. Encryption prevents data leaks. Malicious nodes can be detected and blocked without compromising the entire cluster.

\subsection{Case Studies}
Telecom operators are exploring decentralized inference for real-time analytics. A phone user running an AR application can offload some computations to a local edge server. This reduces network traffic and speeds up responses. Blockchain-based AI marketplaces, such as SingularityNET or DeepBrain Chain, propose token-based incentives for individuals to rent out GPU resources. Healthcare federated learning initiatives let hospitals create robust AI models for disease detection while respecting patient privacy. Projects combining these approaches show the potential for a new wave of distributed AI research.

\section{Discussion and Future Outlook}

By integrating AI with the metaverse, we expand digital experiences beyond simple virtual worlds. Physical systems gain richer modeling with digital twins, AR glasses become deeply attuned to the wearer's perspective, and GPU resources spread across networks can handle the heavy lifting for real-time AI. 

Though promising, there are challenges. Digital twins require continuous high-quality data, and not all systems are equipped with reliable sensors or data pipelines. Egocentric AI raises significant privacy questions, because first-person video and audio can capture sensitive personal details. Decentralized GPU clusters demand robust coordination and security protocols to ensure fairness, trust, and performance. Overcoming these technical and ethical obstacles will shape how quickly we see widespread adoption.

In the near future, we can expect further miniaturization of hardware, so that AR glasses become more comfortable and are able to run advanced AI on-device. The concept of a personal AI agent that sees the world as we do may become mainstream, assisting people at home and in the workplace. Likewise, digital twin platforms will expand from factories into healthcare, transportation, and city planning. Decentralized GPU networks may provide the computational backbone, helping to balance workloads and preserve data privacy. If these technologies evolve in a user-centered and responsible manner, the combination of AI and the metaverse will provide experiences that are practical, personalized, and immersive. 

\section{Conclusion}

This chapter introduced three pillars that connect AI research with the metaverse. First, digital twin-based physical AI leverages virtual replicas to help monitor and optimize real-world systems. Second, egocentric multimodal AI agents bring personal context and assistance to AR glasses. Third, decentralized GPU clusters address the growing computational demands of AI while reducing latency and respecting data privacy constraints.

An undergraduate or new learner in AI should now appreciate how these areas come together to shape next-generation platforms. Digital twins extend our ability to study and improve the physical realm. Egocentric AI fosters more natural human-computer interaction through a first-person perspective. Decentralized compute enables scalable and efficient AI services by sharing the load among many nodes. Collectively, these trends point toward a future in which physical and digital worlds merge seamlessly, aided by intelligent infrastructure that is globally distributed yet intimately personal.

\appendix

\chapter{How to Code}

\section{Introduction}
Learning to code involves understanding fundamental concepts that guide how we structure, organize, and collaborate on software projects. In particular, Object-Oriented Programming (OOP) helps us break down complex problems into manageable parts, and version control systems like Git enable teams (and individuals) to work together smoothly. Let us walk through these ideas step by step.

\section{Classes and Objects}
In OOP, \textbf{classes} act like blueprints that define a type of object. These blueprints outline the properties (data) and behaviors (functions or methods) that the objects created from that class will have. When we create an actual, usable instance of a class, we call it an \textbf{object}.

\paragraph{Example}
Imagine a ``Car'' class specifying attributes like color and speed, and methods such as \texttt{accelerate()} or \texttt{brake()}. Any individual car on the road is an object, created using this blueprint.

\section{Data Abstraction and Encapsulation}
Two foundational OOP concepts that help keep code organized and secure are data abstraction and encapsulation.

\paragraph{Data Abstraction}
Data abstraction focuses on exposing only the essential features of an object, hiding the unnecessary inner details. You see only what you need, which keeps things simpler.

\paragraph{Encapsulation}
Encapsulation bundles the data (attributes) and the methods (functions) that operate on that data. By placing data inside the class, we ensure only those methods in the class can directly access and modify it. This shields the data from unintended changes and makes the code more robust.

\section{Inheritance}
Inheritance allows us to create a new class based on an existing class, so we can reuse code rather than writing it from scratch.

\paragraph{Example}
If you have a ``Vehicle'' class with shared characteristics (wheels, seats), you can create a ``Car'' class that inherits from ``Vehicle'' and adds features (such as a trunk). This makes the code easier to maintain and expand.

\section{Polymorphism}
\textbf{Polymorphism} enables a function or method to behave differently depending on the context or the type of data it is handling. In simpler terms, one function name can perform different tasks.

\paragraph{Example}
A function named \texttt{draw()} could display a circle if it is drawing a circle object, or a rectangle if it is drawing a rectangle object. Each object can respond in its own way.

\section{Benefits of Object-Oriented Programming}
Object-oriented programming simplifies both the creation and maintenance of software projects:
\begin{itemize}
    \item It makes development more organized and easier to manage.
    \item It supports data hiding, which is good for security.
    \item It is well-suited to solving real-world problems by breaking them into logical objects.
    \item It encourages code reuse, reducing repetitive tasks.
    \item It allows for more generic code that can handle various data without rewriting.
\end{itemize}

\paragraph{Example: Linking to Real-World Projects}
One way to strengthen your understanding of these concepts is by exploring real project code. For instance, the \href{https://github.com/karpathy/minGPT}{minGPT repository by Karpathy} demonstrates how object-oriented principles and clean coding can come together in practice. Browsing open-source projects like this illustrates how classes, objects, inheritance, and polymorphism appear in actual programs.

\section{Version Control Systems (VCS)}
After structuring your code with OOP, the next essential step is keeping track of changes. A \textbf{version control system} like Git allows you to manage different versions of your project. This is especially useful when multiple people collaborate, but even solo developers benefit from being able to roll back to earlier versions.

\subsection{Personal and Organizational GitHub Pages}
GitHub is a popular platform for hosting and reviewing code. It can serve two main roles:
\begin{enumerate}
    \item \textbf{Personal GitHub Page}: A space for showcasing your own projects, reflecting your identity as a developer.
    \item \textbf{Organizational GitHub Page}: A shared space representing a company, research group, or community, which helps organize team-based or collective projects.
\end{enumerate}

\subsection{Basic Git Workflow}
To start using Git effectively, you should understand its core steps and commands:

\begin{enumerate}
    \item \textbf{Install Git and Create a GitHub Account:} Download Git and sign up on GitHub.
    \item \textbf{Open Your Terminal/Command Line:} Typing \texttt{git} should show available commands.
    \item \textbf{Configure Your User Name and Email:} 
    \begin{verbatim}
    git config --global user.name "Your Name"
    git config --global user.email "Your Email"
    \end{verbatim}
    \item \textbf{Create a New Repository on GitHub:} This is where your project files will live.
    \item \textbf{Clone an Existing Repository:} Use the \texttt{git clone} command to copy a repo locally.
    \item \textbf{Essential Git Commands:}
    \begin{itemize}
        \item \texttt{git status} : Shows the current state of your repository.
        \item \texttt{git add} : Stages changes you want to commit.
        \item \texttt{git commit} : Records staged changes to the project history.
        \item \texttt{git push} : Sends your commits to the remote repository.
        \item \texttt{git pull} : Fetches and merges changes from the remote repository.
        \item \texttt{git checkout} : Switches between branches or versions.
        \item \texttt{git merge} : Combines changes from one branch into another.
        \item \texttt{git fetch} : Retrieves changes from the remote but does not merge them.
        \item \texttt{git rebase} : An advanced method of integrating changes by rewriting commit history.
    \end{itemize}
\end{enumerate}

\section{Conclusion}
Coding is more than simply writing instructions for a computer. By embracing Object-Oriented Programming, you learn to break your code into logical, secure, and reusable pieces. Understanding and using Git keeps your projects organized and collaborative, whether you work alone or in a team. Together, these concepts form a solid foundation for anyone looking to write efficient, maintainable, and collaborative software. 

Feel free to explore open-source projects like the \href{https://github.com/karpathy/minGPT}{minGPT repository} for practical examples of OOP in action, and set up a personal GitHub page to share your own projects. With these skills and tools, you will be well on your way to coding more effectively and confidently.

\chapter{Exercise 1: Git}

\section{Introduction}

\subsection{What is Version Control?}
\begin{itemize}
  \item \textbf{Definition:} A system that records changes to files, enabling you to recall specific versions later.
  \item \textbf{Benefits:}
  \begin{itemize}
    \item Maintains a detailed history of changes.
    \item Simplifies collaboration for multiple contributors.
    \item Facilitates rollback to earlier file states when necessary.
  \end{itemize}
\end{itemize}

\subsection{Importance of Version Control}
\begin{itemize}
  \item Teams can work on the same code without overwriting each other's work.
  \item Every change is tracked, providing accountability and clarity.
  \item Prevents the confusion of multiple file versions like \texttt{project\_final\_v2\_backup}.
\end{itemize}

\subsection{Why Git?}
\begin{itemize}
  \item \textbf{Distributed Model:} Each user has a complete copy of the repository.
  \item Created by Linus Torvalds in 2005 to manage the Linux kernel.
  \item Widely adopted due to speed, flexibility, and strong community support.
\end{itemize}

\subsection{Git vs. GitHub}
\begin{itemize}
  \item \textbf{Git:} The version control software running on your local machine.
  \item \textbf{GitHub:} A platform that hosts Git repositories online, providing features like pull requests and issue tracking.
  \item Alternatives include GitLab and Bitbucket, but GitHub is very common for open-source.
\end{itemize}

\section{Installing Git and Setting Up a Local Repository}

\subsection{Installation}
\begin{itemize}
  \item \textbf{Windows:} Download ``Git for Windows'' from \href{https://git-scm.com}{git-scm.com}.
  \item \textbf{macOS:} Install via Homebrew (\texttt{brew install git}) or Xcode Command Line Tools.
  \item \textbf{Linux:} Use your package manager, e.g., \texttt{sudo apt install git}.
\end{itemize}
Check installation:
\begin{verbatim}
git --version
\end{verbatim}

\subsection{Configuration}
\begin{itemize}
  \item Set your username and email so that each commit is correctly attributed:
  \begin{verbatim}
  git config --global user.name "Your Name"
  git config --global user.email "your@email.com"
  \end{verbatim}
  \item Optionally configure a default editor:
  \begin{verbatim}
  git config --global core.editor "code --wait"
  \end{verbatim}
\end{itemize}

\subsection{Initializing a New Repository}
\begin{itemize}
  \item Create or navigate to an empty folder, then run:
  \begin{verbatim}
  git init
  \end{verbatim}
  \item A \texttt{.git} folder is created to store project history.
  \item \texttt{git status} checks which files are tracked/untracked and shows repository status.
\end{itemize}

\section{Basic Git Workflow}

\subsection{The Edit-Stage-Commit Cycle}
\begin{itemize}
  \item \textbf{Working Directory:} Your current files/folders.
  \item \textbf{Staging Area:} An intermediate step before finalizing changes in a commit.
  \item \textbf{Repository:} Stores the official commit history.
\end{itemize}

\subsubsection{Add and Commit}
\begin{itemize}
  \item Create a file (e.g., \texttt{hello.txt}) and stage it:
  \begin{verbatim}
  git add hello.txt
  git commit -m "Add hello.txt file"
  \end{verbatim}
  \item \texttt{git log --oneline} shows a compressed history of your commits.
  \item Use clear, descriptive commit messages for easier tracking.
\end{itemize}

\subsubsection{Working with Remotes}
\begin{itemize}
  \item \textbf{Remote:} A copy of the repository hosted on another server (e.g., GitHub).
  \item \texttt{git remote add origin <URL>} connects your local repo to a remote.
  \item \texttt{git push} uploads local commits to the remote; \texttt{git pull} retrieves and merges any changes from it.
\end{itemize}

\subsubsection{Cloning a Repository}
\begin{itemize}
  \item \textbf{Clone:} Download an existing remote repository locally.
  \begin{verbatim}
  git clone https://github.com/user/repo.git
  \end{verbatim}
\end{itemize}

\noindent
\textbf{Exercise:}
\begin{enumerate}
  \item Create a new GitHub repository.
  \item Clone it locally.
  \item Add and commit a file, then push to see the changes on GitHub.
\end{enumerate}

\section{Branching and Merging}

\subsection{Why Use Branches?}
\begin{itemize}
  \item Branches let you work on a new feature or fix independently.
  \item Minimizes disruption to the main code while you're experimenting or developing.
\end{itemize}

\subsection{Creating and Switching Branches}
\begin{verbatim}
git checkout -b feature1
\end{verbatim}
\begin{itemize}
  \item \texttt{-b} creates a new branch, then \texttt{checkout} switches to it.
  \item Commit your changes on \texttt{feature1} without affecting \texttt{main}.
\end{itemize}

\subsection{Merging Branches}
\begin{verbatim}
git checkout main
git merge feature1
\end{verbatim}
\begin{itemize}
  \item Combines changes from \texttt{feature1} into \texttt{main}.
  \item If no conflicting edits, this might be a fast-forward merge.
  \item If there is a conflict, Git will require manual resolution.
\end{itemize}

\subsection{Resolving Merge Conflicts}
\begin{itemize}
  \item Occur when the same lines differ between branches.
  \item Conflict markers appear:
\begin{verbatim}
<<<<<<< HEAD
(your changes)
=======
(other changes)
>>>>>>> feature1
\end{verbatim}
  \item Decide what to keep or combine, then stage and commit the resolution.
\end{itemize}

\noindent
\textbf{Exercise:}
\begin{enumerate}
  \item Create and switch to a branch called \texttt{experiment}, make a commit.
  \item Merge it into \texttt{main}.
  \item Push \texttt{main} to GitHub and confirm the changes are there.
\end{enumerate}

\section{GitHub: Pull Requests and Issues}

\subsection{Pull Requests (PRs)}
\begin{itemize}
  \item \textbf{Definition:} A mechanism to propose merging one branch into another, often used for code review.
  \item \textbf{Typical Steps:}
  \begin{enumerate}
    \item Push a branch to the remote repository.
    \item On GitHub, open a PR by clicking ``Compare \& pull request.''
    \item Describe your changes; reference issues if relevant.
    \item Teammates review and comment.
    \item Merge the PR into \texttt{main}.
  \end{enumerate}
\end{itemize}

\subsection{Issues}
\begin{itemize}
  \item Used to track bugs, enhancements, or discussions.
  \item Can be referenced in commits (\texttt{Fixes \#3}) to close them automatically upon merging.
  \item Can include labels, milestones, assignees for organizing project tasks.
\end{itemize}

\noindent
\textbf{Optional Collaboration Exercise:}
\begin{enumerate}
  \item Open a GitHub issue describing a small feature.
  \item Create a new branch locally to address the feature.
  \item Commit and push, then open a pull request mentioning the issue (e.g., ``Closes \#1'').
  \item Merge the PR to see the issue close automatically.
\end{enumerate}

\section{Advanced Topics: Rebasing and Workflows}

\subsection{Rebasing}
\begin{itemize}
  \item \textbf{Concept:} Moves the base of your branch to a new starting point, creating a linear history.
  \item \textbf{Usage Example:}
  \begin{verbatim}
  git checkout feature
  git rebase main
  \end{verbatim}
  \item Useful for avoiding merge commits but rewrites commit history, so use carefully.
\end{itemize}

\subsection{Common Git Workflows}
\begin{itemize}
  \item \textbf{Feature Branch Workflow (GitHub Flow)}: Simple approach, each feature on its own branch, then merged via PR.
  \item \textbf{Gitflow}: More structured with separate branches for \texttt{develop}, \texttt{main}, features, and releases.
  \item \textbf{Fork \& Pull}: Common in open-source; contributors fork the repository and submit PRs from their fork.
\end{itemize}

\subsection{Best Practices}
\begin{itemize}
  \item Commit frequently with descriptive messages.
  \item Pull or fetch often to stay updated on team changes.
  \item Keep \texttt{main} stable by developing features in separate branches.
  \item Use pull requests for code reviews and structured merging.
\end{itemize}

\section{Homework Assignment: ``Git in Practice''}

\noindent
\textbf{Objective:} Gain hands-on experience with Git's core features, GitHub collaboration, branching, pull requests, issues, and an introduction to rebasing.

\begin{enumerate}
  \item \textbf{Create a Repository on GitHub} \\
  Initialize it (no README), then clone locally.
  \item \textbf{Initial Commit} \\
  Create and commit a \texttt{README.md}, then push to \texttt{main}.
  \item \textbf{Branch and Pull Request} \\
  Create a \texttt{dev} branch, add or modify files, commit, push the branch, and open a pull request to merge \texttt{dev} into \texttt{main}.
  \item \textbf{Issue and Fix} \\
  Open an issue (e.g., ``Typo in README''), create a fix branch referencing the issue, merge, and confirm the issue is closed.
  \item \textbf{Rebase Practice} \\
  Simulate two branches diverging, then rebase one onto \texttt{main}. Resolve conflicts if they occur.
  \item \textbf{Reflection} \\
  Write a short note (200--300 words) on your experience, what you learned, and any challenges faced.
\end{enumerate}

\section{Conclusion}

\noindent
By completing these exercises, you will:
\begin{itemize}
  \item Understand Git’s core commands for day-to-day use.
  \item Gain confidence creating, merging, and managing branches.
  \item Learn to collaborate using GitHub’s pull requests and issues.
  \item Explore rebasing and the concept of rewriting Git history.
\end{itemize}

\noindent
Over time, practice will help you master more advanced features and workflows. Good luck, and enjoy exploring version control with Git!

\chapter{Exercise 2: Python Basics}

\section{Introduction}
Python is a high-level, interpreted programming language known for its readability and versatility. It is widely used in various fields such as web development, data science, machine learning, scripting, and more. Python's large standard library and active community make it a powerful tool for both beginners and experienced developers.

Key features of Python include:
\begin{itemize}
    \item Simple, readable syntax
    \item Interpreted, cross-platform execution
    \item Large standard library
    \item Extensive ecosystem of third-party libraries
\end{itemize}

A simple ``Hello, World!'' program in Python looks like this:

\begin{lstlisting}[language=Python]
print("Hello, World!")
\end{lstlisting}

\noindent This outputs the text \texttt{Hello, World!}. Note that Python does not require semicolons or other boilerplate code, making it easy to get started.

\subsection*{Best Practices}
\begin{itemize}
    \item Practice coding daily to build familiarity with the syntax.
    \item Follow PEP 8 style guidelines (indent using four spaces, limit line length, etc.).
    \item Use comments (\# This is a comment) to clarify intent or non-obvious logic.
\end{itemize}

\section{Variables and Data Types}
Variables store data values in Python. Python is dynamically typed, meaning you do not need to declare a type before assigning a value, but it is also strongly typed, so you cannot combine incompatible types without explicit conversion.

Common data types include:
\begin{itemize}
    \item \textbf{int} (e.g., 5, -3, 0)
    \item \textbf{float} (e.g., 3.14, -0.001)
    \item \textbf{bool} (\texttt{True} or \texttt{False})
    \item \textbf{str} (e.g., \texttt{"Hello"})
    \item \textbf{NoneType} (\texttt{None})
\end{itemize}

Example:
\begin{lstlisting}[language=Python]
x = 10                 # int
y = 3.5                # float
name = "Alice"         # str
is_student = True      # bool

print(type(x))         # <class 'int'>
print(type(name))      # <class 'str'>

# Type conversion (casting):
z = int("123")         # 123 (int)
w = str(5)             # "5" (str)
\end{lstlisting}

\subsection*{Best Practices}
\begin{itemize}
    \item Use meaningful variable names that reflect the stored value.
    \item Follow snake\_case convention for variable/function names.
    \item Be consistent with naming and avoid using reserved keywords (e.g., \texttt{if}, \texttt{for}, \texttt{and}).
\end{itemize}

\section{Operators and Expressions}
Python supports numerous operators for arithmetic, comparison, logical, and more. Expressions combine variables and operators to produce a value.

\subsection{Arithmetic Operators}
\begin{lstlisting}[language=Python]
a = 10
b = 3

print(a + b)    # 13
print(a - b)    # 7
print(a * b)    # 30
print(a / b)    # 3.3333...
print(a // b)   # 3 (floor division)
print(a % b)    # 1 (remainder)
print(a ** b)   # 1000 (10^3)
\end{lstlisting}

\subsection{Comparison and Logical Operators}
\begin{itemize}
    \item \texttt{==}, \texttt{!=}, \texttt{>}, \texttt{<}, \texttt{>=}, \texttt{<=}
    \item \texttt{and}, \texttt{or}, \texttt{not}
\end{itemize}

Example:
\begin{lstlisting}[language=Python]
x = 5
print(x > 0 and x < 10)  # True
print(not (x > 0))       # False
\end{lstlisting}

\subsection*{Best Practices}
\begin{itemize}
    \item Use parentheses for clarity in complex expressions.
    \item Distinguish between \texttt{=} (assignment) and \texttt{==} (comparison).
    \item Use augmented assignment (e.g., \texttt{x += 1}) for conciseness.
\end{itemize}

\section{Control Flow (if-else, loops)}
Control flow statements allow conditional execution and repetition of code blocks.

\subsection{if-elif-else}
\begin{lstlisting}[language=Python]
x = 7
if x > 0:
    print("x is positive")
elif x == 0:
    print("x is zero")
else:
    print("x is negative")
\end{lstlisting}

\subsection{for Loop}
\begin{lstlisting}[language=Python]
fruits = ["apple", "banana", "cherry"]
for fruit in fruits:
    print(fruit)

for i in range(5):
    print(i)  # prints 0,1,2,3,4
\end{lstlisting}

\subsection{while Loop}
\begin{lstlisting}[language=Python]
count = 3
while count > 0:
    print(count)
    count -= 1
print("Blast off!")
\end{lstlisting}

\subsection*{Loop Control Keywords}
\begin{itemize}
    \item \texttt{break} (exit loop immediately)
    \item \texttt{continue} (skip remainder of loop body for current iteration)
\end{itemize}

\begin{lstlisting}[language=Python]
for number in range(1, 10):
    if number == 5:
        break
    if number % 2 == 0:
        continue
    print(number)
\end{lstlisting}

\subsection*{Best Practices}
\begin{itemize}
    \item Indent consistently using 4 spaces.
    \item Avoid deeply nested logic; refactor into functions if needed.
    \item Use \texttt{for i in range(n)} for numeric loops and \texttt{for x in iterable} for collections.
\end{itemize}

\section{Functions and Modules}
Functions group code into reusable units, while modules organize related functions and objects into files.

\subsection{Functions}
\begin{lstlisting}[language=Python]
def greet(name):
    """Return a greeting message."""
    return f"Hello, {name}!"

print(greet("Alice"))  # Hello, Alice!
\end{lstlisting}

Functions can have default parameter values:
\begin{lstlisting}[language=Python]
def power(base, exponent=2):
    return base ** exponent

print(power(5))     # 25
print(power(5, 3))  # 125
\end{lstlisting}

\subsection{Modules and Imports}
A Python file (\texttt{.py}) is a module. You can import standard or third-party modules, or your own.

\begin{lstlisting}[language=Python]
import math
print(math.sqrt(16))  # 4.0
print(math.pi)        # 3.141592653589793

from random import randint
print(randint(1, 10))
\end{lstlisting}

\subsection*{Best Practices}
\begin{itemize}
    \item Keep functions focused on a single task.
    \item Use docstrings (triple-quoted strings) to document function behavior.
    \item Organize related functions into modules.
    \item Import modules at the top of your file.
\end{itemize}

\section{Basic Data Structures: Lists, Tuples, Dictionaries, Sets}
Python provides built-in data structures for storing collections.

\subsection{Lists}
Ordered, mutable collections:
\begin{lstlisting}[language=Python]
fruits = ["apple", "banana", "cherry"]
fruits[0] = "avocado"
fruits.append("date")
\end{lstlisting}

Common methods include \texttt{append()}, \texttt{remove()}, \texttt{pop()}, \texttt{sort()}, etc.

\subsection{Tuples}
Ordered, \textbf{immutable} collections:
\begin{lstlisting}[language=Python]
point = (3, 4)
# point[0] = 5  # Error (immutable)
\end{lstlisting}

\subsection{Dictionaries}
Unordered key-value pairs:
\begin{lstlisting}[language=Python]
student = {"name": "Alice", "age": 25}
student["age"] = 26
student["grade"] = "A"
\end{lstlisting}

\subsection{Sets}
Unordered collections of \textbf{unique} items:
\begin{lstlisting}[language=Python]
nums = {1, 2, 3, 2, 1}
print(nums)  # {1, 2, 3}
nums.add(4)
\end{lstlisting}

\subsection*{Best Practices}
\begin{itemize}
    \item Use lists for ordered, mutable data; tuples for immutable sequences; dictionaries for key-value mappings; sets for unique elements.
    \item Choose the correct data structure for readability and efficiency.
\end{itemize}

\section{File Handling Basics}
Python handles file I/O through the \texttt{open()} function. Common modes are \texttt{"r"} (read), \texttt{"w"} (write), and \texttt{"a"} (append). Use \texttt{with} to ensure proper closure.

\begin{lstlisting}[language=Python]
with open("example.txt", "w") as f:
    f.write("Hello, file!\n")

with open("example.txt", "r") as f:
    content = f.read()
    print(content)
\end{lstlisting}

\subsection*{Best Practices}
\begin{itemize}
    \item Use \texttt{with open(...)} to automatically close files.
    \item Handle exceptions (e.g., \texttt{FileNotFoundError}) when reading files.
    \item Choose appropriate mode (\texttt{"r"}, \texttt{"w"}, \texttt{"a"}).
\end{itemize}

\section{Introduction to Object-Oriented Programming (OOP)}
In OOP, you bundle data (attributes) and methods (functions) into classes. An \textbf{object} is an instance of a class.

\subsection{Classes and Objects}
\begin{lstlisting}[language=Python]
class Dog:
    def __init__(self, name, age):
        self.name = name
        self.age = age

    def bark(self):
        print(f"{self.name} says Woof!")

dog1 = Dog("Buddy", 3)
dog2 = Dog("Max", 5)

dog1.bark()  # Buddy says Woof!
\end{lstlisting}

\subsection*{Best Practices}
\begin{itemize}
    \item Name classes in PascalCase (e.g., \texttt{Dog}, \texttt{Student}).
    \item Use \texttt{self} for instance methods.
    \item Keep the constructor simple (just set up attributes).
\end{itemize}

\section{Exception Handling}
Exceptions are runtime errors that can be caught and handled. Use \texttt{try}/\texttt{except} blocks to handle them gracefully.

\begin{lstlisting}[language=Python]
try:
    num = int(input("Enter a number: "))
    result = 100 / num
    print("Result:", result)
except ValueError:
    print("Invalid input. Please enter a valid integer.")
except ZeroDivisionError:
    print("Cannot divide by zero.")
finally:
    print("Program complete.")
\end{lstlisting}

\subsection*{Best Practices}
\begin{itemize}
    \item Catch specific exceptions (e.g., \texttt{ValueError}, \texttt{ZeroDivisionError}).
    \item Use a \texttt{finally} block for cleanup actions if necessary.
    \item Keep the \texttt{try} block small to handle errors precisely.
\end{itemize}

\section{Homework Assignment: Contact Book Project}

\noindent \textbf{Task:} Build a small \textbf{Contact Book} application that allows a user to:
\begin{itemize}
    \item Add a new contact (name and phone number).
    \item View all contacts.
    \item Search for a contact by name.
    \item Save contacts to a file (and optionally load existing contacts on startup).
\end{itemize}

\noindent \textbf{Requirements:}
\begin{itemize}
    \item Use a \textbf{dictionary} or similar structure to store contacts, with names as keys.
    \item Write at least one \textbf{function} to avoid code repetition.
    \item Use \textbf{file handling} to save or load contact data.
    \item (Optional) Use a simple \textbf{class} to represent a Contact.
    \item Use \textbf{try/except} to handle errors (e.g., file not found).
\end{itemize}

\noindent \textbf{Implementation Hints:}
\begin{enumerate}
    \item Present a menu in a loop: 1) Add, 2) View, 3) Search, 4) Quit.
    \item For ``Add Contact'', prompt for name and phone number, and store them.
    \item For ``View Contacts'', list them all, or report if none.
    \item For ``Search Contact'', prompt for a name, then find it in the dictionary.
    \item Use \texttt{with open(..., "w")} to save contacts to a text file before quitting.
    \item Optionally, load existing contacts at program start if the file exists.
\end{enumerate}

\noindent This exercise reinforces core Python concepts (control flow, data structures, file I/O, OOP, exception handling) and practices your skills in structuring and sharing a project.

\chapter{Exercise 3: Data Ingestion \& Preprocessing}

\section{Introduction}
Multimodal AI applications often require handling multiple data types (e.g., text, images, audio, sensor data). Preparing these data sources involves various tasks such as reading files, cleaning noise, normalizing formats, and storing data in efficient structures. Proper data ingestion and preprocessing are critical steps that can greatly influence model performance and reliability. 

\noindent This excercise provides an overview of:
\begin{itemize}
    \item Common data ingestion techniques for text, image, and audio
    \item Preprocessing steps such as normalization, tokenization, feature extraction
    \item Best practices in organizing and validating multimodal data
\end{itemize}

\section{Data Ingestion: General Workflow}
\noindent In a multimodal pipeline, you may gather data from:
\begin{itemize}
    \item Files (e.g., CSV, JSON, text documents, image folders, WAV files)
    \item Remote data sources (web APIs, cloud storage)
    \item Streaming or real-time sensors (e.g., camera feeds, microphones, wearable devices)
\end{itemize}

\noindent \textbf{Key steps in data ingestion:}
\begin{enumerate}
    \item \textbf{Identify Sources:} Determine all possible input formats and data locations.
    \item \textbf{Extract \& Load:} Read files or streams into a Python environment using libraries like \texttt{pandas}, \texttt{PIL}, \texttt{librosa}, \texttt{cv2}, or custom loaders.
    \item \textbf{Validate \& Clean:} Check for missing or corrupt data, inconsistent labels, and invalid file paths.
    \item \textbf{Transform \& Organize:} Convert raw data into standard formats or consistent shapes (e.g., images resized to a fixed dimension, text tokenized into sequences).
    \item \textbf{Store for Efficiency:} Optionally cache or store intermediate processed data in efficient formats (HDF5, TFRecord, LMDB, etc.) to accelerate repeated training or inference runs.
\end{enumerate}

\section{Text Data Ingestion}
Text is a core modality in chatbots and many AI tasks. Preprocessing tasks typically include:
\begin{itemize}
    \item Reading text from files or databases
    \item Normalizing strings (case-folding, punctuation removal)
    \item Tokenizing (splitting into words or subword units)
    \item Converting tokens to integer IDs (if using a vocabulary-based approach)
    \item Handling special tokens (start/end of sentence, unknown tokens, etc.)
\end{itemize}

\subsection{Reading and Cleaning Text}
To illustrate, assume you have a large text file (e.g., \texttt{dataset.txt}):

\begin{lstlisting}[language=Python]
import string

def load_text_file(filepath):
    lines = []
    with open(filepath, "r", encoding="utf-8") as f:
        for line in f:
            line = line.strip()
            # Remove punctuation (simple approach)
            line = line.translate(str.maketrans('', '', string.punctuation))
            # Convert to lower case
            line = line.lower()
            if line:
                lines.append(line)
    return lines

text_data = load_text_file("dataset.txt")
print(len(text_data), "lines loaded.")
\end{lstlisting}

\noindent Here, we:
\begin{itemize}
    \item \texttt{strip()} each line to remove leading and trailing whitespace
    \item Remove punctuation using \texttt{str.translate}
    \item Convert text to lowercase for uniformity
\end{itemize}

\subsection{Tokenization}
After cleaning, the next step is splitting into tokens:
\begin{lstlisting}[language=Python]
def tokenize(text):
    return text.split()  # simplest form of tokenization

all_tokens = []
for line in text_data:
    tokens = tokenize(line)
    all_tokens.append(tokens)

print(all_tokens[0])  # example of first line token list
\end{lstlisting}

\noindent Advanced tokenizers (e.g., for subwords, Byte Pair Encoding) can be used for sophisticated NLP pipelines, but even simple token splits provide a foundation for basic text tasks.

\section{Image Data Ingestion}
Images often require standardized formats (e.g., same width, height, color channels) and numerical normalization (e.g., pixel values scaled to [0,1] or standardized). 

\subsection{Reading Images}
\noindent Two common methods:
\begin{itemize}
    \item \textbf{PIL (Python Imaging Library)} via \texttt{Pillow}
    \item \textbf{OpenCV} (C++ library with Python bindings), often imported as \texttt{cv2}
\end{itemize}

\begin{lstlisting}[language=Python]
from PIL import Image
import os

def load_images_from_folder(folder_path):
    images = []
    for filename in os.listdir(folder_path):
        if filename.endswith(".jpg") or filename.endswith(".png"):
            img_path = os.path.join(folder_path, filename)
            try:
                img = Image.open(img_path).convert("RGB")
                images.append(img)
            except Exception as e:
                print(f"Error loading {filename}: {e}")
    return images

image_list = load_images_from_folder("images/")
print("Loaded", len(image_list), "images.")
\end{lstlisting}

\subsection{Preprocessing and Normalization}
\noindent Common steps:
\begin{itemize}
    \item \textbf{Resize} to a fixed dimension for batch processing (e.g., 224x224 for many CNNs).
    \item \textbf{Convert to NumPy or PyTorch tensors}.
    \item \textbf{Normalize} pixel values, e.g., scale to [0,1] or subtract mean and divide by standard deviation (common in pretrained models).
\end{itemize}

\begin{lstlisting}[language=Python]
import numpy as np

def preprocess_image(img, desired_size=(224,224)):
    # Resize
    img_resized = img.resize(desired_size)
    # Convert to numpy array
    arr = np.array(img_resized, dtype=np.float32)
    # Scale to [0,1]
    arr /= 255.0
    # (Optional) if using a model that needs mean/std normalization:
    # arr = (arr - mean) / std
    return arr
\end{lstlisting}

\noindent If you use frameworks like PyTorch, you can leverage \texttt{torchvision.transforms} (e.g., \texttt{transforms.Resize}, \texttt{transforms.ToTensor()}) to automate these steps.

\section{Audio Data Ingestion}
Handling audio involves reading waveform data and possibly extracting features (MFCC, spectrograms). Popular libraries include:
\begin{itemize}
    \item \texttt{librosa} (user-friendly, focuses on music and speech analysis)
    \item \texttt{torchaudio} (PyTorch-based audio I/O and transforms)
\end{itemize}

\subsection{Reading Audio Files}
\begin{lstlisting}[language=Python]
import librosa
import numpy as np

def load_audio_file(file_path, sr=16000):
    # sr is the target sampling rate
    audio, sample_rate = librosa.load(file_path, sr=sr)
    return audio, sample_rate

audio_data, sr = load_audio_file("example.wav")
print(f"Loaded audio with {len(audio_data)} samples at {sr} Hz")
\end{lstlisting}

\subsection{Extracting Features}
Many speech or sound classification tasks use features like Mel-Frequency Cepstral Coefficients (MFCCs) or spectrograms as inputs to the model:

\begin{lstlisting}[language=Python]
def extract_mfcc(audio, sr=16000, n_mfcc=13):
    mfccs = librosa.feature.mfcc(y=audio, sr=sr, n_mfcc=n_mfcc)
    return mfccs  # shape: (n_mfcc, time_frames)

mfcc_features = extract_mfcc(audio_data, sr)
print("MFCC shape:", mfcc_features.shape)
\end{lstlisting}

\noindent This produces a 2D representation of audio. You might further transform or normalize it depending on the model’s requirements.

\section{Data Labeling \& Organization}
To train supervised models, you need labels. For multimodal data:
\begin{itemize}
    \item \textbf{Text classification:} Each line or document has a category (spam/ham, sentiment label, etc.).
    \item \textbf{Image classification:} Images grouped by folders or a CSV mapping filenames to labels.
    \item \textbf{Audio tasks:} A JSON or CSV that lists audio filenames and their respective classes or transcripts.
\end{itemize}

\noindent Best practice is to maintain a single metadata file (e.g., \texttt{data.csv}) that maps each sample’s filename or ID to its label. This file can also track other attributes like image resolution, domain, or any additional annotations.

\section{Common Best Practices in Data Ingestion}
\begin{itemize}
    \item \textbf{Check data integrity:} Ensure paths are valid and files are not corrupt. 
    \item \textbf{Handle missing data:} Decide how to fill, drop, or otherwise manage incomplete records. 
    \item \textbf{Balance classes:} If dealing with classification tasks, note class imbalance and plan sampling or weighting strategies.
    \item \textbf{Split early into train/val/test sets:} Avoid data leakage by separating these splits before heavy preprocessing.
    \item \textbf{Avoid manual re-processing:} Cache or store preprocessed data so you don’t redo expensive operations every run (especially for large images, audio files).
    \item \textbf{Documentation:} Keep records of how data was collected, labeled, and transformed for reproducibility. 
\end{itemize}

\section{Example: Multi-Step Pipeline Structure}
For larger projects, you might automate each ingestion/preprocessing step:

\begin{lstlisting}[language=Python]
def pipeline_run():
    # Step 1: Read metadata
    metadata = load_csv("metadata.csv")
    
    # Step 2: Process text entries
    text_records = []
    for item in metadata:
        text_path = item["text_file"]
        text_lines = load_text_file(text_path)
        # Possibly tokenize or clean here
        text_records.append(text_lines)
    
    # Step 3: Process images
    image_records = []
    for item in metadata:
        img_path = item["image_file"]
        img = Image.open(img_path).convert("RGB")
        arr = preprocess_image(img)
        image_records.append(arr)
    
    # Step 4: Process audio
    audio_records = []
    for item in metadata:
        audio_path = item["audio_file"]
        audio, sr = load_audio_file(audio_path, sr=16000)
        mfcc = extract_mfcc(audio, sr)
        audio_records.append(mfcc)
    
    # Step 5: Save or return preprocessed data
    # e.g., save to .npy or custom format
    # ...
    return text_records, image_records, audio_records

text_data, img_data, audio_data = pipeline_run()
print("Text, image, and audio data processed.")
\end{lstlisting}

\noindent This skeleton helps keep each stage of ingestion explicit and organized.

\section{Closing Remarks}
Data ingestion and preprocessing are fundamental to building robust multimodal AI systems. Whether you are working on chatbots that combine text and image inputs or sophisticated speech-vision applications, carefully handling data before modeling can dramatically improve performance and maintainability.

\noindent In the next steps (Model Building and Acceleration with Groq, and Building a Multimodal Chatbot Interface with Gradio), we will see how to feed these cleaned and preprocessed inputs into advanced architectures, and subsequently provide a user-friendly interface for real-time interaction.

\chapter{Exercise 4: Building a Multimodal LLM with Groq}

\section{Introduction}
This exercise guides you in using the \textbf{Groq API} to build a simple \emph{multimodal} large language model pipeline. The pipeline combines:
\begin{itemize}
    \item \textbf{Image understanding}: Provide an image and get a textual interpretation.
    \item \textbf{Speech-to-text}: Convert audio input to text.
    \item \textbf{LLM inference}: Feed the extracted text into a large language model for additional reasoning or summarization.
\end{itemize}

The project is \textbf{inference only}: we will \emph{not} train or fine-tune the model. Instead, we'll tap into a pre-existing LLM or relevant sub-models, running them on Groq hardware/resources with a free tier account.

\section{Objectives}
\begin{enumerate}
    \item Familiarize with Groq's free tier environment.
    \item Understand how to set up and query Groq endpoints for image and speech-based tasks.
    \item Chain the outputs from speech or image modules into an LLM for advanced text-based responses.
    \item Practice secure handling of API keys, avoiding public commits of sensitive credentials.
\end{enumerate}

\section{Prerequisites}
\begin{itemize}
    \item \textbf{Python 3.7+} or a version compatible with your chosen Groq SDK/clients.
    \item Basic familiarity with \textbf{image processing} (e.g., Pillow or OpenCV) and \textbf{audio handling} (e.g., \texttt{librosa} or other speech libraries) if needed locally.
    \item A \textbf{Groq account} with free-tier access. See Groq documentation for sign-up steps.
\end{itemize}

\section{Project Structure}
Below is a high-level structure you can adapt:
\begin{enumerate}
    \item \textbf{Environment Setup:} Install any needed Python libraries and configure your Groq credentials. 
    \item \textbf{Image Preprocessing and Upload}: Send an image to a Groq endpoint (if available) that handles image-to-text or feature extraction.
    \item \textbf{Speech-to-Text Module}: Convert a short audio clip to text using a speech model deployed on Groq (or a local library, if Groq provides only the LLM endpoint).
    \item \textbf{LLM Inference Step}: Combine text from image understanding or speech recognition with user prompts, then send to a Groq-based LLM for a final response.
    \item \textbf{API Security}: Store and load your credentials safely (e.g., environment variables, do not upload keys to GitHub).
\end{enumerate}

\section{Step 1: Environment Setup}
\subsection{Python Dependencies}
A sample \texttt{requirements.txt} might include:
\begin{lstlisting}[language=bash]
groq-api==<your_version>
requests
python-dotenv
Pillow
librosa
# ... any others you need ...
\end{lstlisting}

\subsection{Installing and Configuring Groq CLI (if required)}
Depending on Groq's recommended workflow:
\begin{itemize}
    \item Install the \texttt{groq} CLI or Python SDK: \texttt{pip install groq-cli} or check documentation.
    \item Ensure you have a \textbf{free-tier} account and retrieve your API key from the Groq dashboard.
\end{itemize}

\section{Step 2: Managing API Keys Securely}
\textbf{Never commit your API keys to GitHub.} Instead:
\begin{itemize}
    \item Create a \textbf{.env} file locally (excluded from version control via \texttt{.gitignore}).
    \item Store your credentials there, e.g.:
    \begin{lstlisting}[language=bash]
GROQ_API_KEY="sk-1234abcd..."
    \end{lstlisting}
    \item Load them in Python using \texttt{python-dotenv}:
\begin{lstlisting}[language=Python]
import os
from dotenv import load_dotenv

load_dotenv()  # reads the .env file
GROQ_API_KEY = os.getenv("GROQ_API_KEY")
\end{lstlisting}
    \item If your client library requires environment variables, set them prior to running your script, e.g., \texttt{export GROQ\_API\_KEY=sk\-1234abcd}.
\end{itemize}

\section{Step 3: Image Understanding Endpoint}
Some multimodal pipelines rely on an \emph{image captioning} model or \emph{vision transformer} that outputs a textual summary:
\begin{enumerate}
    \item \textbf{Load \& preprocess} the image in Python (optional resizing or format conversion).
    \item \textbf{Send it to Groq's image endpoint} with your API key in the header or query param.
    \item \textbf{Receive} textual output describing the image.
\end{enumerate}

\noindent Pseudocode:
\begin{lstlisting}[language=Python]
import requests
from PIL import Image

def generate_image_caption(img_path, api_key):
    url = "https://api.groq.com/v1/image-caption"
    headers = {"Authorization": f"Bearer {api_key}"}
    
    # Convert image to bytes
    with open(img_path, "rb") as f:
        img_data = f.read()
    
    # Possibly a multipart/form-data request
    files = {"file": (img_path, img_data, "image/png")}
    
    response = requests.post(url, headers=headers, files=files)
    if response.status_code == 200:
        return response.json()["caption"]
    else:
        raise ValueError(f"Error: {response.text}")
\end{lstlisting}

\subsection*{Notes}
\begin{itemize}
    \item Actual endpoint and request format may vary depending on Groq's APIs. Check documentation.
    \item If the endpoint returns features (embeddings) instead of text, you might forward those to your LLM in the next step.
\end{itemize}

\section{Step 4: Speech-to-Text Module}
If Groq provides a speech recognition endpoint:
\begin{enumerate}
    \item \textbf{Record or load an audio file} (e.g., \texttt{.wav}).
    \item \textbf{Send to Groq's STT (speech-to-text)} endpoint.
    \item Receive recognized text, possibly with timestamps or confidence scores.
\end{enumerate}

\noindent Example:
\begin{lstlisting}[language=Python]
def speech_to_text(audio_path, api_key):
    url = "https://api.groq.com/v1/speech2text"
    headers = {"Authorization": f"Bearer {api_key}"}
    
    with open(audio_path, "rb") as f:
        audio_data = f.read()
    files = {"file": (audio_path, audio_data, "audio/wav")}
    
    response = requests.post(url, headers=headers, files=files)
    if response.status_code == 200:
        return response.json()["transcript"]
    else:
        raise ValueError(f"Error: {response.text}")
\end{lstlisting}

\section{Step 5: LLM Inference (Text-based)}
With text from the image or speech, you can now request an LLM to:
\begin{itemize}
    \item Summarize or reformat the recognized text.
    \item Answer a question about the image content.
    \item Generate a next-step response in a conversational context.
\end{itemize}

\noindent Example:
\begin{lstlisting}[language=Python]
def query_groq_llm(prompt, api_key):
    url = "https://api.groq.com/v1/llm"
    headers = {"Authorization": f"Bearer {api_key}"}
    payload = {"prompt": prompt, "max_tokens": 100}
    
    response = requests.post(url, json=payload, headers=headers)
    if response.status_code == 200:
        return response.json()["text"]
    else:
        raise ValueError(f"Error: {response.text}")
\end{lstlisting}

\noindent Then you might chain them:
\begin{lstlisting}[language=Python]
# 1) Get image caption
img_caption = generate_image_caption("image.jpg", GROQ_API_KEY)

# 2) Convert audio to text
speech_text = speech_to_text("audio.wav", GROQ_API_KEY)

# 3) Combine for final LLM request
combined_prompt = f"The user described an image as '{img_caption}' and said '{speech_text}'. Provide a response:"
llm_response = query_groq_llm(combined_prompt, GROQ_API_KEY)
print("LLM says:", llm_response)
\end{lstlisting}

\section{Step 6: Free Tier Constraints}
\begin{itemize}
    \item \textbf{Usage Limits:} Free-tier often has daily/monthly usage caps for requests, tokens, or runtime. Monitor usage to avoid throttling.
    \item \textbf{Model Size/Response Time:} Some endpoints may be slower or have smaller model variants. Adjust prompts accordingly.
    \item \textbf{Rate Limits:} Groq may impose a rate limit; handle HTTP 429 responses by waiting or retrying.
\end{itemize}

\section{Step 7: Putting It All Together}
\noindent In a final script (\texttt{multimodal\_inference.py}), you might:
\begin{enumerate}
    \item Load your \texttt{.env} file for the Groq API key.
    \item Prompt the user for an image path or audio path.
    \item Generate intermediate text from image or audio.
    \item Pass it to the LLM for further processing.
    \item Display or store the result.
\end{enumerate}

\subsection*{Security Reminder}
\begin{itemize}
    \item Always ignore \texttt{.env} in your \texttt{.gitignore}.
    \item If you deploy or share your project, keep the key in a secure location or use environment variables in your production environment (e.g., Docker secrets).
    \item Rotate or revoke keys if they’re accidentally leaked.
\end{itemize}

\section{Conclusion}
By completing this exercise, you gain experience:
\begin{itemize}
    \item Accessing multiple specialized endpoints on Groq (image, speech, LLM).
    \item Structuring a pipeline that goes from raw multimodal input to textual queries for LLM inference.
    \item Handling security best practices for API keys, especially when using a public repository.
\end{itemize}

\noindent This approach can be extended to more sophisticated scenarios:
\begin{itemize}
    \item Fine-tuning a pipeline or caching model responses (if advanced usage is allowed).
    \item Integrating with \textbf{Gradio} or a web UI to build an interactive chatbot (covered in a subsequent lecture).
    \item Incorporating advanced data preprocessing steps (covered in the previous Data Ingestion \& Preprocessing notes).
\end{itemize}

\chapter{Exercise 5: Multimodal Chatbot with Gradio}

\section{Introduction}
This final exercise culminates the previous lessons and demonstrates how to build a fully functional, multimodal chatbot interface using \textbf{Gradio}. Throughout earlier modules, you learned:
\begin{itemize}
    \item \textbf{Data ingestion \& preprocessing}: Techniques for reading, cleaning, and preparing text, images, and audio.
    \item \textbf{Groq-based inference}: Leveraging accelerated hardware or endpoints for tasks like image captioning, speech-to-text, and large language model (LLM) queries.
    \item \textbf{API key handling}: Avoiding the exposure of credentials by storing them in \texttt{.env} files or environment variables.
\end{itemize}

In this project, you will integrate these components into one seamless user experience, where the chatbot can:
\begin{enumerate}
    \item Accept textual queries
    \item Process uploaded images (via image captioning or other vision tasks)
    \item Convert audio to text (speech-to-text)
    \item Combine all inputs into a coherent conversation with an LLM
\end{enumerate}

\noindent By adding \textbf{Gradio}, you provide a simple and interactive web interface for users, allowing them to upload files, speak directly, and receive immediate responses. This real-time demonstration underscores the power of multimodal AI and helps you practice essential skills for modern AI development—particularly bridging data preprocessing and model inference with accessible UI components.

\section{Objectives}
By completing this exercise, students will be able to:
\begin{itemize}
    \item Implement a Gradio interface that handles text, image, and (optionally) audio inputs
    \item Chain together multiple model inference functions (e.g., image understanding, speech recognition, LLM-based reasoning)
    \item Maintain proper security practices for sensitive API credentials
    \item Test and deploy a proof-of-concept multimodal chatbot application
\end{itemize}

\section{Prerequisites}
\begin{itemize}
    \item \textbf{Python 3.7+} (or compatible with Gradio and Groq libraries)
    \item \textbf{Gradio} installed (\texttt{pip install gradio})
    \item Working knowledge of:
    \begin{enumerate}
        \item Data ingestion (text, images, audio)
        \item Groq inference endpoints/functions for vision, speech, and LLM tasks
        \item Environment variable usage for API keys (e.g., \texttt{python-dotenv})
    \end{enumerate}
\end{itemize}

\section{Project Structure}
A suggested folder structure might look like:
\begin{verbatim}
multimodal_chatbot/
  |- .env
  |- .gitignore
  |- main.py           # Main Gradio script
  |- modules/
  |    |- groq_inference.py
  |    |- data_utils.py
  |    |- ...
  |- requirements.txt
  |- README.md
\end{verbatim}

\noindent The exact organization depends on your earlier exercises, but keep separate files for inference logic, data utilities, and the main Gradio app. Ensure \texttt{.env} is listed in your \texttt{.gitignore} file to prevent committing credentials.

\section{Step 1: Load Environment Variables Securely}
\noindent As with previous assignments, you should never hard-code API keys in source code. Instead, do:
\begin{lstlisting}[language=Python]
# main.py

import os
from dotenv import load_dotenv

load_dotenv()
GROQ_API_KEY = os.getenv("GROQ_API_KEY")
\end{lstlisting}

\noindent Confirm \texttt{.env} is excluded from version control:
\begin{lstlisting}
# .gitignore

.env
__pycache__/
*.pyc
\end{lstlisting}

\section{Step 2: Creating the Multimodal Processing Function}
Assume you have helper functions from previous lessons:
\begin{itemize}
    \item \texttt{generate\_image\_caption(img\_path, api\_key)}: returns a text caption for an image
    \item \texttt{speech\_to\_text(audio\_path, api\_key)}: transcribes audio into text
    \item \texttt{query\_groq\_llm(prompt, api\_key)}: returns a text response from the LLM
\end{itemize}

\noindent You can combine them as follows:

\begin{lstlisting}[language=Python]
# modules/groq_inference.py (example usage)

def multimodal_chat(user_text, user_image_path, user_audio_path, api_key):
    context_parts = []

    if user_image_path:
        caption = generate_image_caption(user_image_path, api_key)
        context_parts.append(f"Image: {caption}")

    if user_audio_path:
        spoken_text = speech_to_text(user_audio_path, api_key)
        context_parts.append(f"Audio: {spoken_text}")

    if user_text.strip():
        context_parts.append(f"User typed: {user_text}")

    # Combine into a single prompt
    combined_prompt = "\n".join(context_parts) + "\nProvide a response:"
    response = query_groq_llm(combined_prompt, api_key)
    return response
\end{lstlisting}

\section{Step 3: Building the Gradio Interface}
\noindent In \texttt{main.py}, you can create a Gradio demo that collects three inputs: text, image, and audio. For instance:

\begin{lstlisting}[language=Python]
import gradio as gr
import os
from dotenv import load_dotenv
from modules.groq_inference import multimodal_chat

load_dotenv()
GROQ_API_KEY = os.getenv("GROQ_API_KEY")

def gradio_multimodal_interface(text_input, image_input, audio_input):
    # image_input and audio_input will be file paths if you set type="filepath"
    return multimodal_chat(text_input, image_input, audio_input, GROQ_API_KEY)

with gr.Blocks() as demo:
    gr.Markdown("# Multimodal Chatbot using Groq + Gradio")

    text_box = gr.Textbox(label="Enter your text prompt")
    image_uploader = gr.Image(type="filepath", label="Upload an image (optional)")
    audio_recorder = gr.Audio(type="filepath", label="Record or upload audio (optional)")

    chat_button = gr.Button("Send")
    output_box = gr.Textbox(label="Chatbot Response")

    chat_button.click(
        fn=gradio_multimodal_interface,
        inputs=[text_box, image_uploader, audio_recorder],
        outputs=output_box
    )

if __name__ == "__main__":
    demo.launch()
\end{lstlisting}

\subsection*{Usage}
\begin{itemize}
    \item Run \texttt{python main.py}
    \item Visit \texttt{http://127.0.0.1:7860} in your browser
    \item Enter text, optionally add an image or audio, then click \texttt{Send}
    \item Observe the chatbot's reply in \texttt{output\_box}
\end{itemize}

\section{Step 4: Handling Data Formats Properly}
\noindent Depending on your exact function signatures:
\begin{itemize}
    \item \textbf{Images}: If your inference function needs an actual image object, you may do:
    \begin{lstlisting}[language=Python]
from PIL import Image

img = Image.open(image_input)  # if you prefer PIL
    \end{lstlisting}
    \item \textbf{Audio}: If you used \texttt{librosa} or other libraries, be mindful of sampling rate vs. the required input for your Groq endpoint. 
\end{itemize}

\section{Step 5: Security Tips and Credentials on GitHub}
\begin{itemize}
    \item Store keys in \texttt{.env}; never commit them.
    \item \texttt{.gitignore} your \texttt{.env} file.
    \item If deploying, set environment variables on your hosting platform rather than bundling them with the code.
\end{itemize}

\section{Step 6: Advanced Features (Optional)}
\begin{itemize}
    \item Maintain conversation \textbf{history} so that each user prompt builds upon previous responses.
    \item Implement a \textbf{session-based} approach in Gradio using \texttt{State} to track multi-turn dialogues.
    \item Add \textbf{logging or analytics} to measure how often image or audio inputs are used.
    \item Customize the \textbf{UI layout} with additional Gradio elements (e.g., \texttt{Tabs}, \texttt{Columns}, \texttt{Dropdowns}).
\end{itemize}

\section{Final Instructions}
\begin{itemize}
    \item \textbf{Submit your repository link} once complete. Your \texttt{README.md} should clearly explain how to install dependencies, set up environment variables, and launch the Gradio interface.
    \item \textbf{Demonstrate usage} via screenshots or screen recordings showing the chatbot responding to multimodal inputs.
    \item Verify that \texttt{.env} is \emph{not} tracked in version control. 
\end{itemize}

\section{Conclusion}
By integrating \textbf{Gradio} with your Groq-based inference functions, you create an accessible, user-friendly \textbf{multimodal chatbot} that serves as a capstone demonstration of everything learned throughout the course. This hands-on experience combines data preprocessing, model orchestration, and interactive user interfaces, providing the foundational skills for real-world AI applications.

\bibliographystyle{plain}
\bibliography{references}
\end{document}